\documentclass{article}

\usepackage{hyperref}

\usepackage{etoolbox}
\providebool{istwocolumn}
\providebool{usebiblatex}
\providebool{loadgeometry}
\providebool{loadhyperref}
\providebool{loadalgorithm2e}
\providebool{altermargins} 
\providebool{twocol}

\setbool{istwocolumn}{true}
\setbool{usebiblatex}{false}
\setbool{loadgeometry}{false}
\setbool{loadhyperref}{false}
\setbool{loadalgorithm2e}{false}
\setbool{altermargins}{false}
\setbool{twocol}{true}

\input{preamble_pkgs}   
\newcommand{\sref}[1]{\S\ref{#1}}

\newcommand{\eg}{e.g., }

\newcommand{\ie}{i.e., }

\newtheorem{theorem}{Theorem}[section]
\newtheorem{definition}[theorem]{Definition}
\newtheorem{proposition}[theorem]{Proposition}

\newtheorem{lemma*}{Lemma}

\mathchardef\mhyphen="2D 

\def\R{\mathbb{R}}

\def\A{\mathbf{A}}

\def\P{\mathbf{P}}

\def\X{\mathbf{X}}

\def\x{\mathbf{x}}
\def\y{\mathbf{y}}

\def\0{\mathbf{0}}

\def\cC{\mathcal{C}}

\def\cF{\mathcal{F}}

\def\cN{\mathcal{N}}

\def\cP{\mathcal{P}}

\def\cT{\mathcal{T}}

\def\cV{\mathcal{V}}
\def\cW{\mathcal{W}}
\def\cX{\mathcal{X}}
\def\cY{\mathcal{Y}}
\def\cZ{\mathcal{Z}}

\DeclareMathOperator*{\argmin}{argmin}

\DeclarePairedDelimiterX{\klx}[2]{(}{)}{#1\;\delimsize\|\;#2}

\DeclarePairedDelimiterX{\relentx}[2]{(}{)}{#1\;\delimsize|\;#2}

\DeclarePairedDelimiterX{\inner}[2]{\langle}{\rangle}{#1, #2} 

\def\st{\medspace|\medspace}

\newcommand{\suchthat}{\;\ifnum\currentgrouptype=16 \middle\fi|\;}

\def\OT{\textup{OT}}

\def\W{\textup{W}}                      
\def\intx{\int_{\mathcal{X}}}

\newcommand{\fvar}[2][]{\frac{\delta#1}{\delta#2}}
\newcommand{\tfvar}[2][]{\tfrac{\delta#1}{\delta#2}}

\makeatletter
\renewcommand*\env@matrix[1][*\c@MaxMatrixCols c]{%
  \hskip -\arraycolsep
  \let\@ifnextchar\new@ifnextchar
  \array{#1}}
\makeatother

\newcommand{\mnist}{\textsc{mnist}\xspace}
\newcommand{\cifar}{\textsc{cifar10}\xspace}
\newcommand{\usps}{\textsc{usps}\xspace}

\newcommand{\emnist}{\textsc{emnist}\xspace}
\newcommand{\kmnist}{\textsc{kmnist}\xspace}

\usepackage{color}

\usepackage[accepted]{icml2021}

\icmltitlerunning{Dataset Dynamics via Gradient Flows in Probability Space}

\begin{document}

\twocolumn[
\icmltitle{Dataset Dynamics via Gradient Flows in Probability Space}

\begin{icmlauthorlist}
\icmlauthor{David Alvarez-Melis}{msr}
\icmlauthor{Nicol\`o Fusi}{msr}
\end{icmlauthorlist}

\icmlaffiliation{msr}{Microsoft Research}

\icmlcorrespondingauthor{David Alvarez-Melis}{alvarez.melis@microsoft.com}

\icmlkeywords{Transfer Learning, Gradient Flows, Wasserstein Distances}

\vskip 0.3in
]

\printAffiliationsAndNotice{}  

\begin{abstract}{}
    Various machine learning tasks, from generative modeling to domain adaptation, revolve around the concept of dataset transformation and manipulation. While various methods exist for transforming unlabeled datasets, principled methods to do so for \textit{labeled} (e.g., classification) datasets are missing. In this work, we propose a novel framework for dataset transformation, which we cast as optimization over data-generating joint probability distributions. We approach this class of problems through Wasserstein gradient flows in probability space, and derive practical and efficient particle-based methods for a flexible but well-behaved class of objective functions. Through various experiments, we show that this framework can be used to impose constraints on classification datasets, adapt them for transfer learning, or to re-purpose fixed or black-box models to classify\,---with high accuracy---\,previously unseen datasets. 
\end{abstract}

\section{Introduction}

One of the hallmarks of machine learning practice is the relative scarcity of domain- and task-specific data, compared to the abundance of general-purpose data. Thus, many problems in machine learning involve dataset manipulation or transformation to re-weight, sub-sample, augment, compress, or build generative models of data. 

Recent work on these problems has focused on unlabeled datasets, either by considering only unsupervised learning settings or ignoring labels in supervised ones. Manipulation of labeled datasets, while less explored, is appealing for its various potential applications. On the one hand, it can unify, formalize, and cast under a new light various data manipulation heuristics core to state-of-the-art supervised learning pipelines, such as data augmentation \citep{simard2003best, krizhevsky2012imagenet, cubuk2019autoaugment}, dataset pooling, or instance mixing \citep{zhang2017mixup}. On the other hand\;---and our principal motivation in this work---\;is the potential of such a framework to tackle less explored or entirely novel problems, such as `shaping' datasets by imposing certain (e.g., geometric or privacy) constraints, or transforming them from one domain to the other to recast transfer learning as transfer of data to the domain of expertise of a model, rather than the other way around. Current data transformation methods are ill suited for these tasks, since they either operate exclusively on features or use labels sparingly and heuristically. In addition, most of them are tailored to specific problems and cannot be easily generalized. Thus, a unified framework for transforming and manipulating datasets is missing.

In response to this, in this work we propose a principled, flexible, and computationally feasible approach to \textit{labeled} dataset transformation. Based on the motivating applications described above, we seek a framework that is model-independent, depending only on intrinsic geometric properties of the data, and applicable to any classification dataset, regardless of size, dimensionality, or number of classes. Our first step towards achieving this is to view datasets as samples from a joint distribution, and to focus on manipulating these distributions instead. Indeed, while a dataset might consist of finitely many samples, the precise number is seldom relevant and often unspecified (\eg in streaming settings). Instead, the true object of interest is the generative process that gives rise to the dataset. Thus, we characterize a dataset as a collection of samples from an (unknown) joint probability distribution $\rho(x,y)$, and cast its transformation as an optimization problem in probability space $\mathcal{P}(\cX \times \cY)$. Formally, we seek to solve problems of the form $\min_{\rho \in \mathcal{P}(\cX \times \cY) }F(\rho)$, for some functional $F$ over distributions that encodes the transformation of interest.

Formalizing and solving optimization problems over such\;---infinite dimensional, non-euclidean---\;probability spaces is conceptually challenging. In this work, we do so by means of \textit{gradient flows}, a linchpin of applied mathematics for modeling dynamics in very general infinite-dimensional spaces \citep{ambrosio2005gradient}, which have in fact been extensively studied in the context of metric measure spaces \citep{santambrogio2017euclidean}, and have deep connections to partial differential equations (PDE) \citep{jordan1998variational}. Gradient flows come with various appealing properties: they are inherently flexible in terms of the underlying space and types of dynamics they can model, they admit rigorous convergence analysis, and they produce\;---in addition to a final minimizing solution---\;a full trajectory of iterates, which are often useful too. It is no surprise then that gradient flows have recently become a popular tool to analyze and derive (parameter) optimization methods in machine learning \citep{javanmard2020analysis, chizat2019sparse}.

But harnessing gradient flows for labeled dataset optimization poses various challenges. First, a suitable representation of feature-label pairs and a meaningful metric between datasets must be defined. Then, one must find a class of functionals that are expressive enough to model interesting objective functions on datasets yet sufficiently well-behaved to enjoy theoretical guarantees and allow for tractable optimization. In response to the first of these challenges, we leverage a recently proposed notion of distance between datasets \citep{alvarez-melis2020geometric} based on optimal transport (OT). Conceptually, this notion endows the space of joint (feature-label) distributions with a metric, allowing us to cast flows in this space in the more general setting of metric measure spaces described above. On a practical level, we can use this distance as a functional to define gradient flows that minimize it, i.e., to encode similarity to a given dataset as an optimization objective. Doing so requires generalizing this distance, which has previously only been used for \textit{static} dataset comparison, and making it differentiable, yielding a loss function that we can efficiently optimize using automatic differentiation. 

To address the second challenge\,---defining suitable functional objectives---\;we borrow a family of flexible functionals over measures that are well-studied in the gradient flow literature, show how they can be re-purposed to encode various handy dataset-related objectives, and discuss in detail how to make use of them within a particle-based approximation of the gradient flow that relies on automatic differentiation. As a result, we end up with a framework for labeled dataset optimization that is flexible, efficient, and has a solid theoretical foundation. 

In our experiments, we first show how this framework for dataset transformation yields novel ways to approach transfer learning problems. For example, it can be used in combination with traditional model adaptation, as a means of data-preprocessing before fine-tuning (\sref{sec:transfer_mnist}), but also \textit{instead of it}, for settings where model adaptation is infeasible (e.g., frozen, black-box, or extremely large models). For such challenging settings, our flow-based methods allow us to `re-purpose' already-trained models to classify previously unseen datasets with high accuracy (\sref{sec:repurposing}). We also show this framework can be used to generate datasets with various geometric constraints (\sref{sec:shaping_experiments}), showing its promise as a principled tool for dataset synthesis.

\section{Related Work}\label{sec:related}

\paragraph{Task dynamics.} \citet{achille2018dynamics} propose a dynamic notion of distance between tasks based on Kolmogorov complexity, and interpret it as a stochastic differential equation. Their notion is defined on parameter space, and is driven by a model-dependent empirical risk minimization loss. Similarly, \citet{wu2020phase} study phase transitions of the information bottleneck, but do so on a learnt representation space, and do not consider gradient flows. 

\paragraph{Dataset adaptation with optimal transport.} Most prior work that uses OT distances for comparing datasets operates exclusively on features (not labels) \citep{seguy2018large-scale}, or relies on a classification loss for the latter \citep{courty2017joint, damodaran2018deepdjot}, which assumes the two label sets are identical. In contrast, the notion of distance between datasets we use in this work, and therefore the resulting framework, does not. Furthermore, these works rely on the static formulation of OT, on a barycentric mapping to transform datasets, and consider only an OT distance as objective. Here instead we rely on more general functional objectives, which might contain an OT distance as one of their components, and use an explicit dynamic formulation of OT to transform the datasets.

\paragraph{Computational Wasserstein gradient flows.} Recent work leverages gradient flows in Wasserstein space for machine learning \citep{liutkus2019sliced, liu2019understanding, arbel2019mmd, kolouri2019generalized, chizat2018global, chizat2019sparse}. Most of these consider flows on model (i.e., parameter) space or feature space, so they are not suitable for handling labeled dataset transformation. 

\paragraph{Generative modeling.} Generative modeling can be understood as optimization in probability space too: its goal is to find a parametrized distribution $\rho_\theta$ that minimizes some notion of dissimilarity to a true data distribution $\rho_{\text{data}}$, i.e., to minimize $F(\rho_{\theta}) = D(\rho_{\theta}, \rho_{\text{data}})$ for some discrepancy $D$. The typical setting assumes that samples from $\rho_{\text{data}} = \argmin F(\rho)$ are available, and that the ultimate goal is to generate this distribution \textit{exactly}. Here we consider more general settings, including those for which the optimal distribution $\rho^*$ is unknown, defined only implicitly as the minimizer of some functional objective $F$. Within generative modeling, there is a flourishing line of work that learns the target distribution by means of \textit{normalizing flows} \citep{dinh2014nice, dinh2017density, kingma2018glow, germain2015made, papamakarios2017masked}, which are sequences of invertible transformations typically parametrized through neural networks. These are different from the \textit{gradient flows} considered here, which are non-parametric, defined implicitly through a functional, and whose infinitesimal dynamics are well understood.   

\section{Technical Background}\label{sec:background}
We begin by introducing the two key notions behind our framework: optimal transport and gradient flows.

\subsection{Setting and Notation}
Let $\cX$ be a Polish space equipped with metric $d$, and $\cP(\cX)$ the set of non-negative Borel measures with finite second-order moment on that space. We consider both continuous and discrete measures, the latter represented as an empirical distribution: $\sum_{i=1}^N p_i\delta_{x_i}$, where $\delta_{x}$ is a Dirac at position $x \in \cX$. When clear from the context, we use $\alpha$ to denote both the measure and its density $\rho_{\alpha}$. For a measure $\alpha$ and measurable map $T\!:\!\cX \rightarrow \cX$, $T_{\sharp}\alpha$ denotes the push-forward measure of $\alpha$ under $T$. For a joint measure $\pi \in \cP(\cX \times \cX)$, we can express its marginals as $P_{1\sharp}\pi$ and $P_{2\sharp} \pi$ using the maps $P_1(x,x')=x$ and $P_2(x,x')=x'$. We focus on supervised learning, so we define a dataset as a collection of feature-label pairs $\{(x^{(i)},y^{(i)}\}_{i}$, which we assume are sampled from some $\alpha \in \cP(\cX \times \cY)$. We denote the dataset as $\mathrm{D}_{\alpha}$ to make this dependence explicit. We will use the shorthand notations: $\cZ\triangleq\cX \times \cY$ and $z\triangleq(x,y)$. Finally, $\nabla \cdot\medspace$ denotes the divergence operator.\looseness=-1

\subsection{Optimal Transport}
For measures $\alpha, \beta\!\in \cP(\cX)$ and cost function $c\!:\!\cX\times\cX\!\!\rightarrow\!\R_{+}$, the optimal transport problem is
\vspace{-0.1cm}
\begin{equation}\label{eq:wasserstein}
    \OT_c(\alpha, \beta) \triangleq \min_{\pi \in \Pi(\alpha, \beta)} \int c(x_1, x_2) \dif \pi(x_1, x_2),
    \vspace{-0.1cm}
\end{equation}
where $\Pi(\alpha, \beta)$ is the set of transport plans between $\alpha$ and $\beta$, i.e., couplings with these two measures as marginals:
\vspace{-0.1cm}
\begin{equation}\label{eq:transportation_polytope}
    \Pi(\alpha, \beta) \triangleq \{\pi \in \cP(\cX\!\times\!\cX) \suchthat P_{1\sharp}\pi = \alpha,  P_{2\sharp}\pi=\beta \}.
    \vspace{-0.1cm}    
\end{equation}
When $c(x,y)=d(x,y)^p$ for $p\geq 1$, $\W_p\triangleq \OT_c(\alpha, \beta)^{1/p}$ is called the p-Wasserstein distance. As its name implies, $\W_p$ defines a \textit{true} distance on $\cP(\cX)$ \citep{villani2008optimal}. Thus, the latter equipped with the former is a metric space $\mathbb{W}_p(\cX) = (\cP(\cX), \W_p)$, called the (p-)Wasserstein space. In practice, a regularized version of Problem \eqref{eq:wasserstein}, with an added entropy term $\lambda \textup{H}(\pi)$, is often solved instead \citep{cuturi2013sinkhorn}. 

The dual formulation of problem \eqref{eq:wasserstein} is:
\vspace{-0.1cm}
\begin{equation}\label{eq:wasserstein_dual}
    \OT_c(\alpha, \beta)  =    \sup_{\varphi\in \cC(\cX)} \intx \varphi \dif \alpha+ \intx \varphi^c \dif \beta,
    \vspace{-0.1cm}
\end{equation}
where $\varphi:\cX \rightarrow \R$ is known as the Kantorovich potential, and $\varphi^c$ is its c-conjugate: $\varphi^c(x) = \inf_{x'\in \cX}c(x,x') - \varphi(x) $.  For $c(x,x')=\|x-x'\|^2$, $\phi^c$ is the Fenchel conjugate. 

In seminal work, \citet{Benamou2000-lk} showed that OT has a \textit{dynamic} formulation too:
\vspace{-0.1cm}
\begin{equation}\label{eq:benamou}
    \W_p^p(\alpha, \beta) = \!\!\min_{\mu_t, V_t } \int_{0}^1\!\!\!\int_{\cX} \|V_t(x)\|^p \dif \mu_t(x) \dif t ,
    \vspace{-0.1cm}
\end{equation}
where the minimum is taken over measure-field pairs satisfying $\mu_0=\alpha, \mu_1 = \beta$, and the continuity equation:
\begin{equation}\label{eq:continuity}
    \vspace{-0.1cm}
    \partial_t \mu_t = - \nabla \cdot (\mu_tV_t).
    \vspace{-0.1cm}
\end{equation}
This formulation amounts to finding, among \textit{paths} of measures $\mu_t$ advecting from $\alpha$ to $\beta$ and velocity fields $V_t$ satisfying a conservation of mass constraint, the `shortest' one, i.e., that which minimizes the path length (formally, the integral of the metric derivative). Thus, the dynamic formulation focuses on \textit{local transfer} (via $\mu_t$), compared to \textit{global correspondence} (via $\pi$) in the static one (Equation~\eqref{eq:wasserstein}).

\subsection{The Optimal Transport Dataset Distance}\label{sec:otdd}
It is appealing to use OT to define a distance between datasets, but this is non-trivial for labeled datasets. The main issue is that problem \eqref{eq:wasserstein} would require an element-wise metric $d$, which for labeled datasets means defining a distance between pairs of feature-label pairs. For the general case where $\cY$ might be a discrete set (\ie classification), this seems daunting. In recent work, \citet{alvarez-melis2020geometric} propose a hybrid metric on this joint space that relies on representing the labels $y$ as distributions over features $\alpha_y$. E.g., for a digit classification dataset, $\alpha_{1}$ would be a distribution over images with label $y=1$. With this, they define a metric on $\cZ$ as $d_{\cZ}(z,z')^p \triangleq  d_{\cX}(x,x')^p  + \text{W}_p^p(\alpha_y, \alpha_{y'})$. Using $d_{\cZ}$ as the ground cost in eq.~\eqref{eq:wasserstein} yields a distance between measures on $\cP(\cZ)$, and therefore between datasets, which they refer to as the Optimal Transport Dataset Distance (OTDD):
\vspace{-0.15cm}
\begin{equation}\label{eq:ottask_general_distance}
	\!\!\textsc{otdd}(\mathrm{D}_{\alpha}, \mathrm{D}_{\beta})\!\triangleq\!\biggl(\!\min_{\pi \in \Pi(\alpha, \beta)}\!\int_{\cZ\!\times\!\cZ}\hspace{-1em}d_{\cZ}(z,z') \dif \pi(z,z')\!\biggr)^{\frac{1}{2}}\!\!.\hspace{-1em}
    \vspace{-0.1cm}
\end{equation}
The main appeal of this distance is that it is defined even if the label sets of the two data sets are non-overlapping, or if there is no explicit known correspondence between them (\eg digits to letters). It achieves this through a purely geometric treatment of features and labels. Another advantage is its computational scalability, which relies on using a Gaussian approximation on the per-label distributions, i.e., modeling each $\alpha_y$ as $\cN(\mu_y, \Sigma_y)$, whose mean and covariance are estimated from samples. In that case, the distances $\text{W}_2^2(\alpha_y, \alpha_{y'})$ can be computed in closed form, so no optimization is needed to evaluate $d_{\cZ}(z,z')$ inside problem~\eqref{eq:ottask_general_distance}.

\subsection{Gradient Flows}
Consider a functional $F: \cX \rightarrow \R$ and a point $x_0 \in \cX$. A \textit{gradient flow} is an absolutely continuous curve $x(t)$ that evolves from $x_0$ in the direction of steepest descent of $F$. When $\cX$ is Hilbertian and $F$ is sufficiently smooth, its gradient flow can be succinctly expressed as the solution of a differential equation $x'(t) = - \nabla F(x(t))$ with initial condition $x(0)=x_0$. Different discretizations of this equation yield popular gradient descent schemes, such as momentum and acceleration \citep{su2016differential, wilson2016lyapunov}.

\section{Dataset Dynamics through Optimization}
Using the notation and concepts introduced in Section~\ref{sec:background}, we now formalize the motivating problem of dataset transformation.  We do so first in general terms here, and discuss specific objective functions in the next section.

\subsection{Functional Minimization via Gradient Flows}\label{sec:into_flows}
Given a dataset objective expressed as a functional $F:\cP(\cZ)\rightarrow \R$, we seek a joint measure $\rho \in \cP(\cZ)$ realizing:
\vspace{-0.15cm}
\begin{equation}
    \min_{\rho \in \cP(\cZ) } F(\rho)    
    \vspace{-0.1cm}
\end{equation}
We propose to approach this problem via gradient flows, \ie by moving along a curve of steepest descent starting at $\rho_0$ until reaching a solution $\rho^*$. Unlike Euclidean settings, here the underlying space $\mathbb{W}_p(\cZ)$ is infinite-dimensional and non-Hilbertian, thus requiring stronger tools. 

First, the notion of derivative can be extended to functionals on measures through the first variation, denoted by $\frac{\delta F}{\delta \rho}$ (and defined in detail in Appendix~\ref{sec:pde_view}). With this, we characterize the gradient flow $(\rho_t)_{t\geq 0}$ of $F$ as the solution of:
\vspace{-0.1cm}
\begin{equation}\label{eq:wass_gradflow_general}
    \partial_t \rho_t = \nabla_{\W}F(\rho_t) \triangleq  \nabla \cdot \left( \rho_t \nabla \tfvar[F]{\rho}(\rho_t) \right), 
\end{equation}
which can also be seen as a continuity equation \eqref{eq:benamou} for the measure $\rho_t$ and the velocity field $- \nabla \tfvar[F]{\rho}(\rho_t) $.

Our main functional of interest will be the Wasserstein distance to a target distribution: $\cT_{\beta}(\rho)\triangleq \W_2(\rho, \beta)$, which we realize using the OTDD (Section~\ref{sec:otdd}). In addition, following the literature on gradient flows \citep{santambrogio2015otam, santambrogio2017euclidean}, we consider three other functional families:
\begin{align}\label{eq:functionals}
    \vspace{-0.1cm}
	\cF(\rho) &= \int f(\rho(z))\dif z \\
	\cV(\rho) &= \int V(z)\dif \rho \\
	\cW(\rho) &= \frac{1}{2}\iint W(z-z')\dif\rho(z)\dif\rho(z')
	\vspace{-0.1cm}
\end{align}
where $f:\R\rightarrow\R$ is convex and superlinear, $V, W:\cX\rightarrow \R$ are convex and sufficiently smooth. These terms have a physical interpretation as internal, potential and interaction energies, respectively. 

This choice of functionals obeys both theoretical and computational motivations. First, gradient flows on these functionals have provable convergence (Appendix~\ref{sec:convergence}), which makes them appealing as optimization objectives. Second, their first variation is simple, making them amenable to gradient-based approaches and automatic differentiation, as discussed in Section~\ref{sec:gradients}. Yet, despite being tractable and well-behaved, these functionals are still sufficiently general to encode various useful objectives and constraints on datasets, as we show in Section~\ref{sec:functionals}.

Hence, for the remainder of this work we assume the objective of interest can be cast as:
\begin{equation}\label{eq:functional_sum}
F(\rho) =  \cT_{\beta}(\rho) + \cF(\rho) + \cV(\rho) + \cW(\rho).
\end{equation}
As mentioned before, the first variations of these four functionals are simple \citep{ambrosio2005gradient}:
\begin{equation}
     \tfvar[\cF]{\rho} = f'(\rho), \medspace
     \tfvar[\cV]{\rho} = V, \medspace
     \tfvar[\cW]{\rho} = W \ast \rho, \medspace
     \tfvar[\cT_{\beta}]{\rho} = \varphi_{\rho},
\end{equation}
where $\ast$ denotes the usual convolution operator between a measurable function and a measure, and $\varphi_\rho$ is the optimal Kantorovich potential in the dual OT formulation \eqref{eq:wasserstein_dual}. In light of this, the gradient flow \eqref{eq:wass_gradflow_general} for functionals of the form \eqref{eq:functional_sum} corresponds to the solution of:
\begin{equation}\label{eq:wass_gradflow_f}
    \partial_t \rho =  \nabla \cdot \bigl(\rho \nabla (f'(\rho) + V + W\ast\rho + \varphi_{\rho}) \bigr).
\end{equation}
Appendix~\ref{sec:pde_view} provides a PDE interpretation of this flow. 

\begin{figure*}
    \centering
    \includegraphics[width=\linewidth, trim={0 19.15cm 11cm 0}, clip]{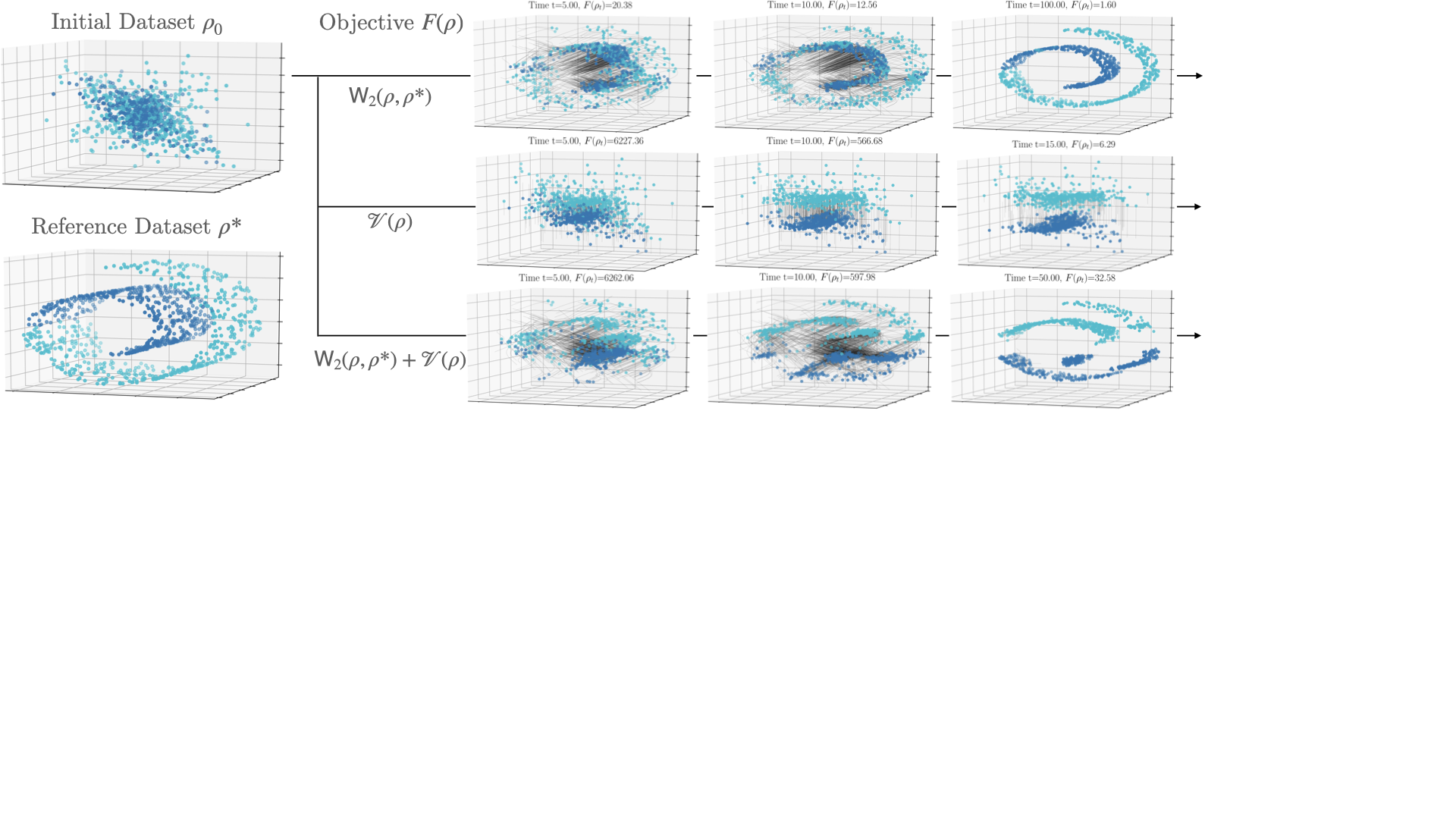}
    \vspace{-7mm}
    \caption{\textbf{Shaping datasets via flows}: our framework allows for simple and principled transformation of classification datasets by following the gradient flow of a functional objective, such as: similarity to a reference dataset (shown here as $\text{W}(\cdot, \rho^*)$, via OTDD), a function enforcing linear separability (shown here as $\mathcal{V}(\rho)$), or a combination thereof.}
    \label{fig:main_diagram}
\end{figure*}

\subsection{Numerical Solution of Gradient Flows}\label{sec:numerical_solution}
The gradient flows defined above have a probability theory counterpart in terms of random variables. Consider a stochastic process $(Z_t)_t$, where each $Z_t\triangleq(X_t, Y_t)$ is a random variable with law $\rho_t \in \cP(\cZ)$. Equation \eqref{eq:wass_gradflow_general} is thus associated with a stochastic differential equation (SDE) on this random variable:
\begin{equation}\label{eq:mckean}
    \dif Z_t = \phi(Z_t, \rho_t) \dif t, \qquad  Z_0 \sim \rho_0, 
\end{equation}
for $\phi(Z_t, \rho_t) = -\nabla \tfvar[F]{\rho}(\rho_t)(Z_t)$. This is a particular (diffusion-less) case of a McKean-Vlasov process \citep{kac1956foundations, mckean1966class}. Eq.~\eqref{eq:mckean} can be interpreted as the trajectory of a single \textit{particle}, with initial position drawn from $\rho_0$, moving according to a potential function that captures its intrinsic dynamics and interactions with other particles, all driven by $F$. Two key observations are in order: (i) this \textit{particle view} of the gradient flow lends itself to computational schemes that operate on finitely many samples, and (ii) the process in eq.~\eqref{eq:mckean} is defined on $\cZ$ (a finite-dimensional space here) rather than on the infinite-dimensional $\cP(\cZ)$, making it amenable to computation.

Numerical approaches to solve SDEs like \eqref{eq:mckean} require time discretization and finite-sample approximation. A simple way to achieve the former is with a forward Euler scheme:
\begin{equation}
    Z_{t+1} = Z_t - \gamma \nabla F_{\alpha_t}(Z_t),   \qquad Z_0 \sim \alpha_0 .
\end{equation}
Computationally, this scheme is approximated by a system of particles that evolve simultaneously. Starting from $\rho_0 \approx \sum_{i=1}^N p_i\delta_{z^{(i)}_{0}}$, each particle $z^{(i)}$ is evolved according to the Euler scheme above, resulting in a system of $n$ updates:
\begin{align}\label{eq:particle_update}
    z_{t+1}^{(i)} = z_{t}^{(i)} - \gamma \nabla_{z^{(i)}_t} F\bigl(z_{t}^{(i)}\bigr)  \qquad \forall i=1,\dots,n.
\end{align}
Then, $\rho_t$ is approximated as $\rho_{N,t} = \sum_{i=1}^N p_i\delta_{z_t^{(i)}}$. For well-behaved functionals, this flow inherits all convexity and stability properties of the exact one, and $\rho_{N,t}(x)$ converges to $\rho_t(x)$ \citep{carrillo2019aggregation}. 

\section{Encoding Dataset Objectives as Functionals on Measures}\label{sec:functionals}
Most of the applications motivating this work involve objectives that refer to one or more `reference' datasets. In particular, we are interested in the functional $\cT_{\beta}(\rho)\triangleq \text{OTDD}(\rho, \beta)$, where $\beta$ is a (fixed) \textit{target} dataset distribution. Various applications can be modeled with this functional, such as dataset interpolation or sample generation for transfer learning (\sref{sec:transfer_mnist}). This objective can be further combined with some of the other functionals introduced in Section~\ref{sec:background} to `shape' the optimal distribution, balancing similarity to $\beta$ with some other (\eg size, regularity, or style) constraints imposed on it. This is often called \textit{controlled synthesis} in generative modeling \citep{wang2018high}. Next, we discuss specific examples of dataset objectives encoded using each of the three canonical functionals. 

\paragraph{Potential energy.}
Functionals of the form $\cV(\rho) = \int V(z) \dif \rho(z)$ can be used to enforce local (per-instance) constraints on the minimizer $\rho^*$. For example, one might be interested in constraining the norm of the features of this dataset by taking $V(z) = V(x,y) = \|x\|$, or more generally $V(z) = \|\A x - b\|$ for some $\A \in \R^{d\times d}, b\in \R^d$. Alternatively, one might be interested in enforcing such constraints only on certain classes, or have different parameters for every class, \eg $V(z) = \|\A_{y}x - b_{y}\|$. Another possibility is to enforce linear separability in binary classification datasets through a margin-inducing potential such as $V(z) = \max\{0, y(x^\top w - b)\}$ (used in Fig.~\ref{fig:main_diagram}). Note that in all of these cases $V$ is convex, as required for convergence. 

\paragraph{Interaction energy.} The functional $\cW(\rho) = \iint W(z - z') \dif \rho(z) \dif \rho(z')$ can encode objectives that model interaction or aggregation between samples in the dataset. An illustrative example of this is the class-repulsion functional obtained by taking:
\vspace{-0.2cm}
\[  W(z-z') =  \begin{cases} \exp\{-\|x-x'\|\} &\text{ if } y\neq y' \\  0 &\text{otherwise}\end{cases} \medspace, \]
which encourages class separation by penalizing pairs $(z,z')$ that have different labels but similar features. We show an example of a flow using this functional in a simple 2D dataset in Appendix~\ref{sec:additional_gaussian}.

\paragraph{Internal energy.}
Compared to the previous three, the functionals $\cF(\rho) = \int f(\rho(z)) \dif z$ are less relevant for our purposes, since an explicit density $\rho(z)$ will rarely be available in closed form. Furthermore, care must be taken with these functionals as they yield diffusion terms (e.g., $\Delta \rho$ in the associated PDE \eqref{eq:wass_gradflow_f}) that, despite leading to theoretically well-behaved gradient flows, are challenging to solve numerically via particle methods \citep{carrillo2019aggregation}. A notable exception is the functional obtained by taking $f(t)=t\log t$, which corresponds to adding an entropy term to the objective (equivalently, a Brownian motion term in Equation~\eqref{eq:mckean}), and can be solved via particle methods, e.g., by using an Euler-Maruyama scheme. For more general functionals of this form, recent work circumvents this problem through stochastic \citep{huang2017error,liu2017random}, deterministic \citep{carrillo2017numerical}, and regularization-based schemes \citep{carrillo2019blob}. We leave their exploration for future work. 

\section{Practical Implementation}\label{sec:implementation}
The particle scheme \eqref{eq:particle_update} provides a template for a gradient-based approach to solve the dataset optimization problem. In this section we discuss how the per-particle gradients are to be computed efficiently, provide practical implementation considerations for the functionals considered in this work, and discuss general computational aspects.

\subsection{Dataset Distance}\label{sec:practical_otdd}

\begin{figure*}
    \centering
    \includegraphics[width=0.3\linewidth, trim={0.3cm 0.3cm 0.3cm 0.3cm},clip]{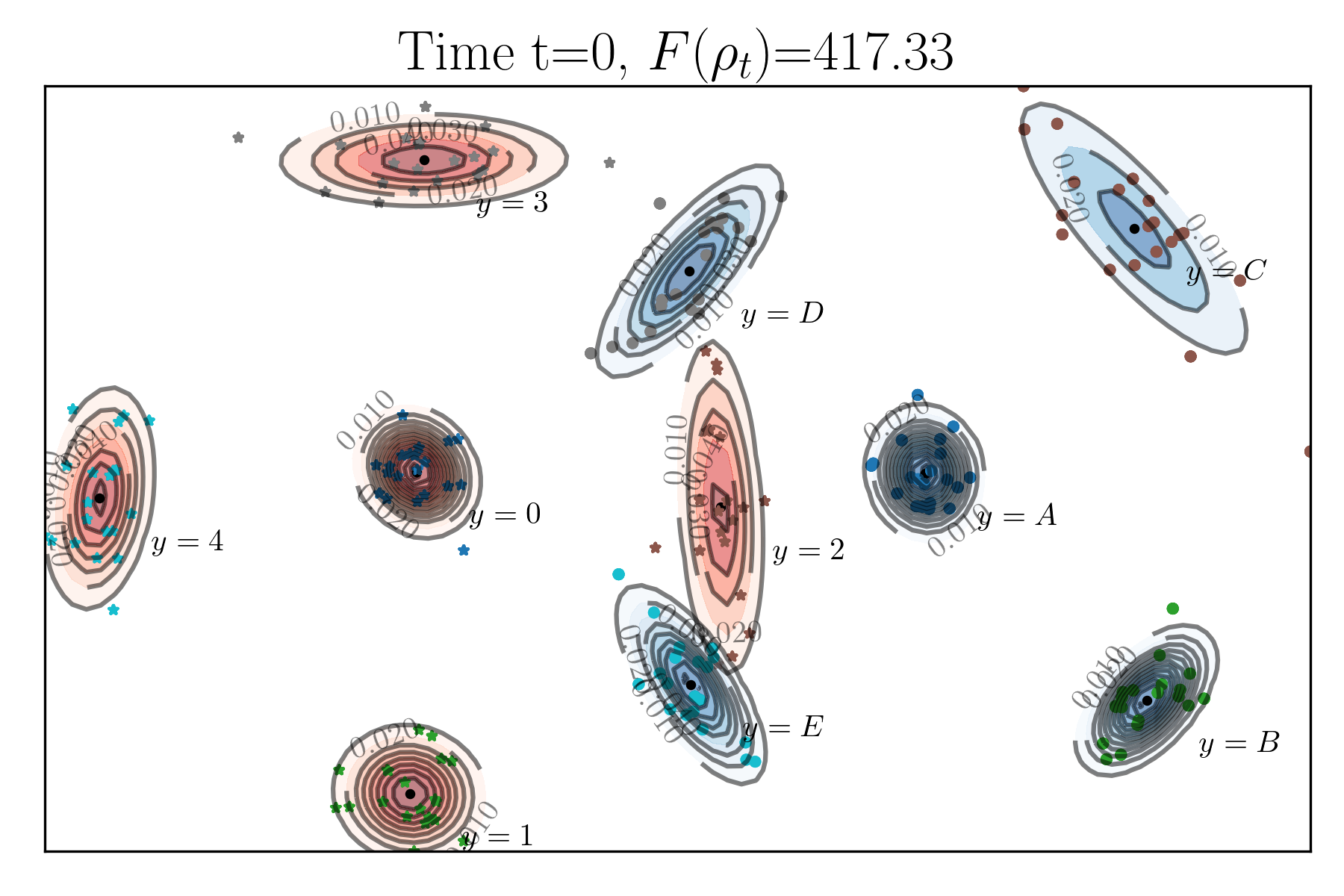}%
    \includegraphics[width=0.3\linewidth, trim={0.3cm 0.3cm 0.3cm 0.3cm},clip]{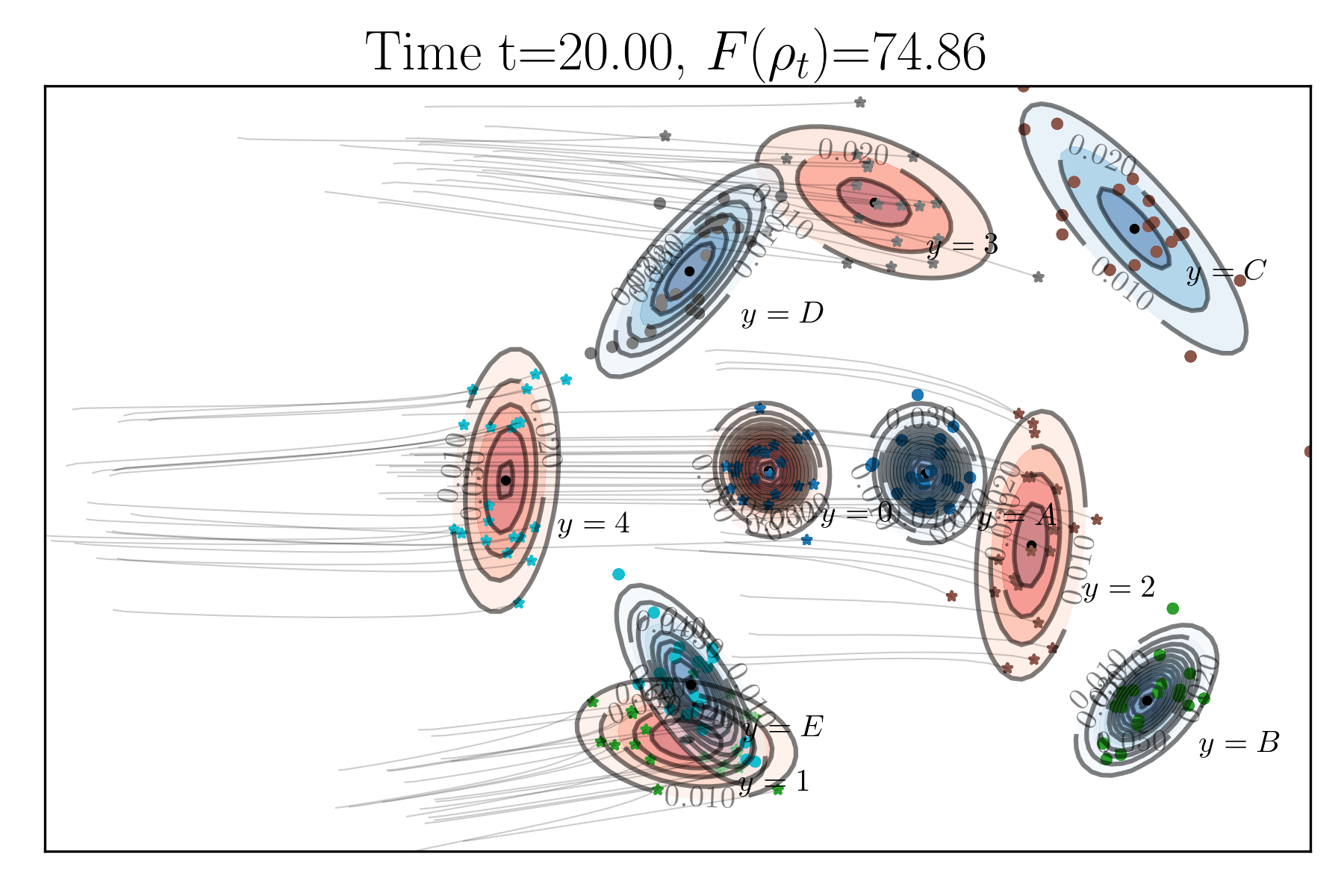}%
    \includegraphics[width=0.3\linewidth, trim={0.3cm 0.3cm 0.3cm 0.3cm},clip]{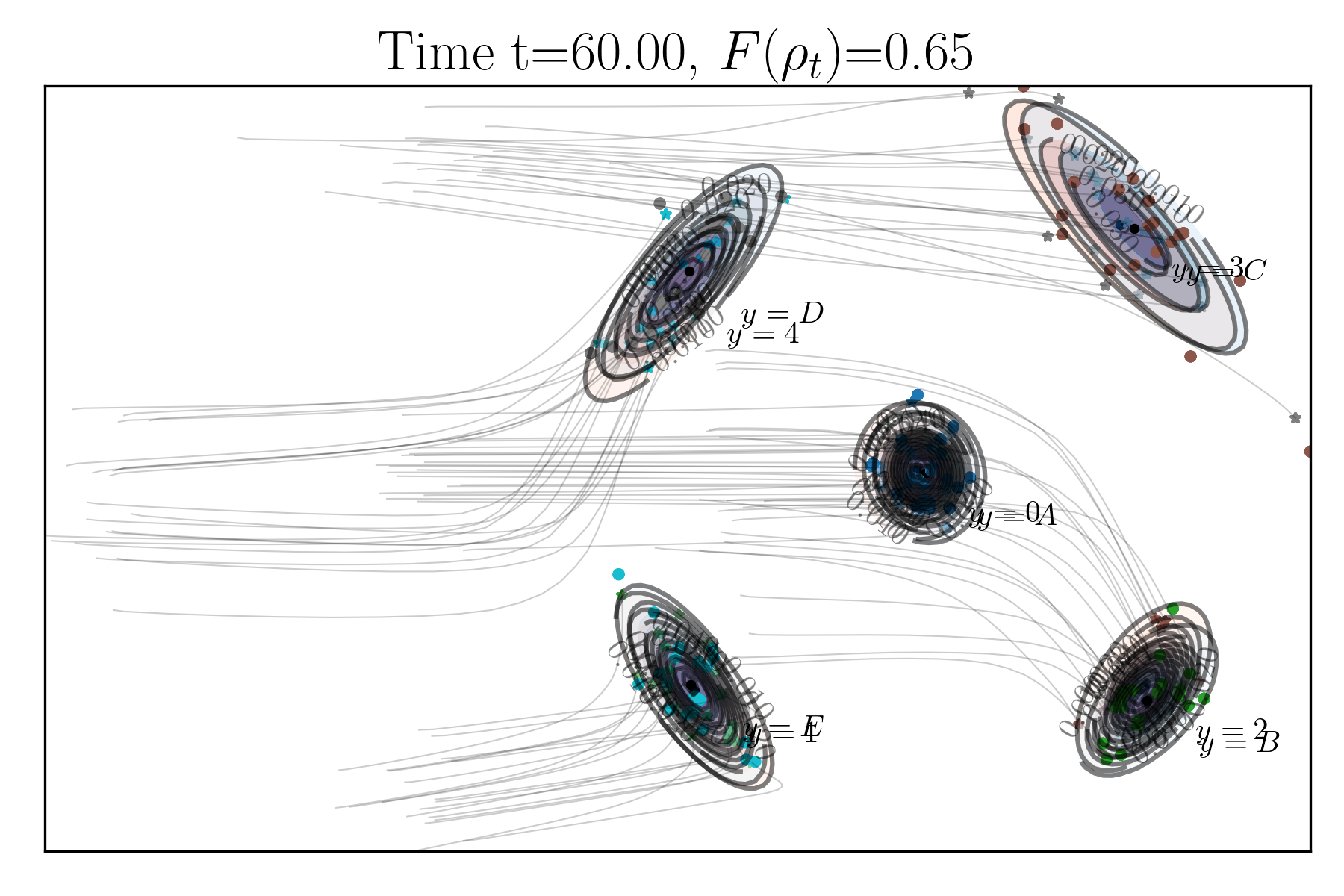}
    \includegraphics[width=0.3\linewidth, trim={0.3cm 0.3cm 0.3cm 0.3cm},clip]{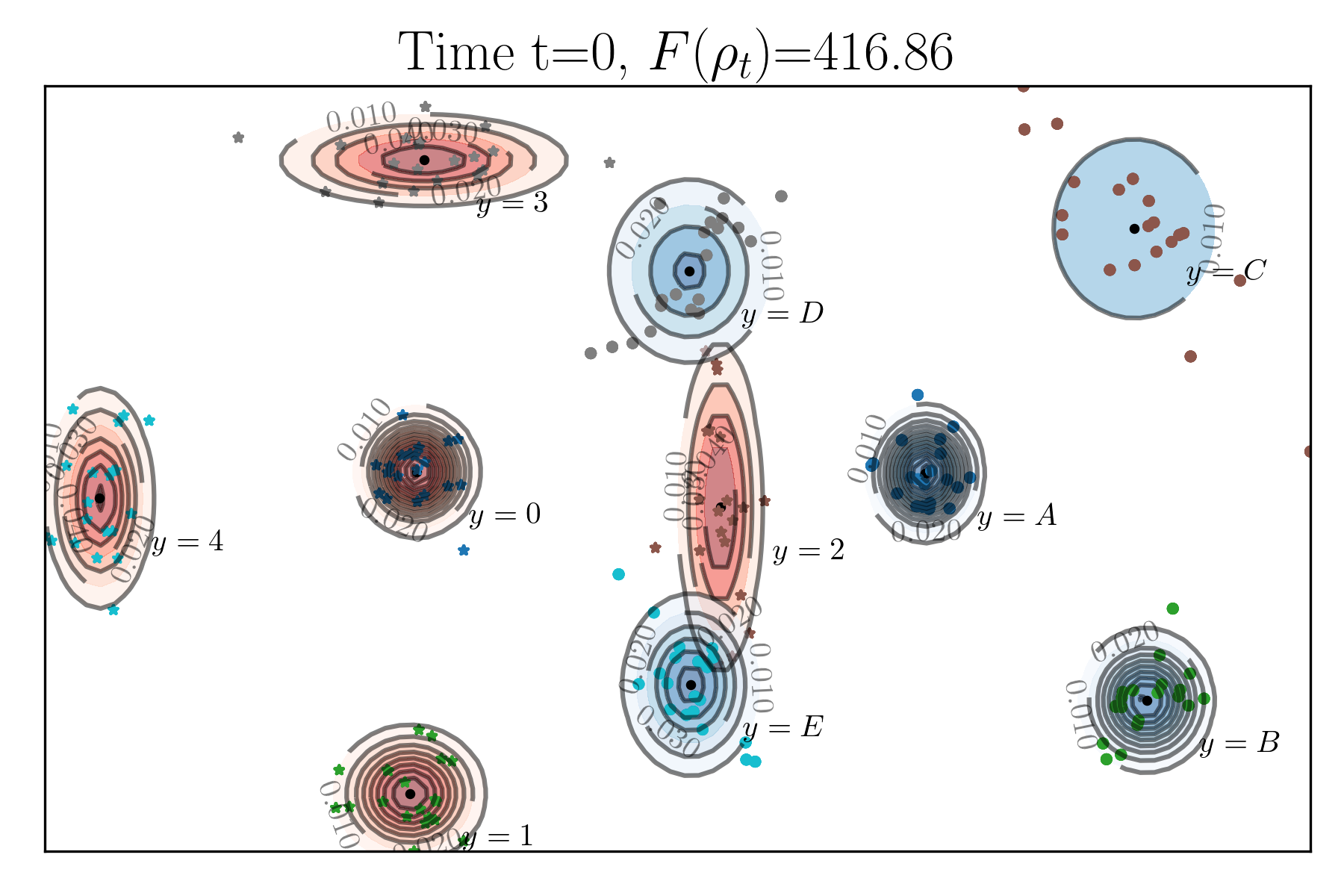}%
    \includegraphics[width=0.3\linewidth, trim={0.3cm 0.3cm 0.3cm 0.3cm},clip]{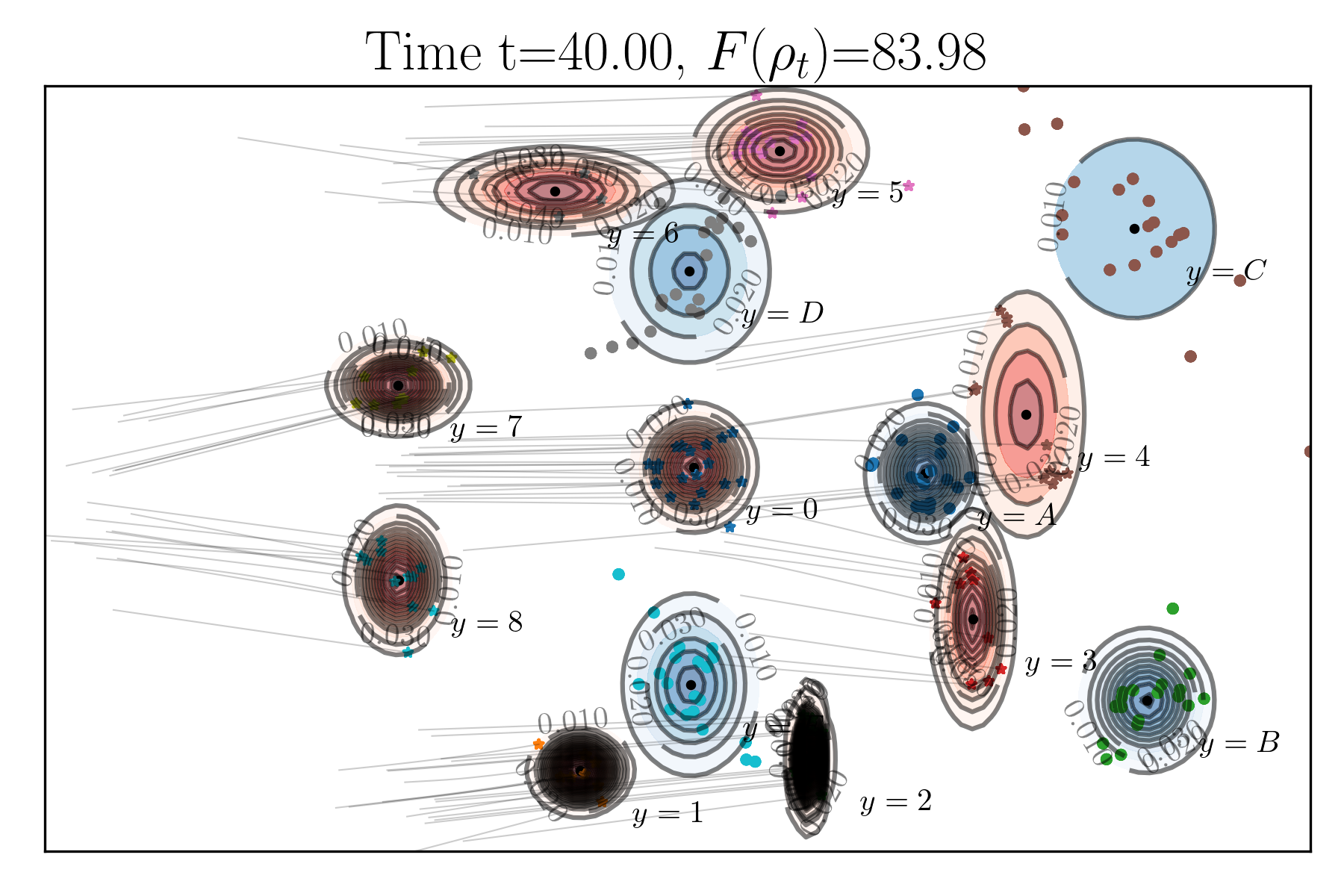}%
    \includegraphics[width=0.3\linewidth, trim={0.3cm 0.3cm 0.3cm 0.3cm},clip]{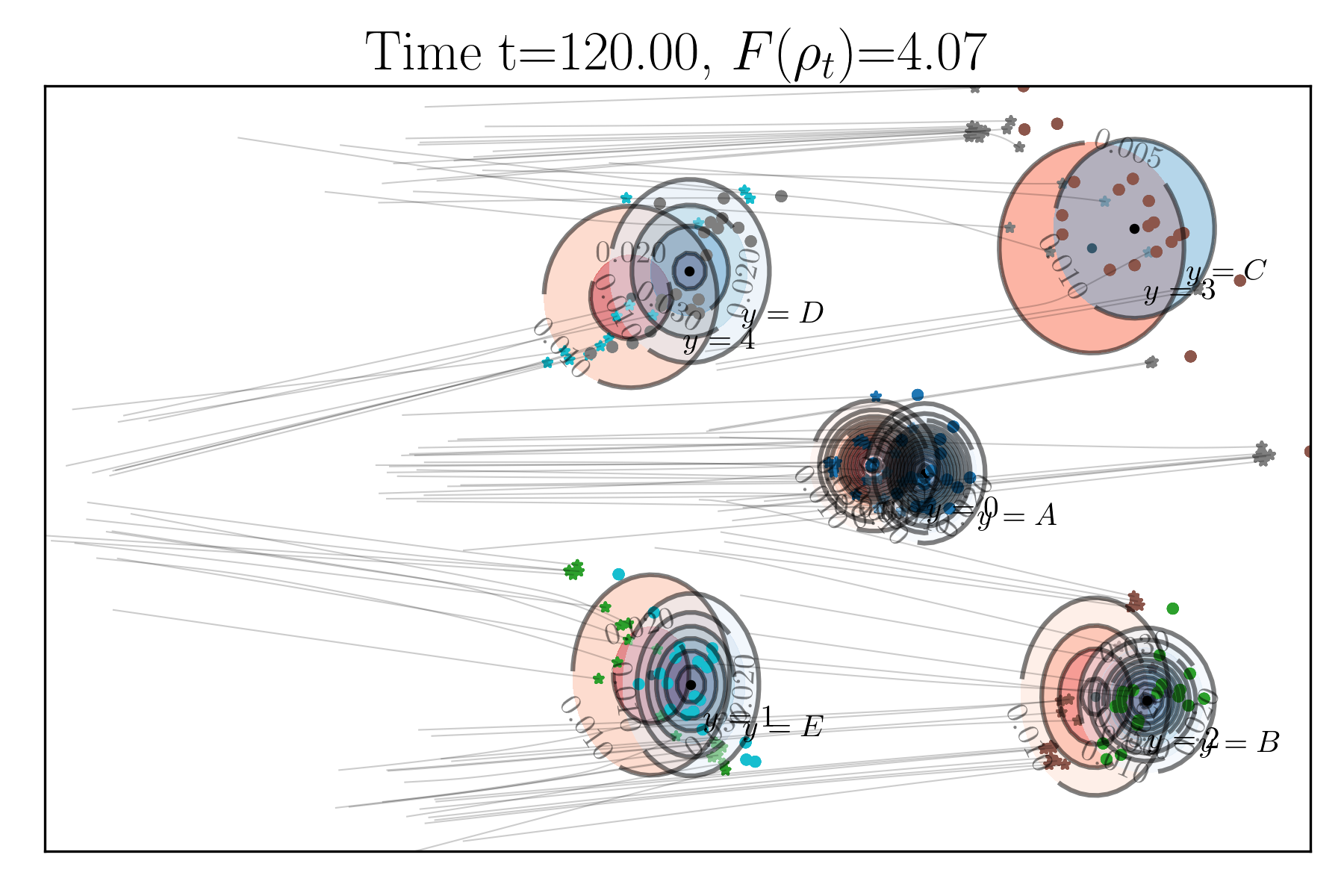}
    \vspace{-4mm}
    \caption{\textbf{OTDD dynamics comparison}: gradient flows driven by $\cT_{\beta}(\rho) = \text{OTDD}(\mathrm{D}_{\rho}, \mathrm{D}_{\beta})$ starting from dataset $\rho_0$ (red) advecting towards $\beta$ (blue), using: \textsc{sgd}+\texttt{jd-fl} (top) and  \textsc{sgd}+\texttt{jd-vl} dynamics (bottom), the latter allowing for variable class assignments. }
    \label{fig:gaussian_flows}
\end{figure*}

The main challenge in computing $\nabla_{z^{(i)}} \text{OTDD}(\mathrm{D}_{\rho_t}, \mathrm{D}_{\beta})$ arises ---unsurprisingly--- from the discrete nature of the labels $y$. Recall that each point (particle) is a pair $(x,y)$, where $x \in \R^d$ and $y \in \{c_i,\dots,c_K\}$. The OTDD framework described in Section~\ref{sec:otdd} provides an alternative representation of this point as $(x,\upsilon_y)$ where $\upsilon_y \in \cP(\cX)$ is a distribution over $\cX$ associated with the label $y$.

For small datasets the Gaussian approximation for OTDD (\sref{sec:otdd}) is not needed, so the label-to-label distances $\text{W}(\upsilon_y, \upsilon_{y'})$ are computed non-parametrically on-the-fly. In this case, only the features need to be updated, so it suffices to compute $\nabla_{x^{(i)}} \text{OTDD}(\mathrm{D}_{\rho_t},\!\mathrm{D}_{\beta})$. By using a differentiable solver for the inner and outer OT problems, we compute these gradients through automatic differentiation.

If, instead, we use the Gaussian approximation in the OTDD, there is now a parametric representation of the distributions $\upsilon_y = \cN(\mu_y, \Sigma_y)$ that must be explicitly updated. A simple way to implement this is to backpropagate gradients into $x^{i}$,\footnote{Here we drop parentheses in superscripts for convenience.} take a gradient step on those, and then re-compute means and covariances (and, as a result, the distributional representation of labels $\upsilon_y$). Formally,
\vspace{-0.15cm}
\begin{align*}
    x_{t+1}^i &= x_{t}^i - \tau \nabla_{x^i}F(\rho) \quad &i \in \{1, \dots, n\}\\ 
	\mu_{t+1}^j &= \text{mean}(\{x_{t+1}^i\}_{i \medspace:\medspace y^i =j}) \quad &j \in \{1, \dots, k\}  \\
	\Sigma_{t+1}^j &=  \text{cov}(\{x_{t+1}^i\}_{i \medspace:\medspace y^i =j}) \quad &j \in \{1, \dots, k\}   \\
	\upsilon_{t+1}^t &= \mathcal{N}(\mu_{t+1}^j , \Sigma_{t+1}^j ) \quad &j \in \{1, \dots, k\} \vspace{-0.2cm}
\end{align*}
Note that here $\mu^j, \Sigma^j$ evolve in response to the dynamics of the their associated empirical sample $\{x^i\}_i$, \textit{but not directly because of a gradient step}. Thus, we refer to this type of flow as \textbf{feature-driven} (\texttt{fd}) dynamics. The main drawback of this approach is that the label assignments are fixed, i.e., if $\rho_0$ has $k$ classes so will every $\rho_t$, and every $x_t^i$ maintains its label throughout. This might be acceptable if $\mathrm{D}_{\rho_0}$ and $\mathrm{D}_{\beta}$ have the same number of classes. 

In order to relax the approach above to allow direct updates on the label distributions $\upsilon_y$, one can instead apply independent updates on features and labels, while keeping the label assignments fixed throughout. This \textbf{joint-driven fixed-label} (\texttt{jd-fl}) scheme corresponds to:
\vspace{-0.15cm}
\begin{align*}
    x_{t+1}^i &= x_{t}^i - \tau \nabla_{x^i}F(\rho) \quad &i \in \{1, \dots, n\} \\ 
	\mu_{t+1}^j &= 	\mu_t^j  -\tau \nabla_{\mu^j}F(\rho)\quad &j \in \{1, \dots, k\}  \\
	\Sigma_{t+1}^j &= 	\Sigma_t^j  -\tau \nabla_{\Sigma^j}F(\rho)\quad &j \in \{1, \dots, k\} \\
	\upsilon_{t+1}^j &= \mathcal{N}(\mu_{t+1}^j , \Sigma_{t+1}^j ) \quad &j \in \{1, \dots, k\} \vspace{-0.2cm}
\end{align*}
Relaxing the constraint that label assignments are fixed through the flow requires evolving the distribution associated to each point individually. To do so, at time $t=0$ we decouple the distributions $\upsilon_j$, yielding now one per particle instead of one per class. We then evolve each particle (both its feature and label-distribution) independently: 
\begingroup
\setlength{\abovedisplayskip}{1pt}%
\setlength{\belowdisplayskip}{1pt}%
\setlength{\abovedisplayshortskip}{1pt}%
\setlength{\belowdisplayshortskip}{1pt}%
\begin{align*}
    x_{t+1}^i &= x_{t}^i - \tau \nabla_{x^i}F(\rho) \quad &i \in \{1, \dots, n\}\\ 
	\mu_{t+1}^i &= 	\mu_t^i  -\tau \nabla_{\mu^i}F(\rho)\quad &i \in \{1, \dots, n\}  \\
	\Sigma_{t+1}^i &= 	\Sigma_t^i  -\tau \nabla_{\Sigma^i}F(\rho)\quad &i \in \{1, \dots, n\} \\
	\upsilon_{t+1}^i &= \mathcal{N}(\mu_{t+1}^i , \Sigma_{t+1}^i ) \quad &i \in \{1, \dots, n\}
\end{align*}
\endgroup
However, recovering discrete labels from the (now decoupled) $\upsilon_t^i$ requires aggregating particles based on these. We do so by clustering the pairs $(\mu_i, \Sigma_i)$. Crucially, we use non-parametric clustering methods that \textit{do not require} the number of clusters be specified, so that they can freely change throughout the flow. We refer to this third approach as \textbf{joint-driven variable-label} (\texttt{jd-vl}) dynamics.

Figure~\ref{fig:gaussian_flows} shows a comparison of flows driven by two different dynamics on a simple 2D dataset. Additional qualitative comparison can be found in Appendix~\ref{sec:additional_gaussian}.

\subsection{Gradients of Energy Functionals}\label{sec:gradients}
Recall that the first variation of potential functionals $\cV$ is given by $\fvar[\cV]{\rho} = V:\!\cZ\!\rightarrow\!\R$, so it suffices to compute gradients of this scalar-valued function, which can be done using automatic differentiation. For interaction functionals, $\fvar[\cW]{\rho}(z) = (W \ast \rho) (z) = \int W(z-z') \dif \rho(z')$, which albeit scalar-valued, now involves an integral over $\cZ$. We approximate it as an empirical expectation over the particles. Finally, for internal energy functionals $\fvar[\cF]{\rho} = f'(\rho)$, so if the density $\rho(\cdot)$ is available and can be computed with automatic differentiation, so can the gradient of $f' \circ \rho: \cZ \rightarrow \R$.

\subsection{Flowing Unlabeled Data}\label{sec:unlabeled_flows}
In many applications, such as in semi-supervised or unsupervised transfer learning, the data available to initialize the flow might be partially or completely unlabeled. However, so far we have assumed all particles are labeled, as required by the OTDD. Using label-dependent flows even for semi- or unsupervised settings might be desirable to explicitly model class-conditional geometric structure. Thus, we propose two approaches to adapt our flows to settings with limited or no labeled data. 

\paragraph{Parametric Flows for Partially Labeled Data.}
 If some labels are available, say, for the data samples indexed by $\mathcal{L}$, but not for those indexed by $\mathcal{U}$, one can first run a flow starting from $\{x^{i}, y^{i}\}_{i\in \mathcal{L}}$, to obtain a collection of final particles $\{x^{i}_T, y^{i}_T\}_{i\in \mathcal{L}}$. Next, the action of the flow $h_{\text{flow}}:x_0 \mapsto x_T$ can be parametrized (e.g., with a neural network) and learnt by fitting $\{x^{i}\}_{i\in \mathcal{L}}$ to $\{x^{i}_T\}_{i\in \mathcal{L}}$. This learnt mapping $\hat{h}$ can then be used to flow the unlabeled samples, i.e., as $\hat{x}^{i}_T = \hat{h}(x^{i}), i\in \mathcal{U}$. We evaluate this approach for semi-supervised transfer learning in Section~\ref{sec:transfer_mnist}.
 
\paragraph{Pseudo-Labels for Fully Unlabeled Data.} If no labels are available at all, we propose to generate \textit{pseudo-labels} for the flow. This can be done, for example, by clustering the initial data and using the identities of the clusters as imputed labels. We evaluate this approach in Section~\ref{sec:repurposing}.

\subsection{Miscellaneous Implementation Aspects}\label{sec:general_aspects}
Traditional gradient flow implementations rely on constant or scheduled step-size schemes, i.e., vanilla gradient descent. In order to accelerate the flows and account for potential violations in the convexity assumptions of $F$, here we also consider various adaptive step-size methods like \textsc{sgd} with momentum and \textsc{adam}, which have been shown to aid in escaping local minima in deep learning.

In our experiments we will study the behavior of classifiers as we flow data between datasets, often of different dimensionality and with different classes. To deal with the former in image classification applications, we up- or down-sample as necessary to obtain equally-sized images --- that is, to put them on the same $\cX$ space. A more general approach to deal with mismatch in dimensionality is discussed in \citep{alvarez-melis2020geometric}.

We deal with mismatch in classes by means of correspondence. For instance, in order to be able to use a classifier trained on a dataset with $k$ classes on a flow originating in one with $k'$ classes even when using fixed-label dynamics, we propose to take advantage of the label-to-label distances computed by the OTDD under the hood (see Section~\ref{sec:otdd}) to obtain correspondences between classes. Specifically, we solve an OT problem between the two class probability histograms using these distances, and append the $k'\times k$ optimal coupling $\pi^*$ as a final linear layer to the classifier. Appendix~\ref{sec:experiment_details} provides further computational details.

\section{Experiments}
We first evaluate our approach for imposing constraints on low-dimensional synthetic datasets (Section~\ref{sec:shaping_experiments}) and then on two settings (Sections~\ref{sec:transfer_mnist} \& \ref{sec:repurposing}) involving transfer learning with benchmark image classification datasets. 

\subsection{Imposing Dataset Constraints via Flows}\label{sec:shaping_experiments}
For a synthetic 3D setting, we experiment with flows driven by functionals that combine distance to a reference dataset (here, a swiss roll) and various other functional objectives such as those described in Section~\ref{sec:functionals}. The results (Figure~\ref{fig:main_diagram}, bottom row and Figure~\ref{fig:additional_shaping} in the Appendix) show that the flows indeed converge to solutions that are geometrically similar to the reference dataset, but are `shaped' by the additional constraints imposed by the functional objectives. 

\subsection{Transfer Learning via Flows}\label{sec:transfer_mnist}
\begin{figure}[t]
	\centering
	\begin{subfigure}[t]{0.21\linewidth}\vspace{0pt}
		\centering
		$\displaystyle \xrightarrow{\quad \text{time} \quad}$
        \includegraphics[width=\linewidth, trim={1cm 0 0 0},clip]{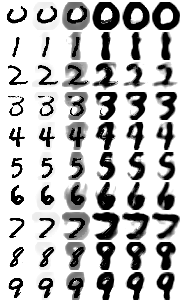}%
		\caption{$z_t$:\textsc{m}$\shortrightarrow$\textsc{u}}\label{fig:mu_traj}		
	\end{subfigure}%
	~
	\begin{subfigure}[t]{0.21\linewidth}\vspace{0pt}
		\centering
		$\displaystyle \xrightarrow{\quad \text{time} \quad}$
        \includegraphics[width=\linewidth, trim={1cm 0 0 0},clip]{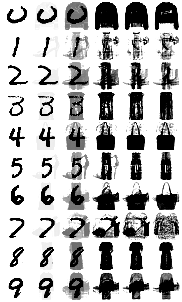}%
		\caption{$z_t$:\textsc{m}$\shortrightarrow$\textsc{f}}\label{fig:mf_traj}				
	\end{subfigure}%
	~
	\begin{subfigure}[t]{0.21\linewidth}\vspace{0pt}
		\centering
		$\displaystyle \xrightarrow{\quad \text{time} \quad}$
        \includegraphics[width=\linewidth, trim={1cm 0 0 0},clip]{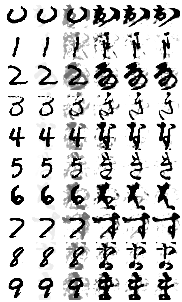}%
		\caption{$z_t$:\textsc{m}$\shortrightarrow$\textsc{k}}\label{fig:mk_traj}				
	\end{subfigure}%
	~
	\begin{subfigure}[t]{0.21\linewidth}\vspace{0pt}
		\centering
		$\displaystyle \xrightarrow{\quad \text{time} \quad}$
        \includegraphics[width=\linewidth, trim={1cm 0 0 0},clip]{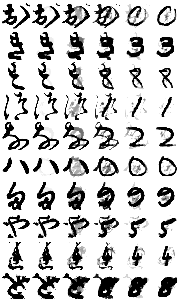}%
		\caption{$z_t$:\textsc{k}$\shortrightarrow$\textsc{m}}\label{fig:km_traj}				
	\end{subfigure}
 	~
	\begin{subfigure}[t]{\linewidth}\vspace{0pt}
		\centering
        \includegraphics[width=0.9\linewidth, trim={0.7cm 0.75cm 0.74cm 0.6cm}, clip]{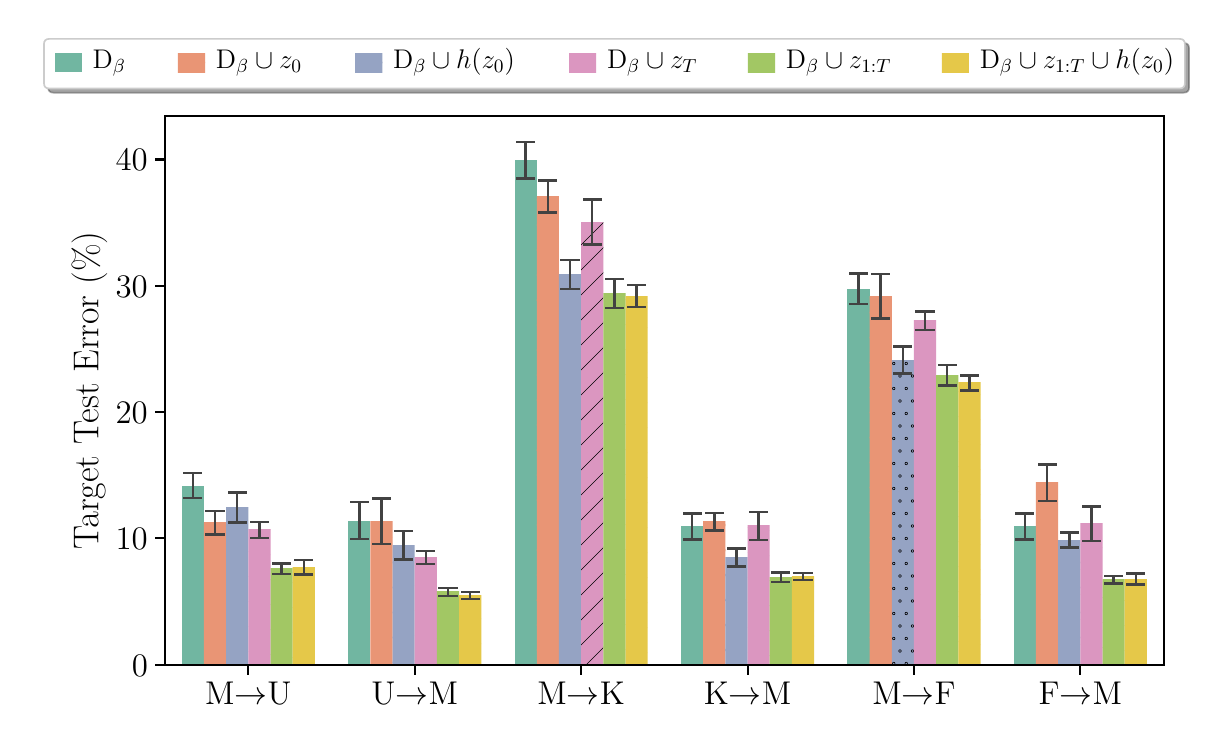}
		\caption{Transfer learning with extra source (dotted) / flow (dash) data}\label{fig:mnist_adapt}
	\end{subfigure}
    \vspace*{-4mm}
    \caption{\textbf{Transfer learning flows}: flows between *\textsc{nist} datasets using 2K particles (see \sref{sec:transfer_mnist} for dataset/legend key).}
    \vspace{-0.5cm}
    \label{fig:mnist_flows}
\end{figure}

Next, we investigate the use of OTDD gradient flows for transfer learning. Specifically, starting from a source domain dataset $\mathrm{D}_{\alpha}$, we transform it using a flow driven by the functional $\cT_{\beta}(\rho) = \text{OTDD}(\mathrm{D}_{\rho}, \mathrm{D}_{\beta})$ towards a target dataset of interest $\mathrm{D}_{\beta}$. 
We consider four classification datasets: \mnist \citep{lecun2010mnist}, \usps, \textsc{fashion-mnist} \citep{xiao2017fashion-mnist} and \textsc{kmnist} \citep{clanuwat2018deep}, denoted here as \textsc{m,u,f,k}. Figures~\ref{fig:mu_traj}-\subref{fig:km_traj} show particle trajectories between these.

Since in most transfer learning settings the number of samples from the target domain is limited, we investigate if the\;---highly non-linear---\;action of a flow can be learnt from samples, so as to `flow' additional unlabeled samples, as proposed in Section~\ref{sec:unlabeled_flows}. For this purpose, we parametrize $h_{\text{flow}}\!:x_0 \mapsto x_T$ as a neural net, and fit it using the initial and final states of the particles ($\{x_0^{i}\}_{i=1}^n$ and $\{x_T^{i}\}_{i=1}^n$). Figure~\ref{fig:mnist_mapped} (Appendix~\ref{sec:additional_nist}) shows that particles mapped with the learnt $h$ reasonably approximate those properly derived from the flow.

For evaluation, we emulate a $\mathsf{k}$-shot learning task, i.e., only $\mathsf{k}$ samples per class from the target domain $\mathrm{D}_{\beta}$ are available, in addition to 2000 from the source domain. With this data, we flow the source samples to the target domain, so that we have at our disposal: the particles in their initial ($z_0$) and final ($z_T$) states, the full trajectories $z_{1:T}$, and additional mapped examples $h(z_0)$ using a neural-net push-forward map $h$ as described above.

\begin{figure}[t]
    \centering
    \includegraphics[height=2.5cm, trim={0.25cm 0.98cm 3cm 0.25cm}, clip]{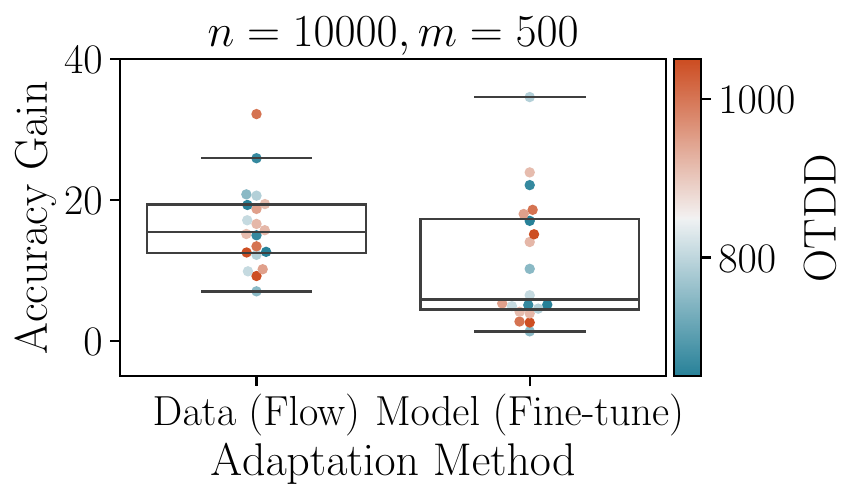}%
    \includegraphics[height=2.5cm, trim={1cm 0.98cm 0cm 0.25cm}, clip]{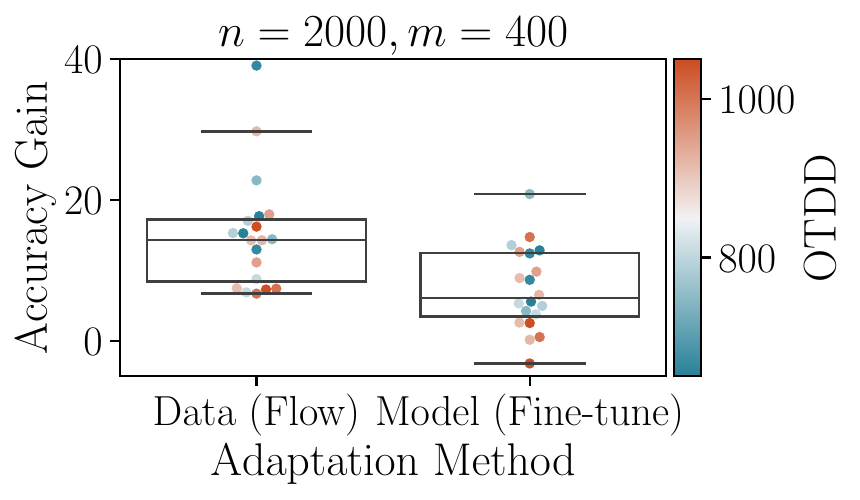}%
    \vspace{-2mm}
    \caption{\textbf{Data vs. model adaptation}: both regimes use the same number of source ($n$) and target ($m$) labeled samples. Each point is a pair of source/target datasets, color-coded by their distance.}
    \label{fig:data_vs_model_adapt}
\end{figure}

We compare various adaptation settings: a no-adaptation baseline (i.e., using target data $\mathrm{D}_{\beta}$ only), and adapting from various combinations of the additional data sources. Details of the classifier architecture and training configuration are provided in Appendix \ref{sec:experiment_details}. We present results for the 100-shot setting here (Figure~\ref{fig:mnist_adapt}), and 5- and 10-shot in Appendix~\ref{sec:additional_nist}. In most cases all additional data is beneficial, but the full trajectories provide the most gains, indeed improving over using final-state particles only. In addition, using samples mapped via $h$ provides some improvement over the baseline, but less so than using `clean' particles from the flow, as expected. When compared equitably (using the same number of labeled samples) against a usual model-adaptation (fine-tuning) baseline, our data-adaptation approach yields overall better performance gain (Figure~\ref{fig:data_vs_model_adapt}).\looseness=-1 

\subsection{Model Re-purposing and Oracle Evaluation}\label{sec:repurposing}

Next, we take the application of flows for transfer learning one step further, foregoing model adaption completely. We now assume the pretrained classifier is frozen and accessible through queries only, and use dataset flows to `re-purpose' it to solve a different task. Concretely, we flow samples from the dataset we seek to classify towards that on which the model was trained, and then we query the model directly on this flowed data without further modifications. Besides providing a litmus test for no-fine-tuning transfer learning, this setting also provides a general strategy to quantitatively evaluate the `quality' of a dataset flow, i.e., by using the model's accuracy on the flowed data as an oracle measure of evolution of the flow.\looseness=-1 

\begin{figure}[t]
    \centering
    \includegraphics[width=0.9\linewidth, trim={0cm 0.2cm 0cm 0}, clip]{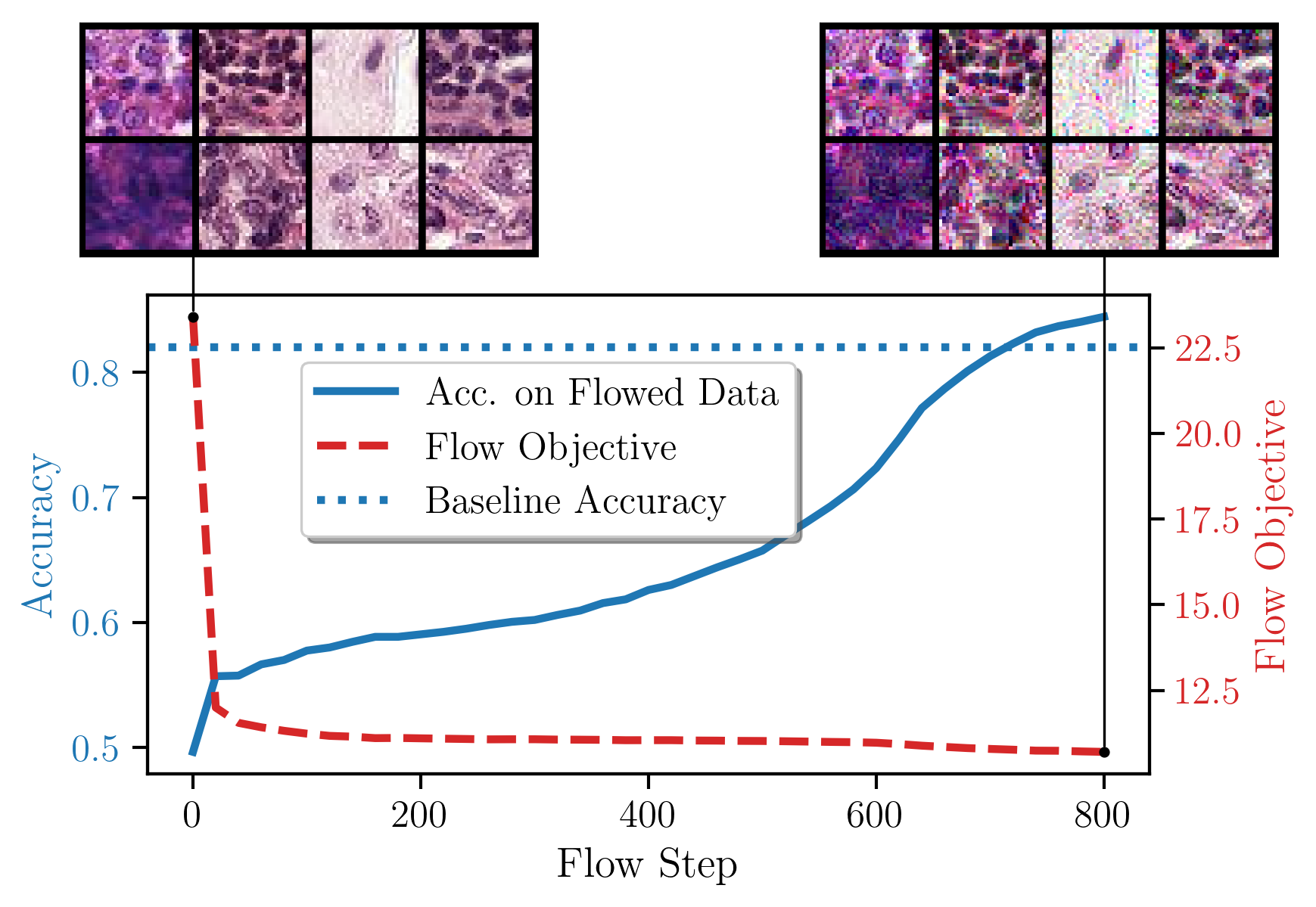}%
    \vspace{-4mm}
    \caption{\textbf{Model re-purposing}: we use a model trained on \cifar, without any modifications, to classify \textsc{camelyon} by flowing the latter to the former. After flowing the data, this model beats a strong baseline model trained only on \textsc{camelyon} data.
    }
    \vspace{-4mm}
    \label{fig:camelyon}
\end{figure}

\begin{figure*}[ht!]
    \centering
    \includegraphics[width=0.33\linewidth, trim={0.2cm 0 0cm 0}, clip]{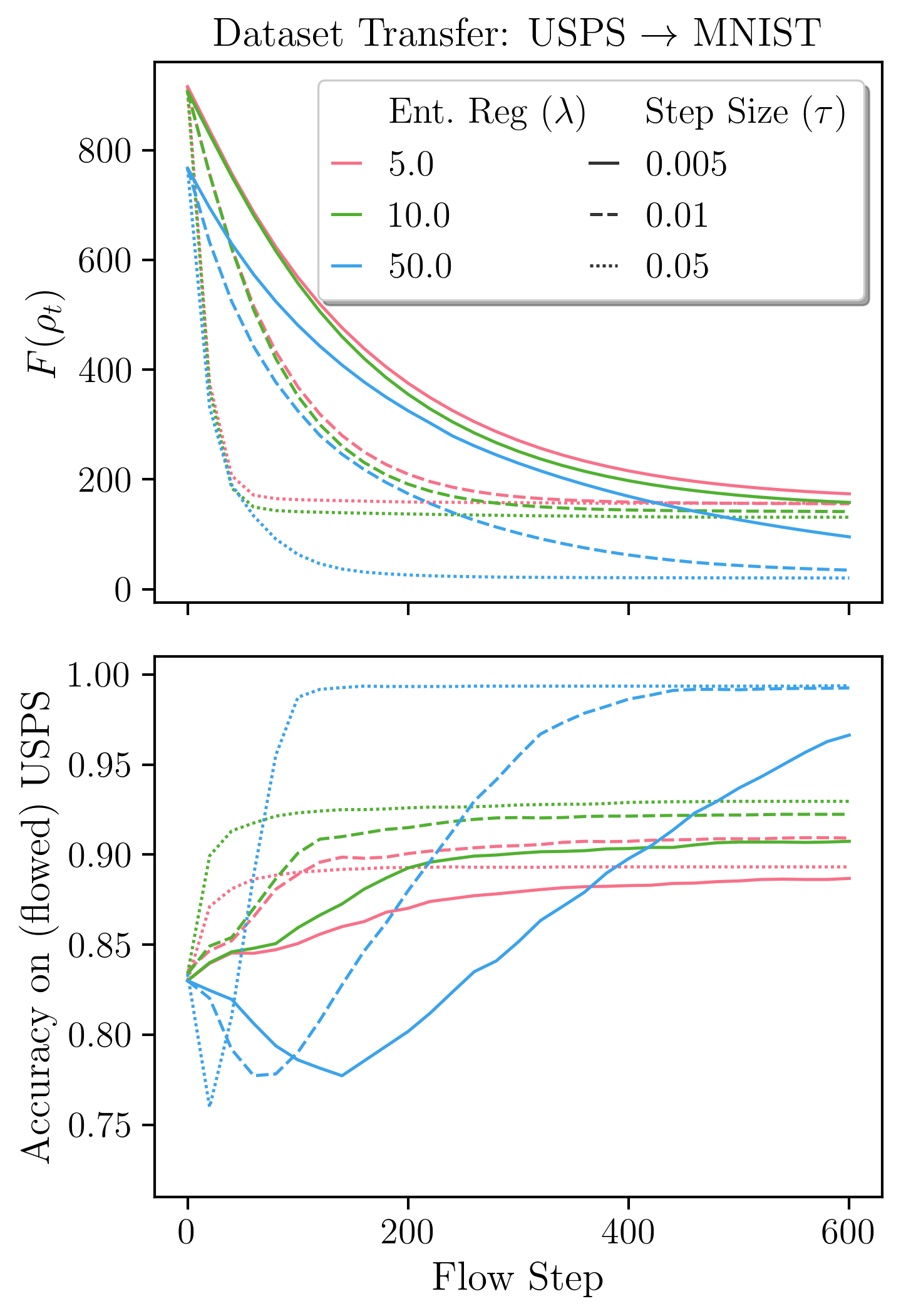}%
    \includegraphics[width=0.33\linewidth, trim={0.2cm 0 0cm 0}, clip]{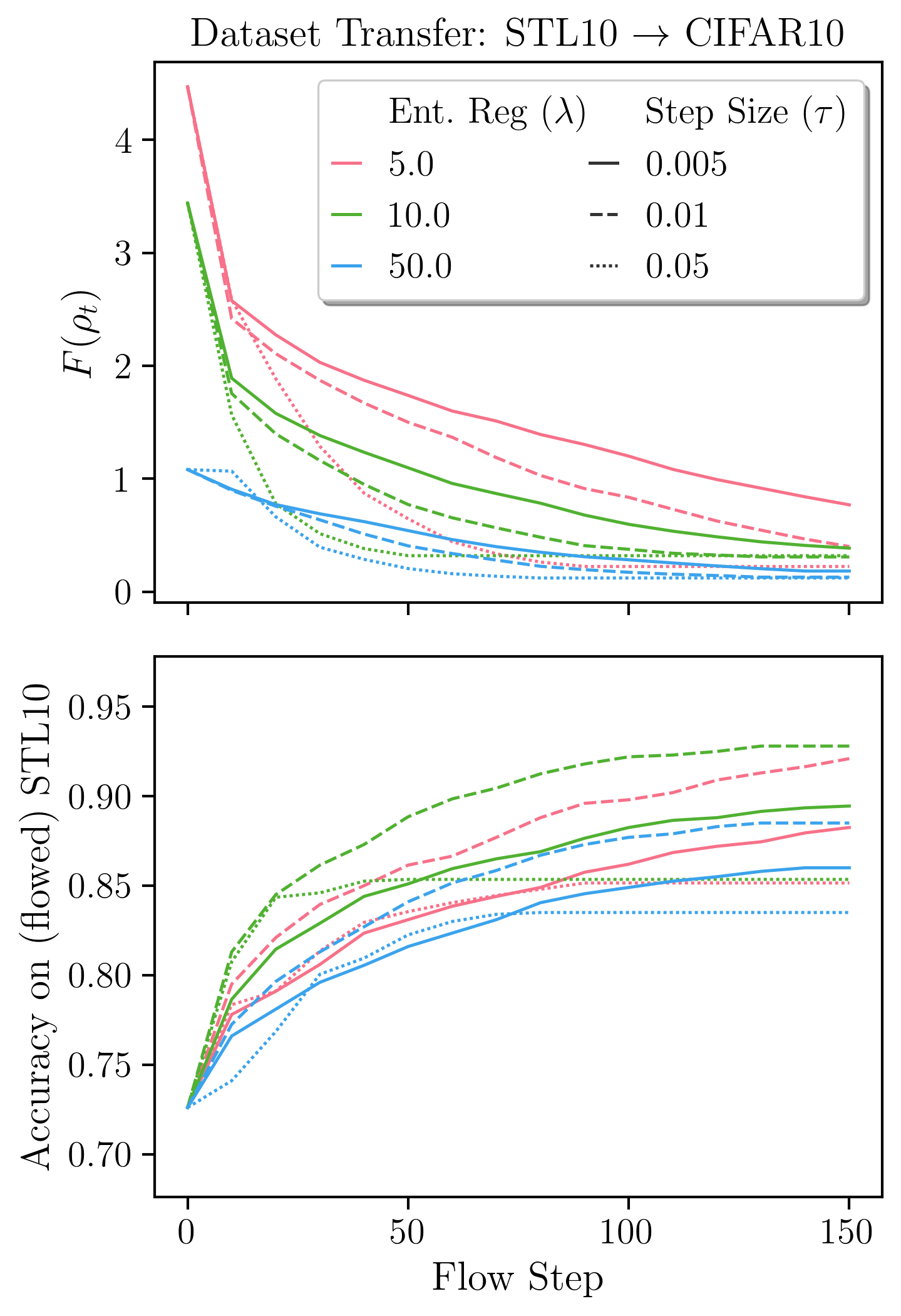}%
    \includegraphics[width=0.325\linewidth, trim={0.2cm 0 0cm 0}, clip]{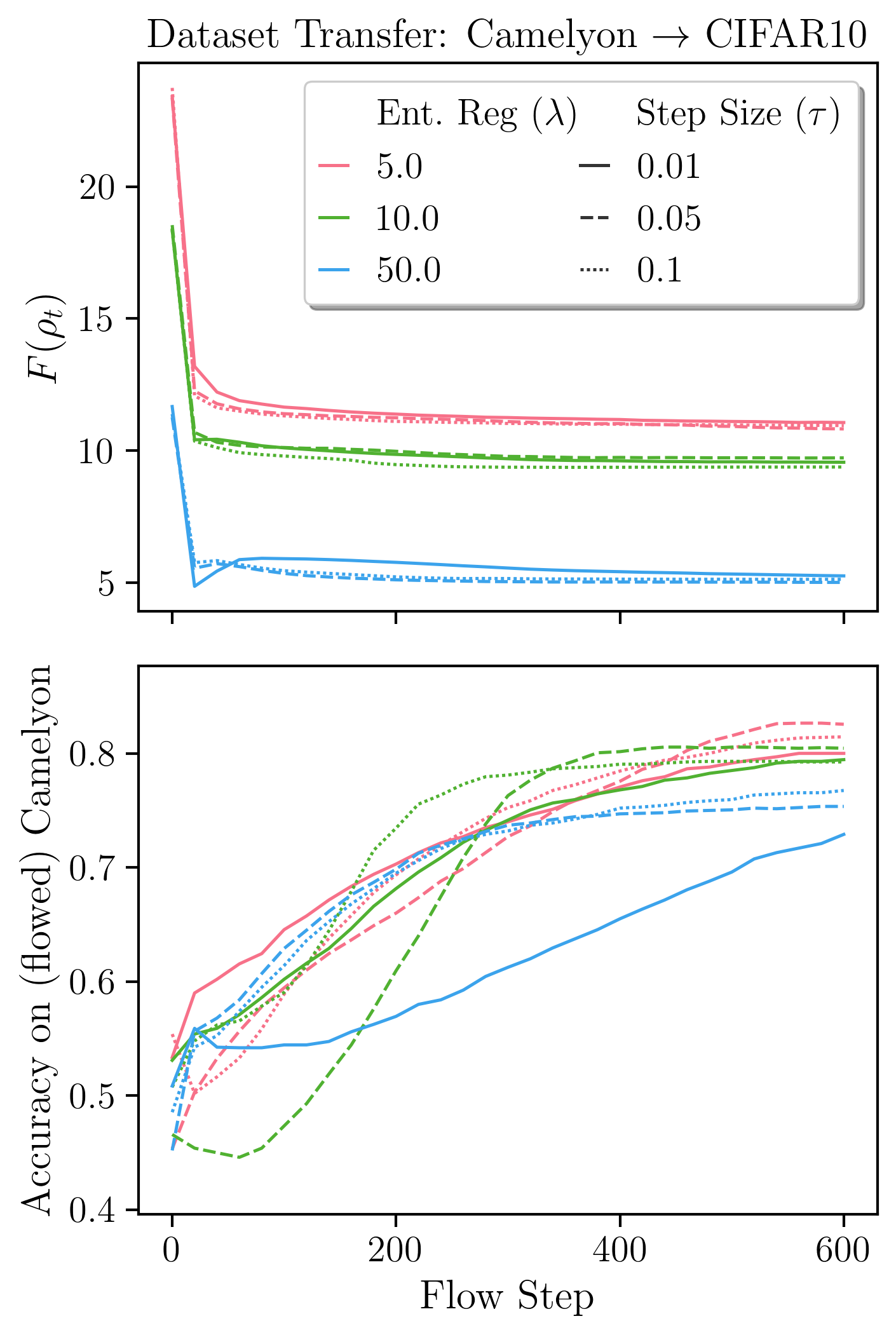}%
    \vspace*{-4mm}
    \caption{\textbf{Oracle evaluation of flows}: given a large classifier pretrained on a target domain, we use our method to `flow' data from a target domain back to the source domain. The plots show the value of the flow objective and target accuracy throughout the flow.}
    \label{fig:oracle_experiments}
\end{figure*}

In addition to the *NIST datasets, we use \textsc{cifar10}, \textsc{stl10} and the \textsc{camelyon} histopathology dataset \citep{litjens2018camelyon}.  The results (Figures~\ref{fig:oracle_experiments} \& \ref{fig:oracle_experiments_correlations}) show that the flows achieve high classification accuracy, even when the \textsc{oracle} is trained on a completely unrelated dataset (e.g., \textsc{camelyon}$\rightarrow$\textsc{cifar10}). Notably, the entropy regularization $\lambda$ used in OTDD has a much stronger effect on final accuracy than the step size. For the \textsc{camelyon}$\rightarrow$\textsc{cifar10} setting, Figure~\ref{fig:camelyon} shows a sample of flowed particles at different times. Although visible only through subtle artifacts in the images, flowing the data has a clear impact on the pretrained model's accuracy.

\begin{table}[t]
    \small
    \caption{\textbf{One model to classify them all}: we use the exact same model trained on \mnist without modifications to classify samples from various datasets, by flowing all datasets to \mnist, using either our vanilla flow with an OTDD functional objective (supervised) or its unsupervised variant that uses pseudo-labels (\sref{sec:unlabeled_flows}). 
    Shown here is the model's accuracy ($\%$) on 5K samples.
    }\label{tab:polimodality}   
    {\centering
    \begin{tabular}{>{\centering\arraybackslash}m{2cm} cccc}
        \toprule
        Dataset & \usps  & \kmnist & \textsc{fmnist} & \emnist \\
        \midrule
        Original (before flow) & 77.1  & 5.05 & 8.22  & 3.62$^\dagger$ \\[.2em]
        Flowed (supervised) & 99.1 & 99.2 & 99.1 & 36.7 \\
        Flowed (unsupervised) & 66.4 & 59.2 & 54.9 & 20.7 \\ 
        \bottomrule
    \end{tabular}
    }
    \hspace{1cm} $^\dagger$stacking a random $26\!\times\!10$ linear layer on top of the classifier.
    \vspace{-0.2cm}
\end{table}

But adapting datasets rather than models comes with another unexpected bonus: the exact same model can be used to classify \textit{multiple} different datasets \textit{simultaneously}, by flowing all of them to the domain on which the model was trained (stacking, if needed, a class-matching layer \sref{sec:general_aspects}). 

Here, we test this `model multi-purposing' setting using a classifier trained on \mnist (the source domain), without modifications, to classify the other \textsc{*nist} datasets (target domains). For this, we use the functional $\cT(\rho) = \textsc{otdd}(\rho, \rho_{\textsc{mnist}})$ as before in two flow modalities: supervised (using labeled target samples for the flow) and unsupervised (no labels used) with pseudo-labels obtained by k-means clustering as described in Section~\ref{sec:unlabeled_flows}. The former is an idealized\,---and unrealistic---\,scenario aimed at investigating the intrinsic limits of model-repurposing, and should be interpreted as an upper bound on transfer accuracy, since labeled data from the target domain is indeed used by the flows, although not by the classifier. The latter, however, is a \textit{bona-fide} method for unsupervised\footnote{However, the identities of the pseudo-labels and true labels must be aligned, manually or automatically (see Appendix \ref{sec:experiment_details}).} model-repurposing, applicable in practice. 

The results in Table~\ref{tab:polimodality} show that the frozen \textsc{mnist} classifier achieves almost perfect accuracy on most of the other datasets after transforming them with a supervised flow. As expected, when using an unsupervised flow the accuracy is lower, but still significantly higher than if no flow-based data adaptation is performed, except for \usps, which the \mnist model can already classify well without adaptation.

\section{Discussion and Future Work}
We have shown how gradient flows in probability space can be used to tackle various types of problems in machine learning that involve transformation of labeled datasets. We believe this work lays out a path for a largely uncharted \textit{data-centric} paradigm in machine learning, orthogonal and complementary to the prevailing model-centric one. Exciting directions of research within this paradigm abound. For example, we are interested in understanding meta-learning and related paradigms from a data-centric perspective, and to investigate whether other aspects of the deep learning toolkit (beyond adaptive gradient schemes) might prove useful in the context of dataset optimization.\looseness=-1

\bibliography{references}
\bibliographystyle{icml2021}

\clearpage
\pagebreak

\appendix
\onecolumn
\thispagestyle{empty}

\section{Brief Discussion of Convergence and Guarantees}\label{sec:convergence}
Proving existence and uniqueness of solutions of gradient flows in Wasserstein space, or convergence of their discretized schemes, is challenging. But it can be achieved through different types of assumptions on the spaces, metrics, and functionals. Here, we will briefly discuss guarantees that depend on one of the simplest such assumptions: geodesic convexity. 

\begin{definition}
    Let $\cX$ be a geodesic metric space and 
    $F: \cX \rightarrow \R \cup \{+\infty\}$ a functional. We say that $F$ is ($\lambda$-)\textbf{geodesically convex} if it is ($\lambda$-)convex along geodesics in $\cX$, \ie for every pair of points $x_0, x_1 \in \cX$, there exists a constant-speed geodesic $\omega$ connecting $\omega(0) = x_0$ to $\omega(1) = x_1$ such that $t\mapsto F(\omega(t)), t\in [0,1]$ is ($\lambda$-)convex.
\end{definition}
Note that if $\cX$ is Euclidean, the definition above is simply $\lambda$-convexity. On the other hand, this concept is well defined for metric measure spaces like $\mathbb{W}_p(\Omega)$ too. In particular, for $\mathbb{W}_2(\Omega)$, all geodesics are displacement geodesics, so the condition above is also known as\textit{ displacement convexity}.

Thus, a functional $F: \mathbb{W}_p(\cX) \rightarrow \R \cup\{+\infty\}$ is $\lambda$-geodesically convex if and only if for every pair $\mu^1, \mu^2 \in \cP(\cX)$ there exists an optimal transport coupling $\pi\in \Pi(\mu^1, \mu^2)$ such that
\begin{equation}\label{eq:displacement_convex}
 F(\mu_t^{1\shortrightarrow2}) \leq (1-t)F(\mu^1)+tF(\mu^2) -\frac{\lambda}{2}t(1-t)\W_p^2(\mu^1,\mu^2) \quad \forall t\in [0,1]     
\end{equation}
where $\mu_t^{1\shortrightarrow2}\triangleq ((1-t)x + ty)_{\sharp}\pi$ is a geodesic in $\mathbb{W}_p(\cX)$ interpolating between $\mu^1$ and $\mu^2$. 

It can be shown that the functionals used in this work (i.e., those in Eq.~\eqref{eq:functionals}) are displacement convex under suitable conditions \citep{santambrogio2017euclidean}. Specifically, $\cV$ and $\cW$ are $\lambda$-displacement convex if the underlying potentials $V$ and $W$ are $\lambda$-convex. For the internal energy functional $\cF$, some technical assumptions on $f$ are needed, such as requiring that $f(0)=0$, $s\mapsto s^df(s^{-d})$ is convex and decreasing, and the underlying space is convex \citep[Thm 7.28]{santambrogio2017euclidean}. It is easy to see that simple functions, such as the entropy term discussed before, or power functions with exponent $q>1$, all satisfy this condition. 

The following result, one of the simplest in such family of guarantees, shows the crucial importance of $\lambda$-geodesic convexity for establishing guarantees of gradient flows in Wasserstein space:

\begin{proposition}[{\citealt[Prop 4.13]{santambrogio2017euclidean}}]\label{prop:uniqueness_flow_simple}
    Suppose that $F: \mathbb{W}_2(\cX) \rightarrow \R \cup\{+\infty\}$ is $\lambda$-geodesically convex and that the two curves $\rho_t^0$ and $\rho_t^1$ are solutions of \eqref{eq:wass_gradflow_general}. Then, setting $\delta(t) = \frac{1}{2}\W_2^2(\rho_t^1, \rho_t^2)$, we have
    \[ \delta'(t) \leq -2 \lambda d(t) \]
    This implies uniqueness of the solution of \eqref{eq:wass_gradflow_general} for a fixed initial state, stability and exponential convergence of the flow as $t\rightarrow +\infty$ if $\lambda>0$.
\end{proposition}

Unfortunately\;---and somewhat unexpectedly---\;the functional  $\cT_{\beta}(\rho) = \W_2^2(\rho, \beta)$ turns out to be not displacement convex in general. However, it does satisfy an alternate and more general notion of convexity: along generalized geodesics. 

\begin{definition}
    Let $\rho \in \cP(\cX)$ be fixed. For every pair $\mu^1, \mu^2 \in \cP(\cX)$, a \textbf{generalized geodesic} between them with base $\rho$ in $\mathbb{W}_2(\cX)$ is given by the curve $\mu_t = \bigl((1-t)T_0 + tT_1)_{\sharp}\rho$ where $T_i$ is the optimal transport map (for the squared cost) from $\rho$ to $\mu^i$. 
\end{definition}

Thus, a functional $F: \mathbb{W}_p(\cX) \rightarrow \R \cup\{+\infty\}$ is $\lambda$-geodesically convex along generalized geodesics if it satisfies condition \eqref{eq:displacement_convex} for \textit{generalized} geodesics. Under the same assumptions as above, the functionals $\cV, \cW$, and $\cF$ are all convex along generalized geodesics too \citep{santambrogio2017euclidean, ambrosio2005gradient}. But now, as hinted at before, so is $\cT_{\beta}(\rho)$ if we choose $\beta$ as the base point of the generalized geodesics \citep{santambrogio2015otam}. 

The notion of convexity along generalized geodesics can be used to establish results analogous to Proposition~\ref{prop:uniqueness_flow_simple} but which apply to more general functionals, including $\cT_{\beta}(\rho)$. Such results usually involve appealing to a characterization of gradient flows known as the evolution variational inequality (EVI):
\begin{equation}
    \frac{\dif}{\dif t} \frac{1}{2} d(\rho_t, \beta) \leq F(\beta) - F(\rho_t) -\frac{\lambda}{2}d(\mu_t, \beta)^2 \qquad \forall \beta \in \cP(\cX)
\end{equation}
Convexity along generalized geodesics can be used to prove the EVI conditions holds for a certain functional, which in turn implies uniqueness and stability of the flow. We refer the reader to \citep{santambrogio2017euclidean} for further details.

\section{First Variations, Gradient Flows, and Connections to PDEs.}\label{sec:pde_view}

\subsection{First variation of a functional}
As mentioned in Section~\ref{sec:into_flows}, having a notion of derivative of functionals over measures is a crucial step towards defining gradient flows in that space. The notion we rely on here is that of first variation of a functional \citep{santambrogio2017euclidean}:
\begin{definition}
	Given a functional $F:\cP(\Omega) \rightarrow \R$, consider perturbations $\chi$ such that at least for every $\epsilon \in [0, \epsilon_0]$,  $\rho+\epsilon \chi \in \cP(\Omega)$. If there exists a function $G$ such that 
	\[ \eval[1]{\frac{\dif}{\dif \epsilon} F(\rho + \epsilon \chi)}_{\epsilon=0} = \int G(\rho) \dif \chi \]
	for every such perturbation $\chi$, we call it the \textbf{first variation} of $F$ at $\rho$, and denote it by $\frac{\delta F}{\delta \rho}$.
\end{definition}

\subsection{Gradient flows and PDEs}
The connection between OT and certain diffusive partial differential equations (PDE) has been well studied over the past two decades \citep{jordan1998variational, otto2001geometry}. Indeed, equation \eqref{eq:wass_gradflow_general} defines a PDE over densities $\rho$. As mentioned before, it has a fluid dynamics interpretation as a continuity equation on a density-dependent flow velocity vector field $\mathbf{u} \triangleq  -\nabla \fvar[F]{\rho}(\rho)$, or a conservation-of-energy PDE for the energy flux $\mathbf{q} \triangleq -\rho \nabla \fvar[F]{\rho}(\rho)$. In the context of densities and datasets, this PDE can be roughly understood as a conservation-of-mass principle: no probability mass is created or destroyed in the sequence of densities on $\cX\times\cY$ that solve this system. 

For a functional of the form \eqref{eq:functional_sum} with only $\cF, \cV, \cW$ terms, the corresponding PDE (eq.~\eqref{eq:wass_gradflow_f}) is known as a diffusion–advection–interaction equation. Certain choices of functionals $\cF, \cV, \cW$ recover familiar PDEs. For example, taking $F(\rho) = \cF(\rho) + \cV(\rho)$, and $f(t)=t\log t$, the gradient flow of $F$ solves a Fokker-Planck equation \citep{santambrogio2015otam}:
\[ \partial_t \rho - \Delta \rho - \nabla \cdot (\rho \nabla V) = 0. \]
In dataset space, this equation can be interpreted as the time evolution of a dataset subject to a drift force imposed by the potential function $V$ and a constant-variance diffusion term ($\Delta \rho$) resulting from the entropy-inducing functional $\cF$. Other choices of functionals allow us to recover the advection equation, porous-media equation, and various other diffusion–advection–interaction PDEs \citep{santambrogio2017euclidean}. As we did for the Fokker-Planck equation, interpreting these PDEs in our context of dataset dynamics might yield interesting insights for designing objective functions. 

\section{Implementation and Experimental Details}\label{sec:experiment_details}
We implement our method on \href{https://pytorch.org/}{\texttt{PyTorch}} \citep{paszke2019pytorch}, using the \href{https://www.kernel-operations.io/geomloss/}{\texttt{geomloss}} \citep{feydy2019interpolating} and \href{https://pythonot.github.io/}{\texttt{POT}} \citep{flamary2021pot} libraries for OT-related computations, including the \textsc{OTDD} distance needed at every step. The three types of feature-label dynamics described in Section~\ref{sec:implementation} are implemented by detaching parts of the computational graph in order to make gradient updates only in some of them. For the variable label dynamics, there are two options for clustering: fixed-size or nonparametric. We use k-means for the former and density-based spatial clustering of applications with noise (DBSCAN) with parameters $\epsilon=5$ and minimum points per cluster $4$ for the latter. Pseudocode for the three types of feature-label gradient flow dynamics described in Section~\ref{sec:implementation} is shown here in \Cref{algo:fd-fl,algo:jd-fl,algo:jd-vl}.

\ifbool{loadalgorithm2e}{
\begin{algorithm}
\SetKwInOut{Input}{input}
\DontPrintSemicolon
\KwIn{Initial particle feature matrix $X_0 \in \R^{d\times n}$ and corresponding labels $\y \in \{0,\dots,k\}^{n}$.}
 \text{requires\_gradient}$(X_0) \gets \texttt{True}$\; 
 \For{$t=0,1,\dots,T$}{
  $\ell \gets F(\X_{t}, \y)$\;
  $\X_t \gets \text{optim\_step}(\nabla_{\X}\medspace \ell)$\;
  \For{every class $j=1,\dots,k$}{
  $\mu_i, \Sigma_i \gets \text{getstats}(\{\x_t^i \st y_i = j \})$\;
  }
  $\text{recompute\_label\_distances}(\{\mu_i\}, \{\Sigma_i\})$ \tcp*{subroutine in OTDD \sref{sec:otdd}}
 }
 \caption{Gradient flow with \texttt{fd-fl} dynamics.}\label{algo:fd-fl}
\end{algorithm}
}{
\begin{algorithm}
    \caption{Gradient flow with feature-driven fixed-label (\texttt{fd-fl}) dynamics.}\label{algo:fd-fl}
    \begin{algorithmic}
        \STATE {\bfseries Input:} Initial particle feature matrix $X_0 \in \R^{d\times n}$ and corresponding labels $\y \in \{0,\dots,k\}^{n}$.
        \STATE  \text{requires\_gradient}$(X_0) \gets \texttt{True}$
        \FOR{time $t=0$ {\bfseries to} $T$}
            \STATE $\ell \gets F(\X_{t}, \y)$
            \STATE  $\X_t \gets \text{optim\_step}(\nabla_{\X}\medspace \ell)$
            \FOR{every class $j=1$ {\bfseries to} $k$}
                \STATE $\mu^j_t, \Sigma^j_t \gets \text{getstats}(\{\x_t^i \st y_i = j \})$
            \ENDFOR
            \STATE $\text{recompute\_label\_distances}(\{\mu^j_t\}, \{\Sigma^j_t\})$ \COMMENT{subroutine in OTDD \sref{sec:otdd}}
        \ENDFOR
\end{algorithmic}
\end{algorithm}
\begin{algorithm}
    \caption{Gradient flow with joint-driven fixed-label (\texttt{jd-fl}) dynamics.}\label{algo:jd-fl}
    \begin{algorithmic}
        \STATE {\bfseries Input:} Initial particle feature matrix $X_0 \in \R^{d\times n}$ and corresponding labels $\y \in \{0,\dots,k\}^{n}$.
        \STATE  \text{requires\_gradient}$(X_0, \Sigma^j_0, \mu^j_0) \gets \texttt{True}$
        \FOR{time $t=0$ {\bfseries to} $T$}
            \STATE $\ell \gets F(\X_{t}, \y)$
            \STATE  $\X_t \gets \text{optim\_step}(\nabla_{\X}\medspace \ell)$
            \FOR{every class $j=1$ {\bfseries to} $k$}
                \STATE  $\mu^j_t \gets \text{optim\_step}(\nabla_{\mu^j}\medspace \ell)$
                \STATE  $\Sigma^j_t \gets \text{optim\_step}(\nabla_{\Sigma^j}\medspace \ell)$
            \ENDFOR
            \STATE $\text{recompute\_label\_distances}(\{\mu^j_t\}, \{\Sigma^j_t\})$ \COMMENT{subroutine in OTDD \sref{sec:otdd}}
        \ENDFOR
    \end{algorithmic}
\end{algorithm}
\begin{algorithm}
    \caption{Gradient flow with joint-driven variable-label (\texttt{jd-vl}) dynamics.}\label{algo:jd-vl}
    \begin{algorithmic}
        \STATE {\bfseries Input:} Initial particle feature matrix $X_0 \in \R^{d\times n}$ and corresponding labels $\y \in \{0,\dots,k\}^{n}$.
        \STATE  \text{requires\_gradient}$(X_0, \Sigma^j_0, \mu^j_0) \gets \texttt{True}$
        \FOR{time $t=0$ {\bfseries to} $T$}
            \STATE $\ell \gets F(\X_{t}, \y)$
            \STATE  $\X_t \gets \text{optim\_step}(\nabla_{\X}\medspace \ell)$
            \FOR{every particle $i=1$ {\bfseries to} $n$}
                \STATE  $\mu^i_t \gets \text{optim\_step}(\nabla_{\mu^i}\medspace \ell)$
                \STATE  $\Sigma^i_t \gets \text{optim\_step}(\nabla_{\Sigma^i}\medspace \ell)$
            \ENDFOR
            \STATE $\mathbf{y}_t \gets \text{clustering\_method}(\{\mu_t\}, \{\Sigma_t\})$ \COMMENT{recompute discrete labels by clustering}
            \STATE $\text{recompute\_label\_distances}(\{\mu^j_t\}, \{\Sigma^j_t\})$ \COMMENT{subroutine in OTDD \sref{sec:otdd}}
        \ENDFOR
    \end{algorithmic}
\end{algorithm}
}

For the parametrized flow mapping $h_{\text{flow}}$ (\sref{sec:transfer_mnist}), we use an autoencoder-type architecture with an encoder consisting of 2 convolutional and 5 fully-connected layers, and the decoder is a inverted copy of the encoder. It was trained for 20 epochs using \textsc{adam} with learning rate \SI{1e-3}, using ten different random restarts and choosing the best performing one in a held-out set. For transfer learning (\sref{sec:transfer_mnist}), we use a LeNet-5 architecture with ReLU nonlinearities trained for 20 epochs using \textsc{adam} with learning rate \SI{1e-3} and weight decay \SI{1e-6}. It was fine-tuned for 10 epochs on the target domain(s) using the same optimization parameters. For the experiments in Table~\ref{tab:polimodality}, we use 5K source (\mnist) and target (other *\textsc{nist} datasets) samples. For both supervised and unsupervised flows, we use a flow step size of \SI{1e-1}, 1000 steps, and entropy regularization $\lambda=$\SI{1e2}. For the unsupervised flow, we permute the values of the pseudo-labels obtained through clustering to match them to the indices of the target labels so as to allow accuracy computation.

All experiments were run on the same machine with an Intel Xeon 32-core 2.00GHz CPU with a single GeForce RTX 2080 Ti GPU. In this machine, the flows on synthetic datasets of Section~\ref{sec:implementation} run at $<$0.2s per step, while the flows for the image classification datasets of Sections~\ref{sec:implementation} and \ref{sec:transfer_mnist} run at $\sim$~5s per step for 2K particles, for a total flow runtime of less than 5 minutes. Information about all the datasets used, including references, are provided in Table~\ref{tab:dataset_details}.


\begin{table}
  \small
  \centering
    \caption{Summary of datasets used. $\ast$: we rescale the \usps digits to $28\times 28$ for comparison to the *\textsc{NIST} datasets, and the \textsc{stl-10} and \textsc{camelyon} to $32\times 32$ for comparison to \textsc{cifar-10}. 
    }\label{tab:dataset_details}  
      \flushleft
         \resizebox{\linewidth}{!}{%
		\begin{tabular}{r c c c c c} 
			\toprule
			 Dataset & Input Dimension & Number of Classes & Train Examples & Test Examples & Source \\
			\midrule
            \textsc{usps} & $16\times 16^{\ast}$ & $10$ & $7291$ & $2007$ & \citep{hull1994database} \\
            \textsc{mnist} & $28\times 28$ & $10$ & $60$K & $10$K & \citep{lecun2010mnist}\\
            \textsc{kmnist} & $28\times 28$ & $10$ & $60$K & $10$K & \citep{clanuwat2018deep}\\
            \textsc{fashion-mnist} & $28\times 28$ & 10 &$60$K & $10$K & \citep{xiao2017fashion-mnist}\\
            \textsc{camelyon}  & $128\times 128^{\ast}$ & 2 &$262$K & $32$K & \citep{litjens2018camelyon} \\
            \textsc{cifar-10} & $32\times 32$ & $10$ & $50$K & $10$K & \citep{krizhevsky2009learning} \\
            \textsc{stl-10} & $96\times 96^{\ast}$ & $10$ & $5$K & $8$K & \citep{coates2011analysis} \\
			\bottomrule
		\end{tabular}%
    }
\end{table}

\clearpage
\pagebreak

\section{Additional Experimental Results on Gaussian Flows}\label{sec:additional_gaussian}

\begin{figure}[H]
    \centering
    \begin{subfigure}{\linewidth}
    \includegraphics[width=0.33\linewidth, trim={0.3cm 0.3cm 0.3cm 0.3cm},clip]{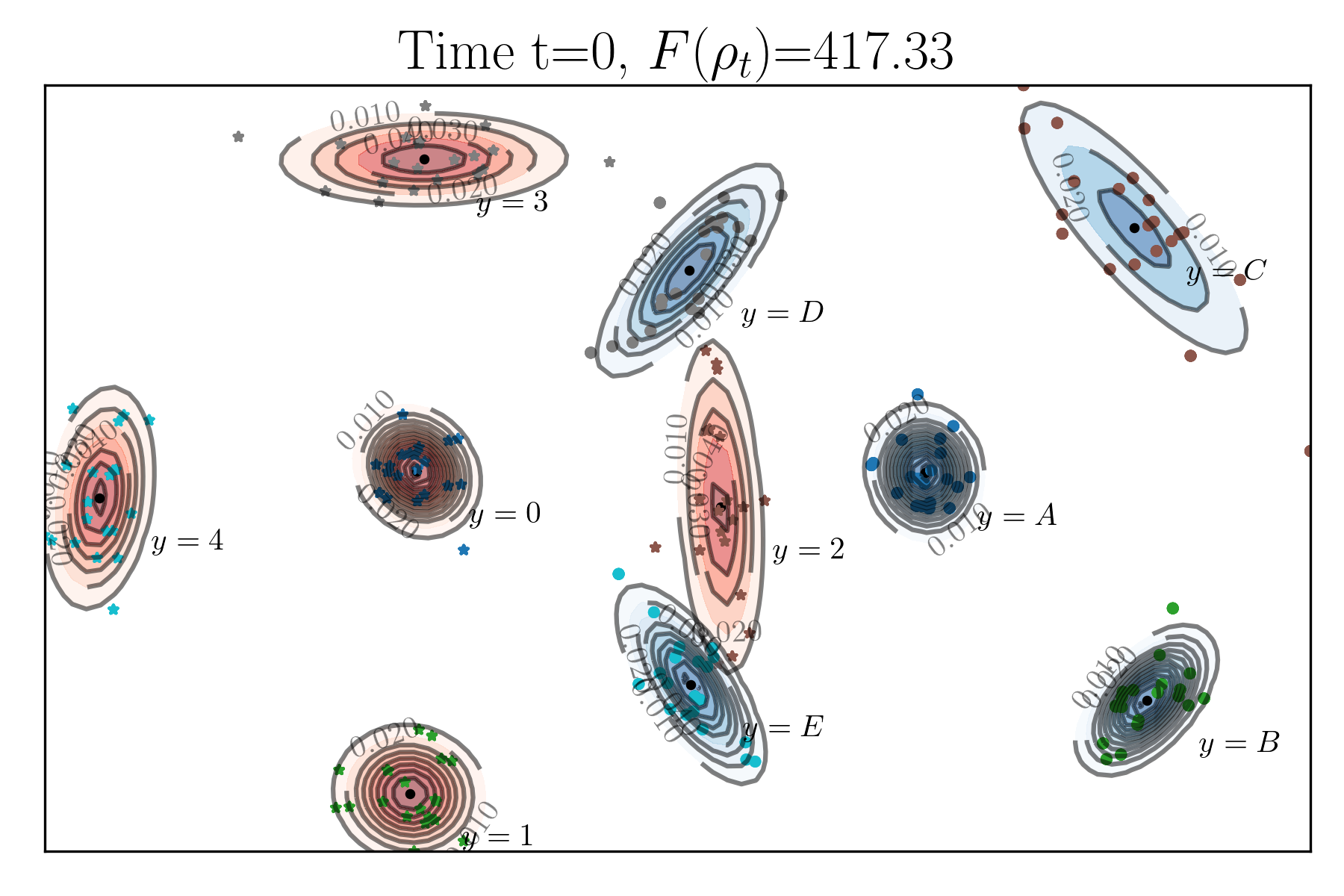}%
    \includegraphics[width=0.33\linewidth, trim={0.3cm 0.3cm 0.3cm 0.3cm},clip]{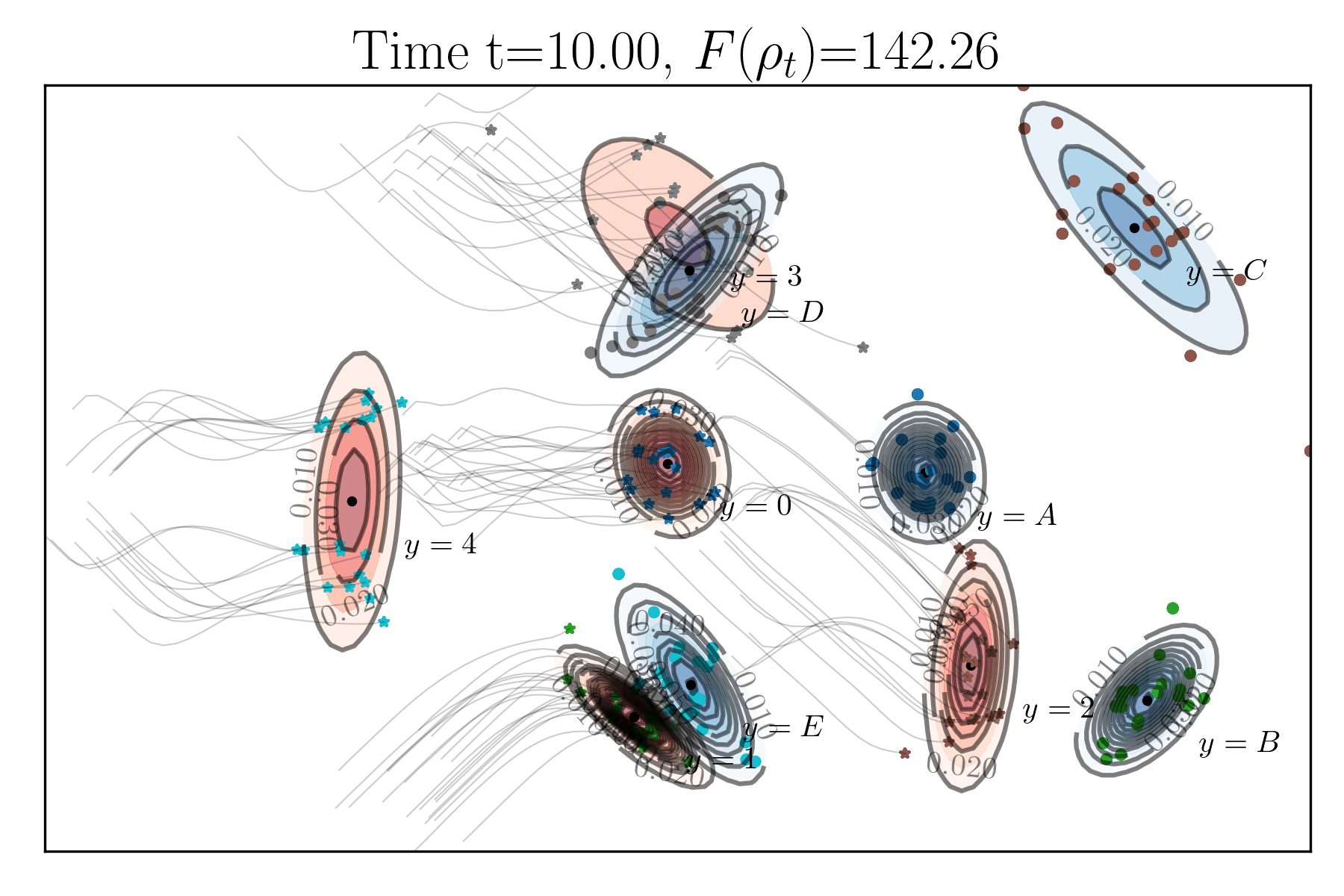}%
    \includegraphics[width=0.33\linewidth, trim={0.3cm 0.3cm 0.3cm 0.3cm},clip]{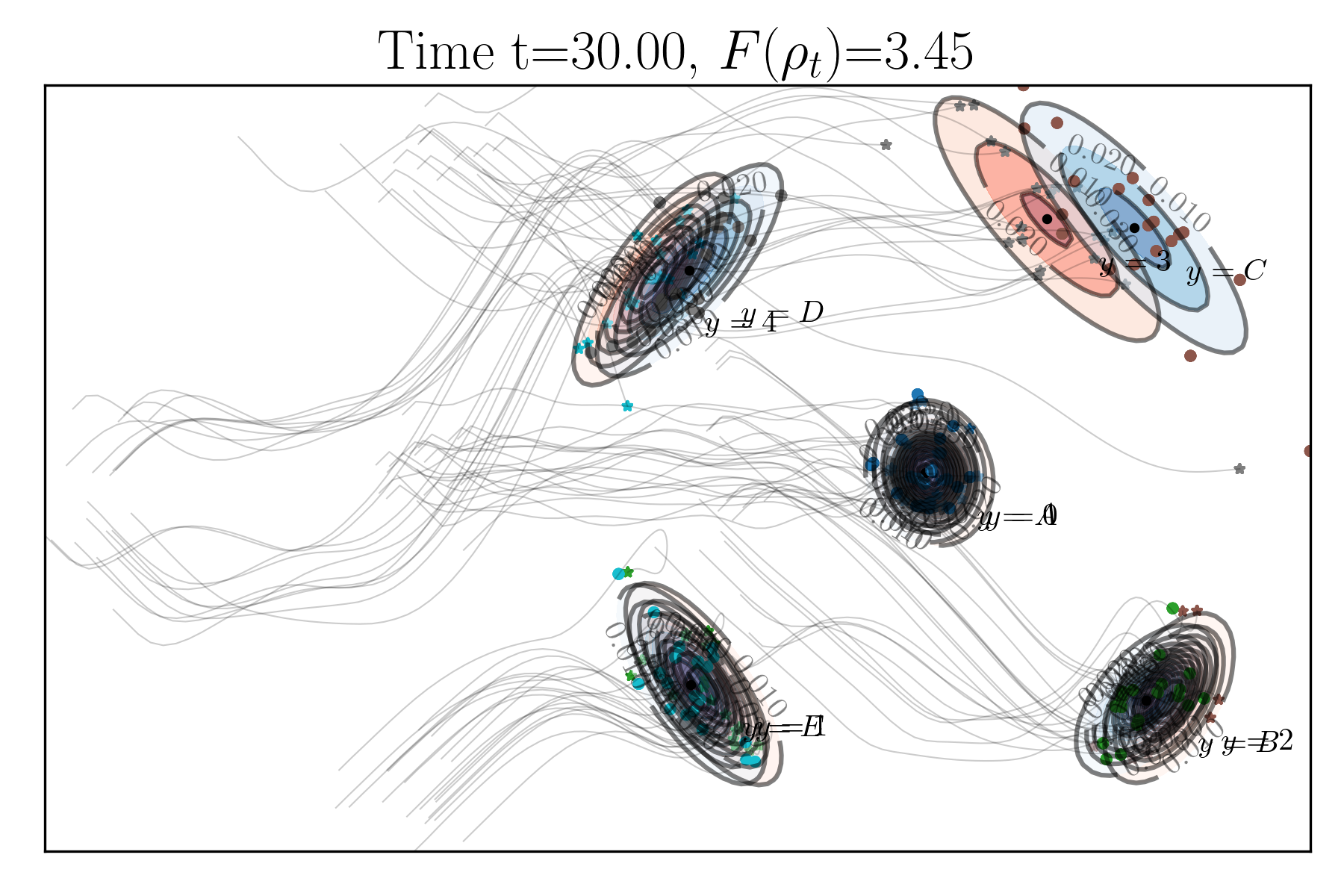}
	\caption{Feature-driven (\texttt{fd}) dynamics, \textsc{adam} optimizer. }\label{fig:flow_methods_gaussians_1}
	\vspace{0.5cm}
	\end{subfigure}
	\begin{subfigure}{\linewidth}
    \includegraphics[width=0.33\linewidth, trim={0.3cm 0.3cm 0.3cm 0.3cm},clip]{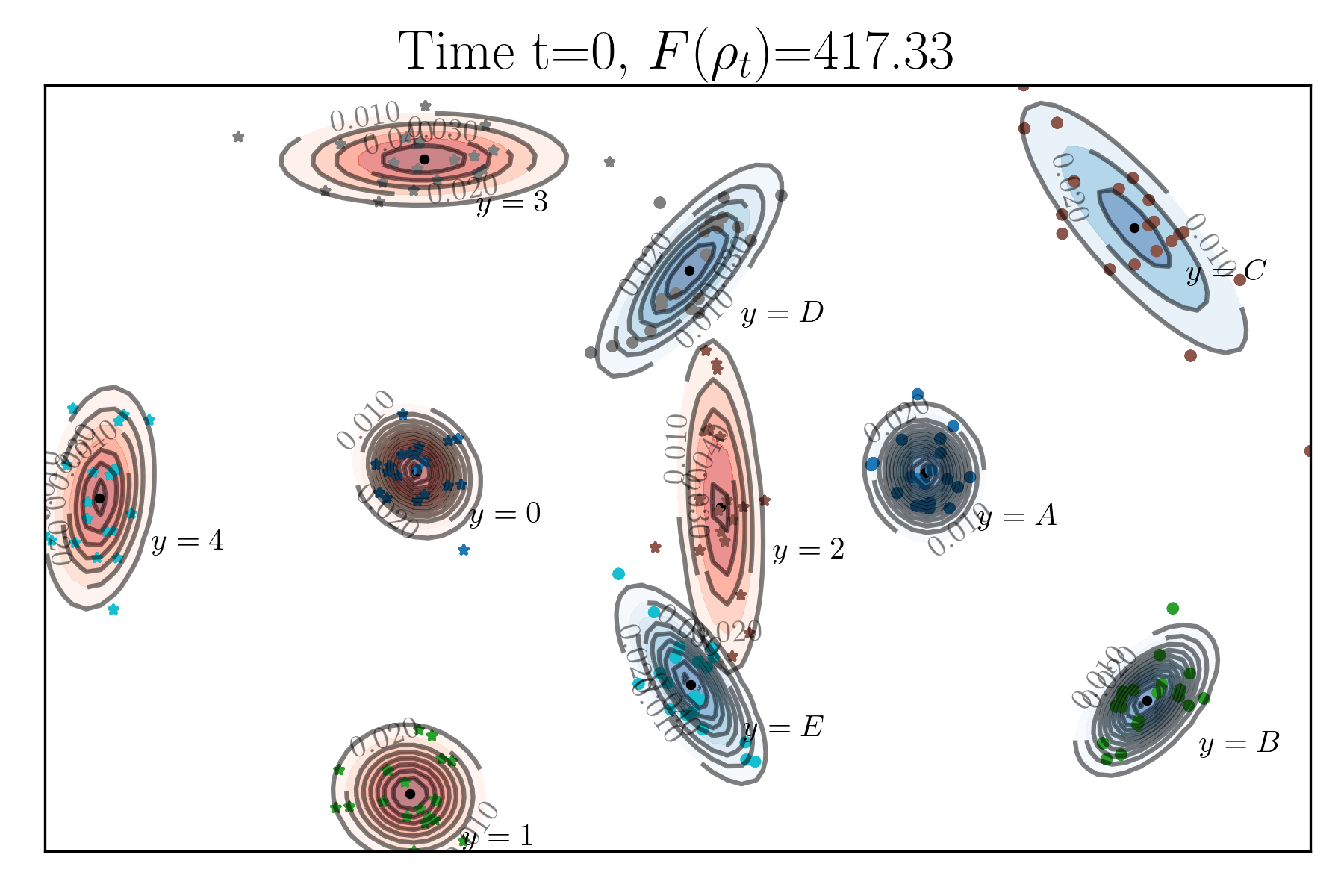}%
    \includegraphics[width=0.33\linewidth, trim={0.3cm 0.3cm 0.3cm 0.3cm},clip]{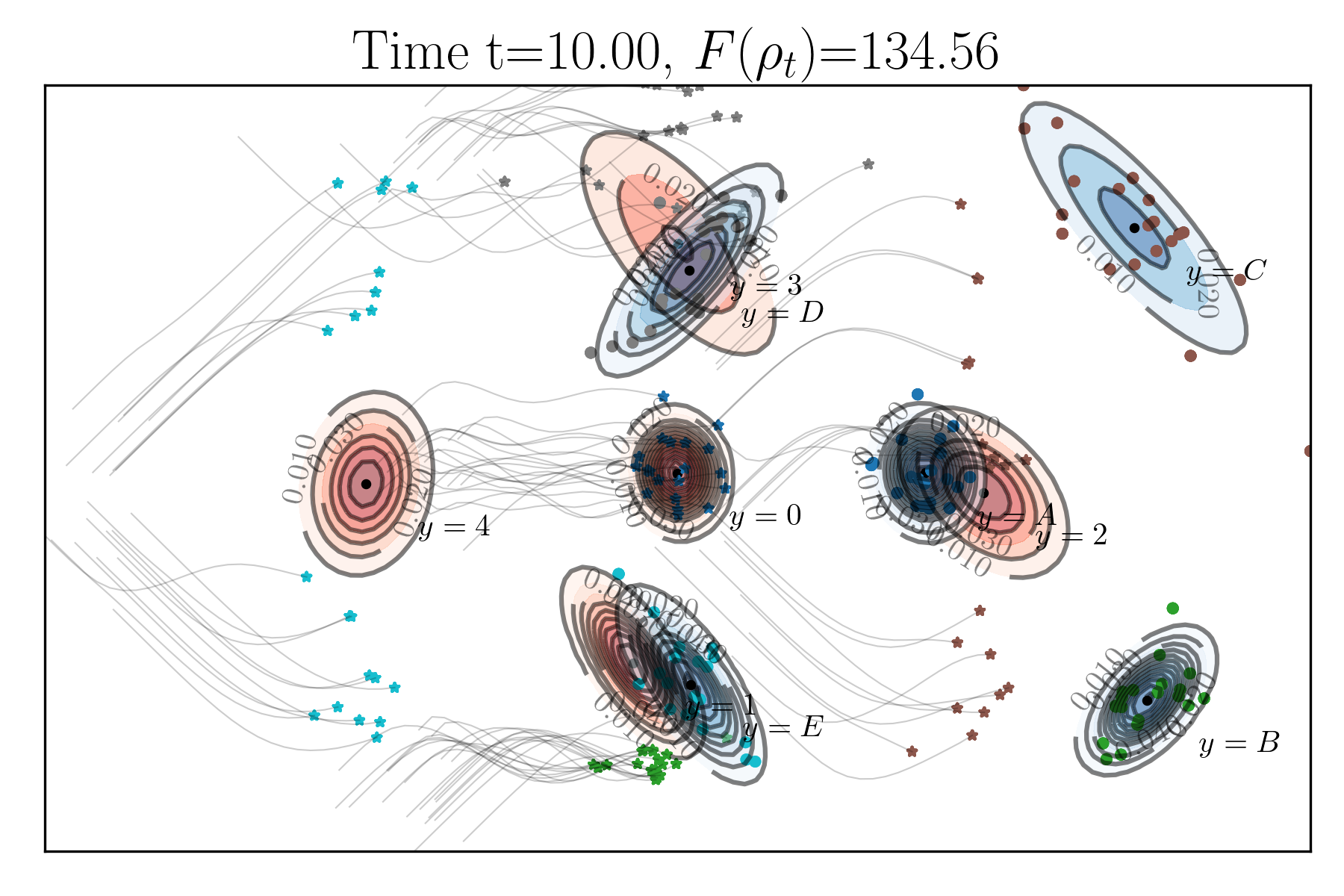}%
    \includegraphics[width=0.33\linewidth, trim={0.3cm 0.3cm 0.3cm 0.3cm},clip]{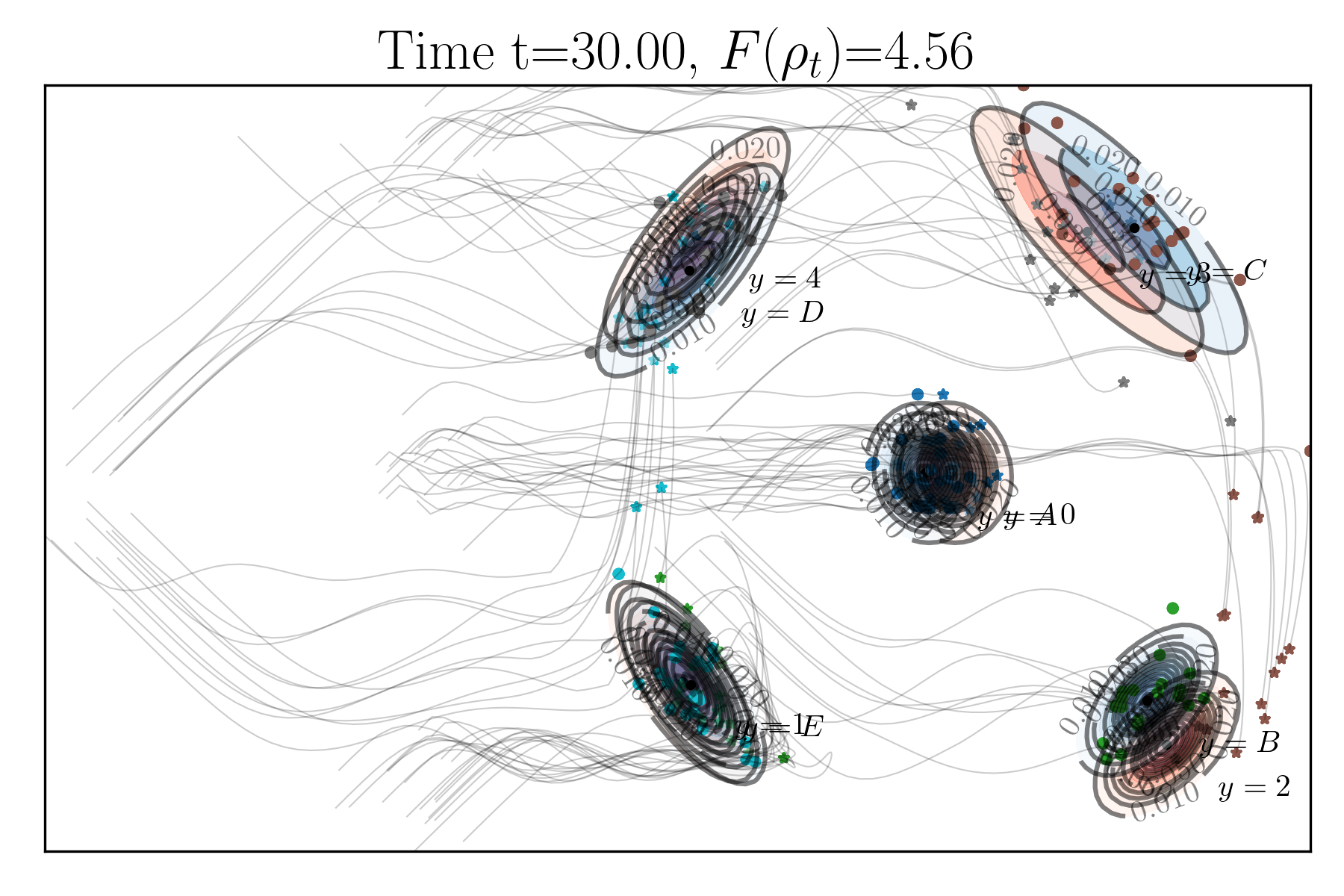}
	\caption{Joint-driven fixed-label (\texttt{jd-fl}) dynamics, \textsc{adam} optimizer.}\label{fig:flow_methods_gaussians_2}		
	\vspace{0.5cm}
	\end{subfigure}	
    \begin{subfigure}{\linewidth}
    \includegraphics[width=0.33\linewidth, trim={0.3cm 0.3cm 0.3cm 0.3cm},clip]{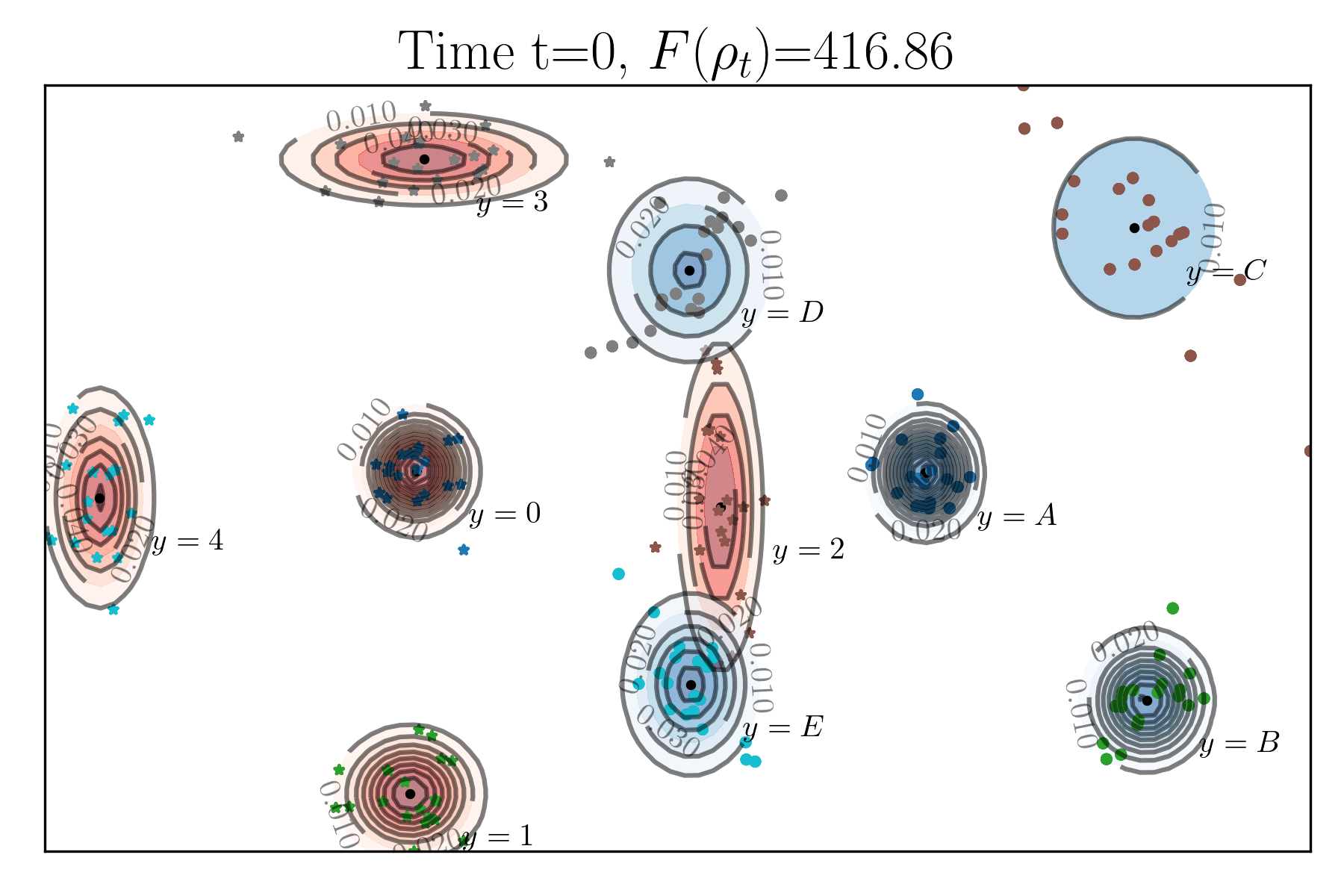}%
    \includegraphics[width=0.33\linewidth, trim={0.3cm 0.3cm 0.3cm 0.3cm},clip]{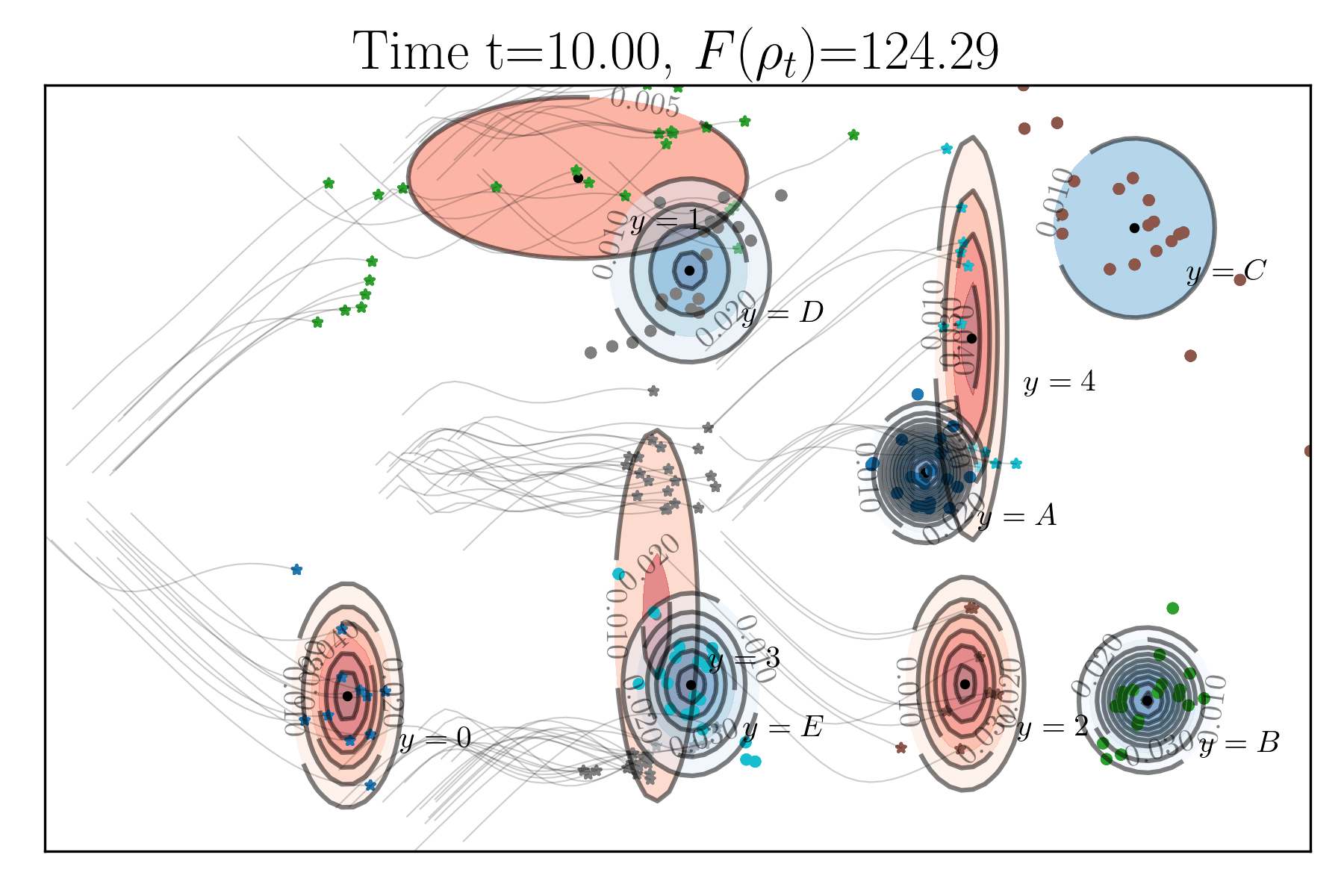}%
    \includegraphics[width=0.33\linewidth, trim={0.3cm 0.3cm 0.3cm 0.3cm},clip]{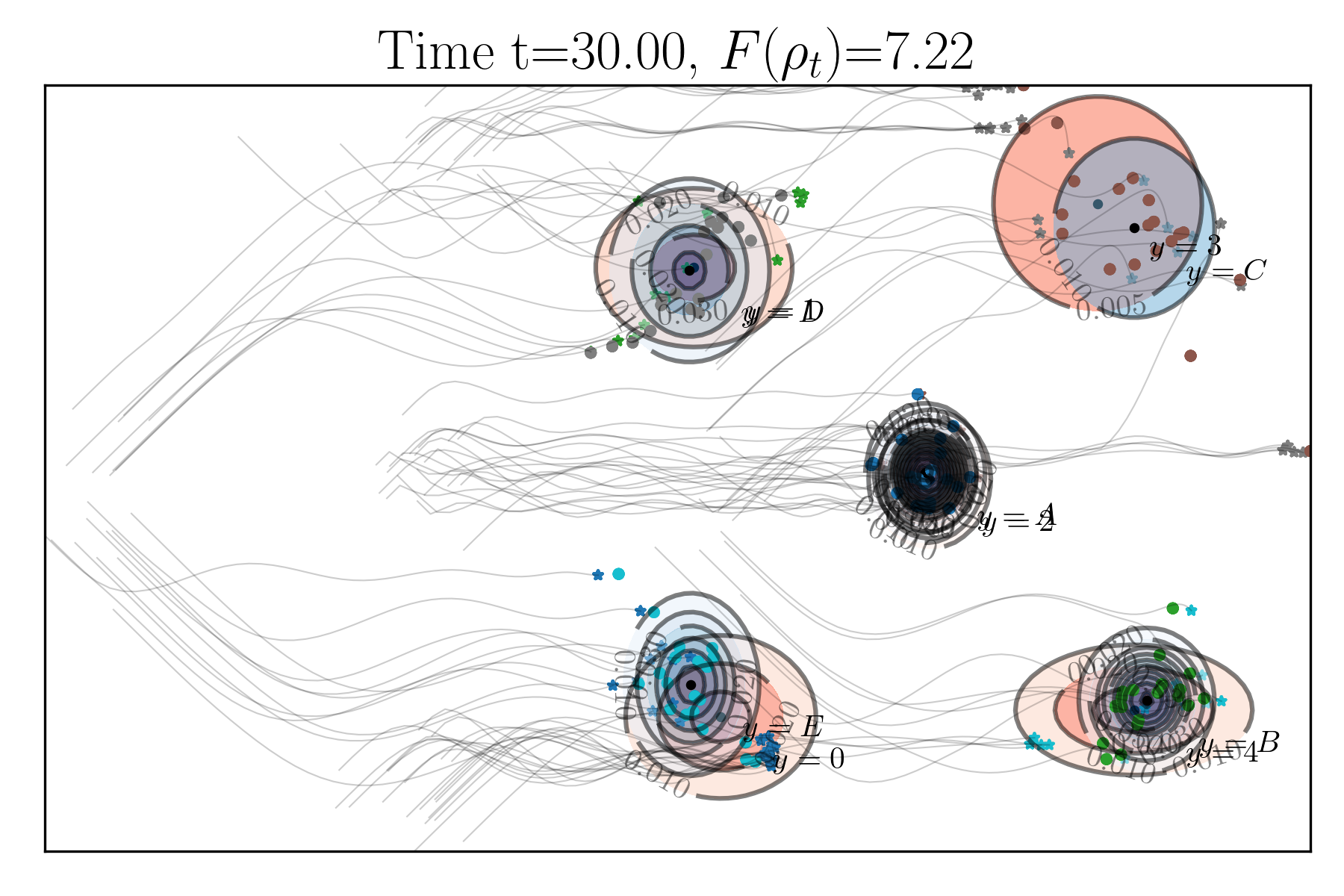}    
	\caption{Joint-driven variable-label (\texttt{jd-vl}) dynamics, k-means clustering, \textsc{adam} optimizer. }\label{fig:flow_methods_gaussians_3}
	\vspace{0.5cm}
	\end{subfigure}    
    \begin{subfigure}{\linewidth}
	\includegraphics[width=0.33\linewidth, trim={0.3cm 0.3cm 0.3cm 0.3cm},clip]{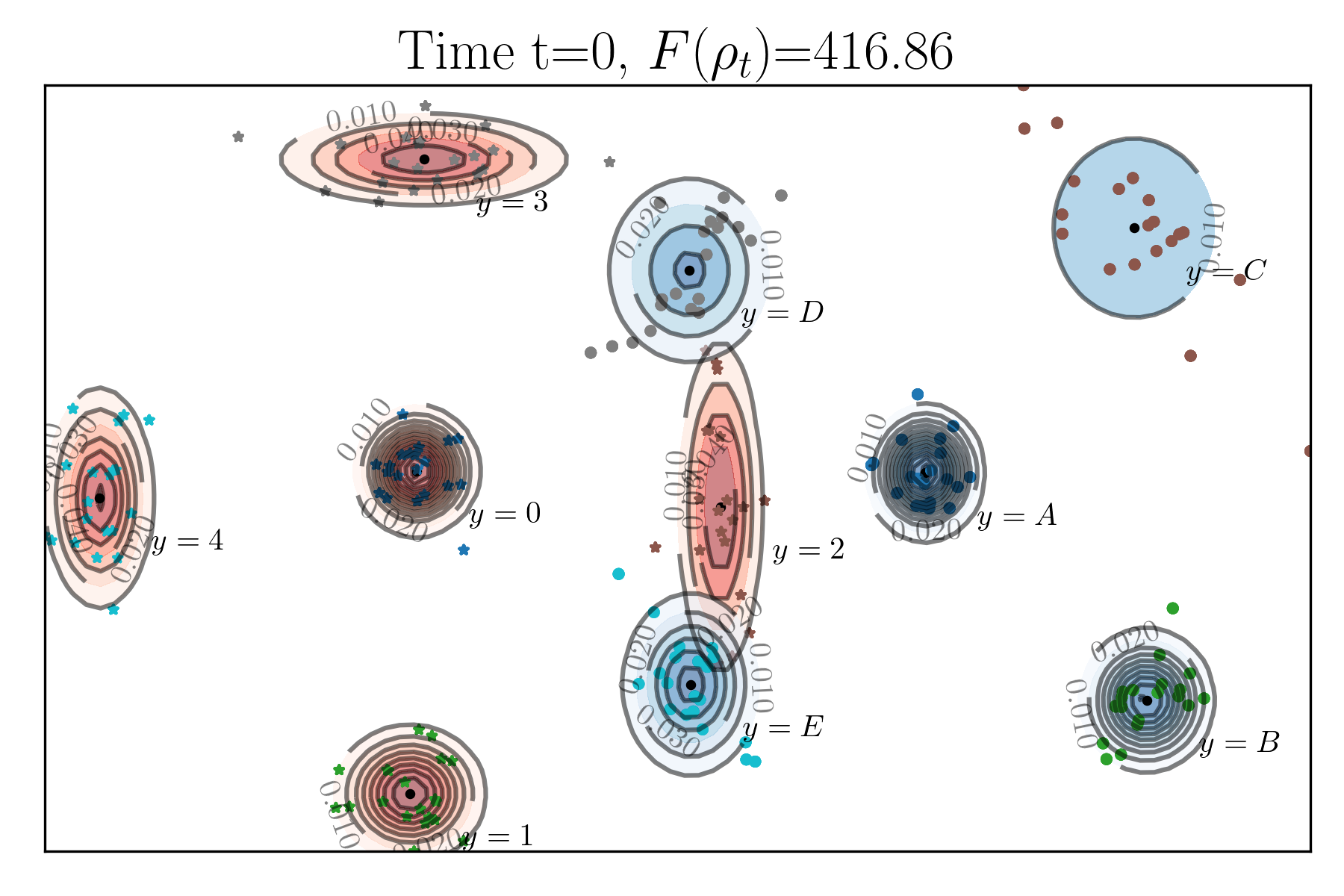}%
    \includegraphics[width=0.33\linewidth, trim={0.3cm 0.3cm 0.3cm 0.3cm},clip]{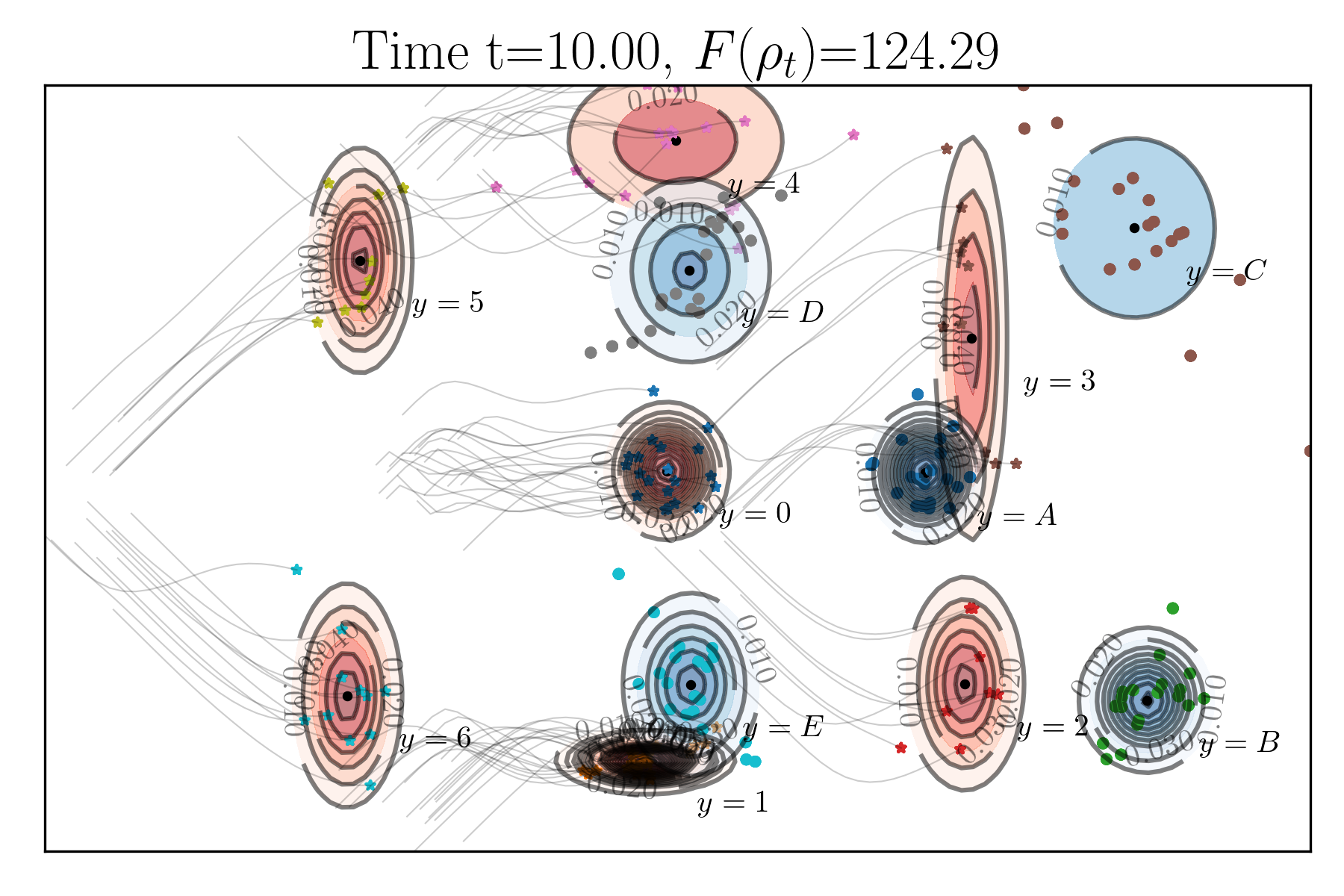}%
    \includegraphics[width=0.33\linewidth, trim={0.3cm 0.3cm 0.3cm 0.3cm},clip]{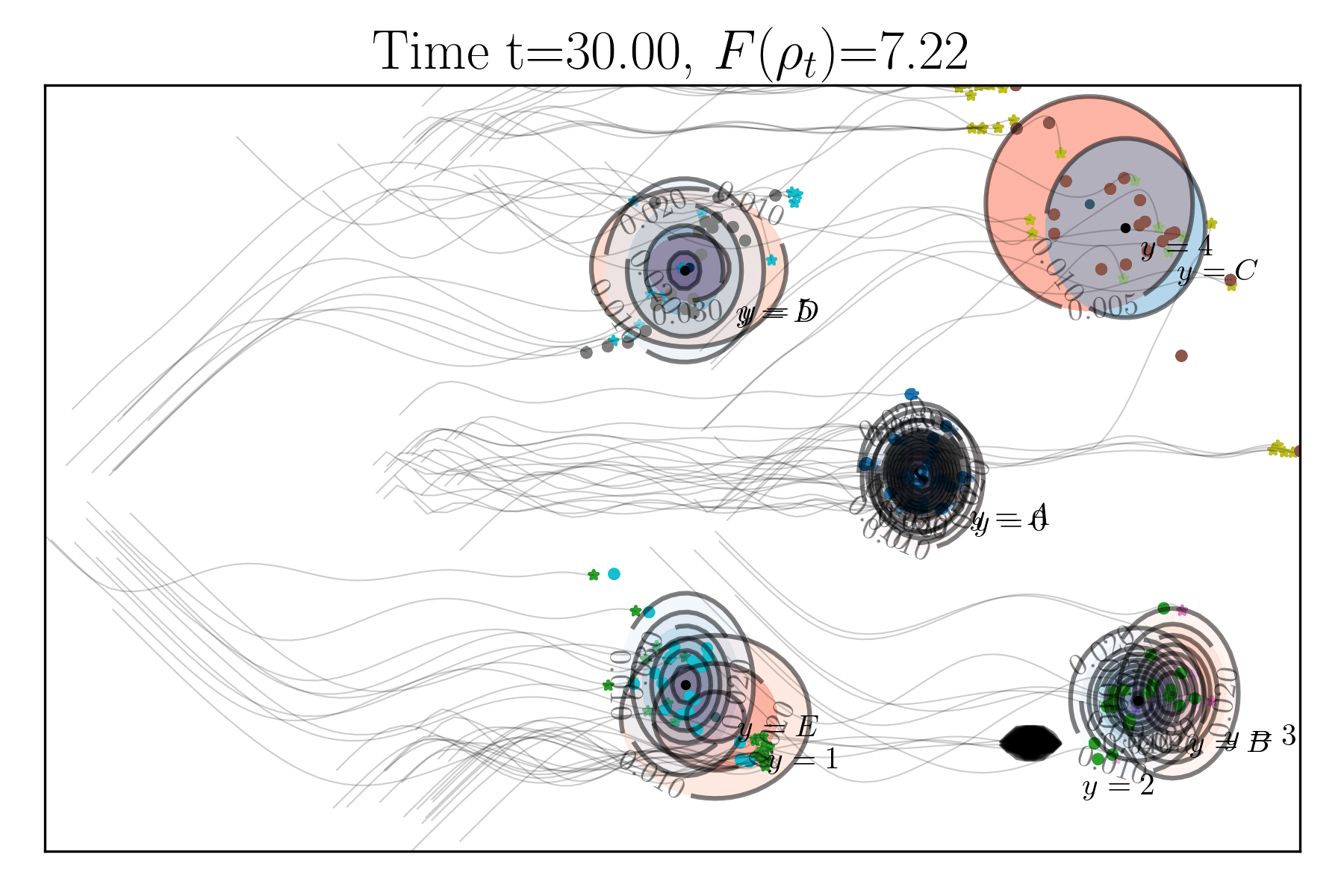} 
	\caption{Joint-driven variable-label (\texttt{jd-vl}) dynamics, DBSCAN clustering, \textsc{adam} optimizer.  }\label{fig:flow_methods_gaussians_4}
	\end{subfigure}
    \caption{Gradient flows driven by functional $\cT_{\beta}(\rho) = \text{OTDD}(\mathrm{D}_{\rho}, \mathrm{D}_{\beta})$ starting from dataset $\rho_0$ (red) advecting towards $\beta$ (blue) for various dynamic schemes (\sref{sec:implementation}).}
    \label{fig:flow_methods_gaussians}
\end{figure}

\begin{figure*}[p]
    \centering
    \begin{subfigure}{\linewidth}
        \includegraphics[width=0.33\linewidth, trim={0.3cm 0.3cm 0.3cm 0.3cm},clip]{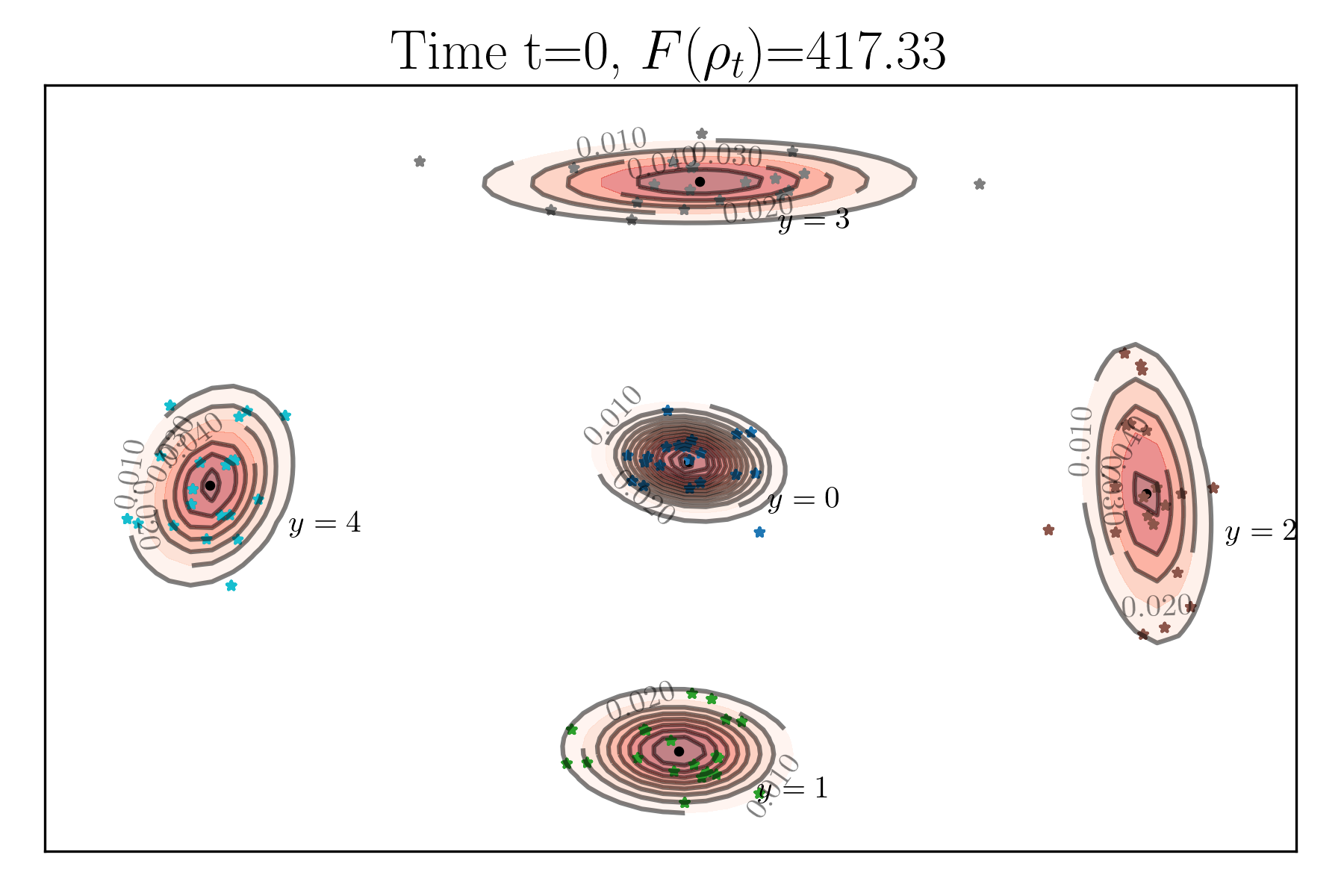}%
        \includegraphics[width=0.33\linewidth, trim={0.3cm 0.3cm 0.3cm 0.3cm},clip]{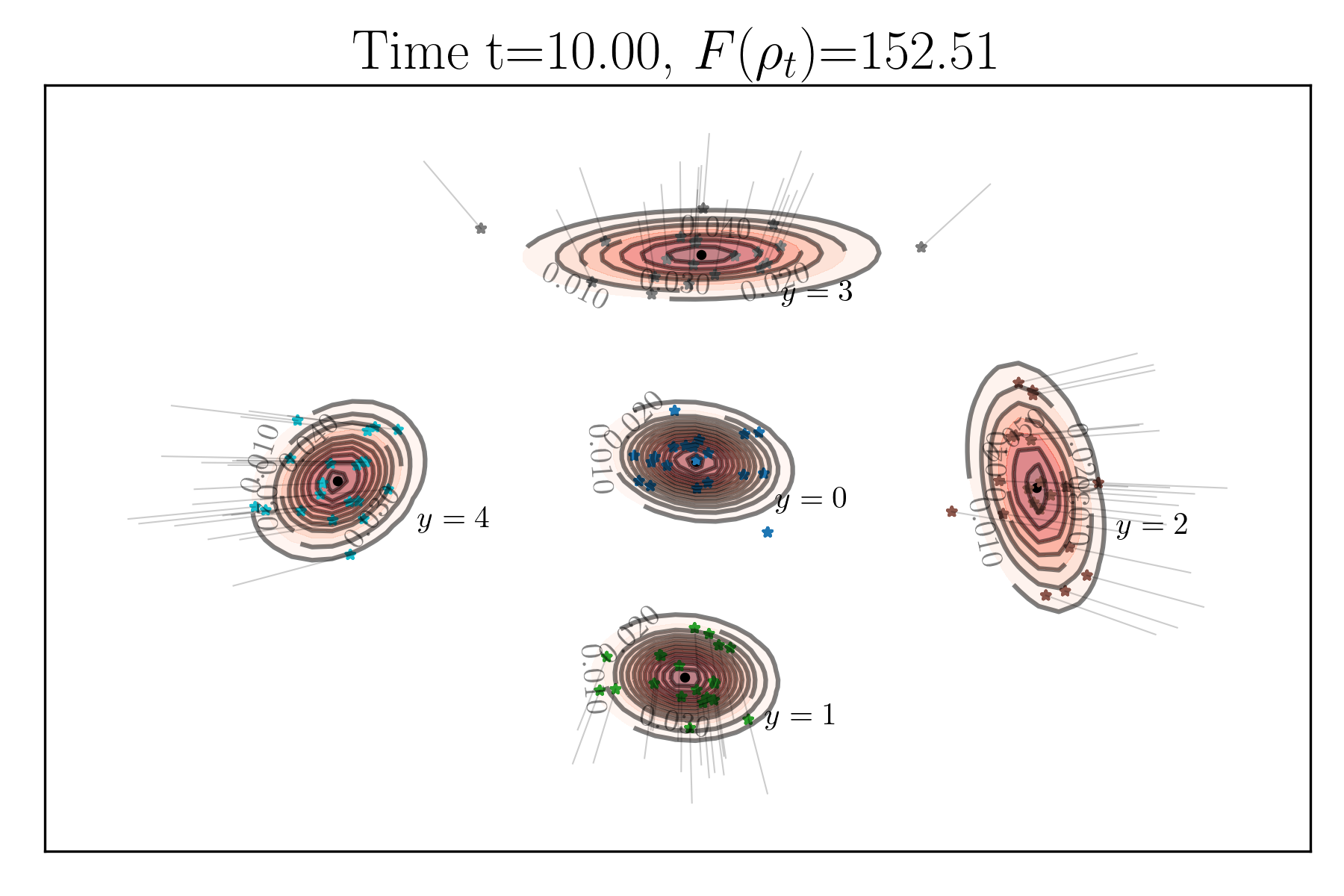}%
        \includegraphics[width=0.33\linewidth, trim={0.3cm 0.3cm 0.3cm 0.3cm},clip]{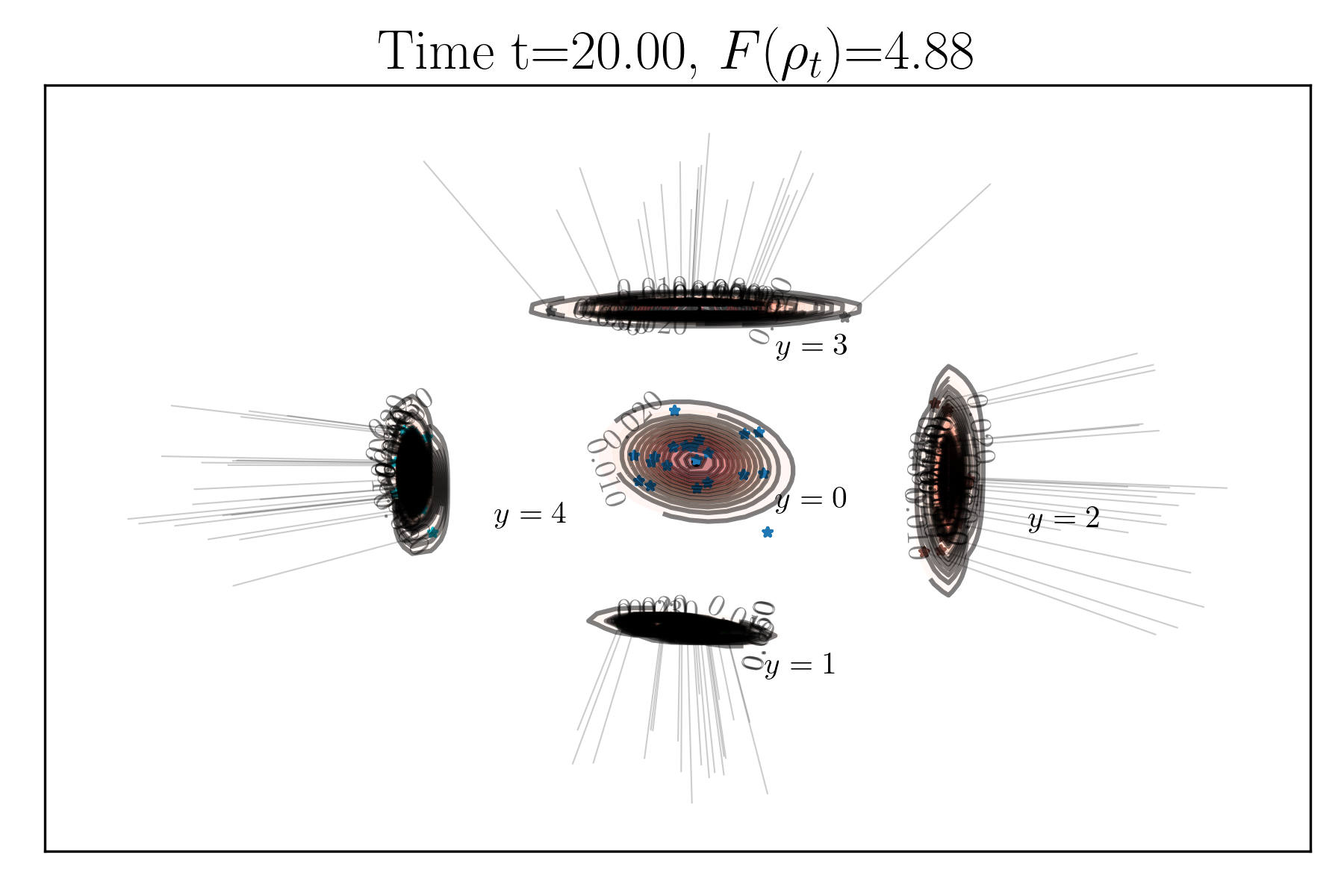}
    \caption{Functional: $F(\rho) =  \int (\|\x - \x_0\| - \tau)_{+} \dif \rho(z)$, \textsc{sgd} optimizer. }\label{fig:functionals_gaussians_1}
	\vspace{0.5cm}
	\end{subfigure}
    \begin{subfigure}{\linewidth}
        \includegraphics[width=0.33\linewidth, trim={0.3cm 0.3cm 0.3cm 0.3cm},clip]{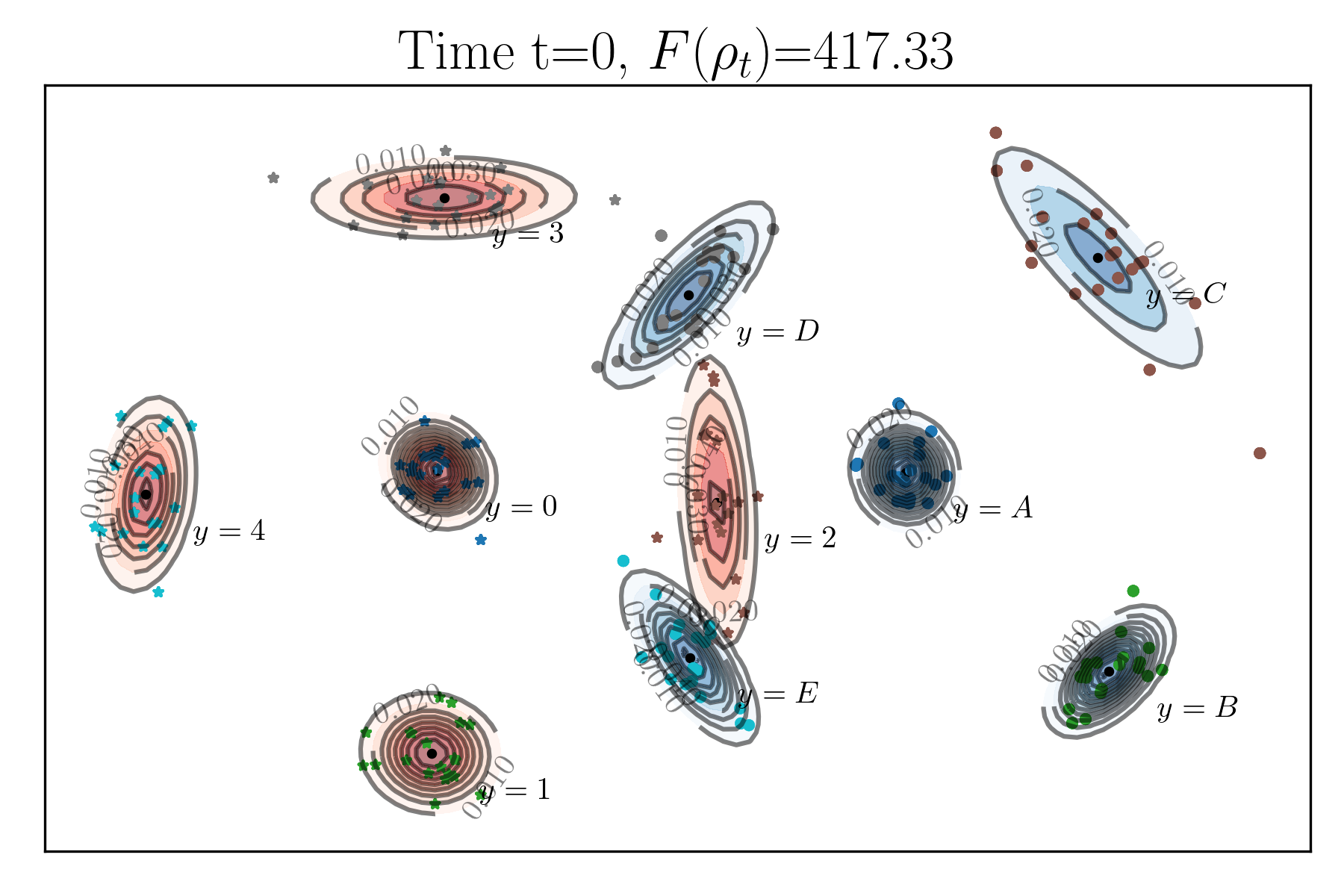}%
        \includegraphics[width=0.33\linewidth, trim={0.3cm 0.3cm 0.3cm 0.3cm},clip]{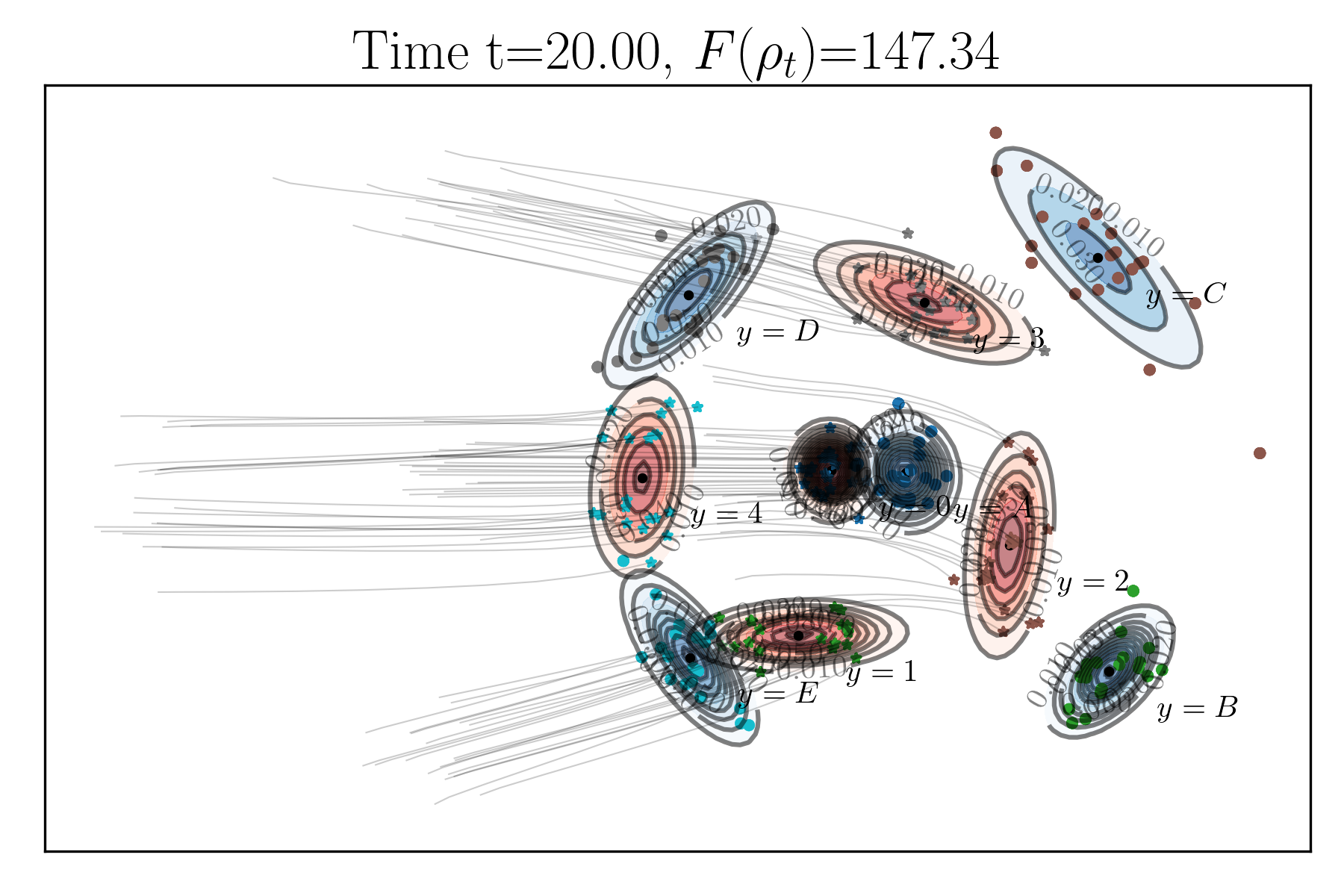}%
      \includegraphics[width=0.33\linewidth, trim={0.3cm 0.3cm 0.3cm 0.3cm},clip]{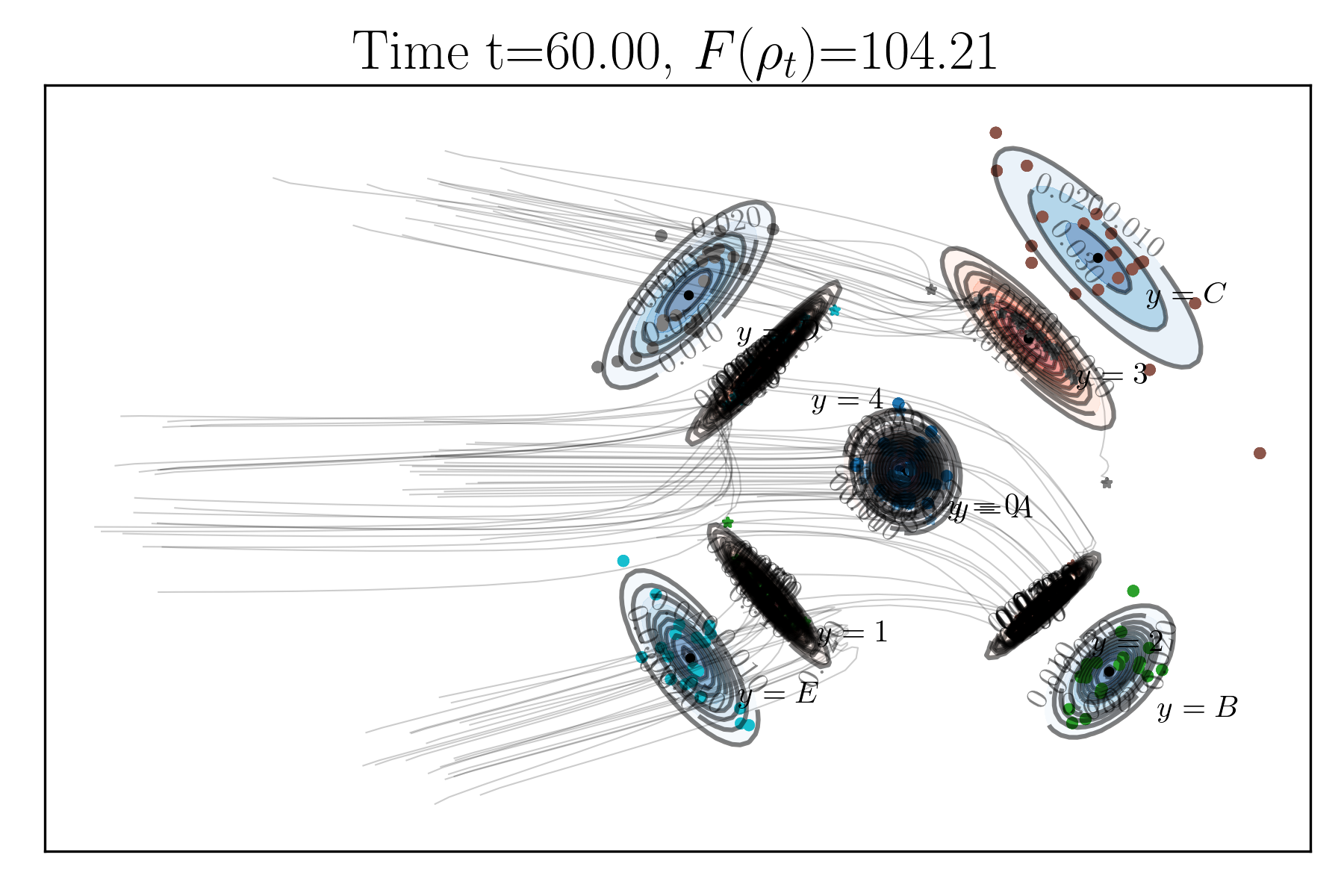}
    \caption{Functional: $F(\rho) = \text{OTDD}(\mathrm{D}_{\rho}, \mathrm{D}_{\beta}) + \lambda \int  (\|\x - \x_0\| - \tau)_{+} \dif \rho(z)$, \textsc{sgd} optimizer. }\label{fig:functionals_gaussians_2}
	\vspace{0.5cm}
	\end{subfigure}
    \begin{subfigure}{\linewidth}
        \includegraphics[width=0.33\linewidth, trim={0.3cm 0.3cm 0.3cm 0.3cm},clip]{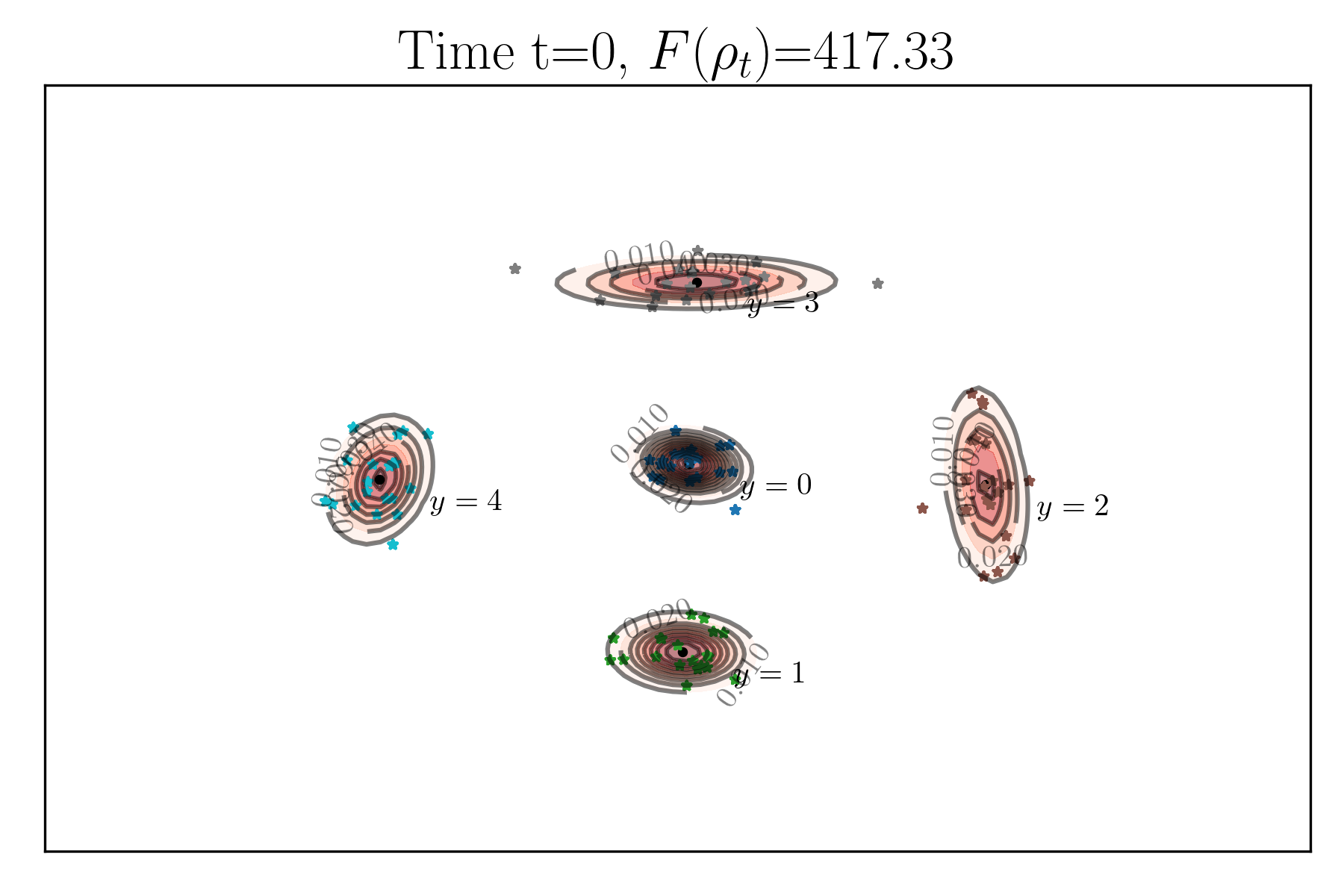}%
        \includegraphics[width=0.33\linewidth, trim={0.3cm 0.3cm 0.3cm 0.3cm},clip]{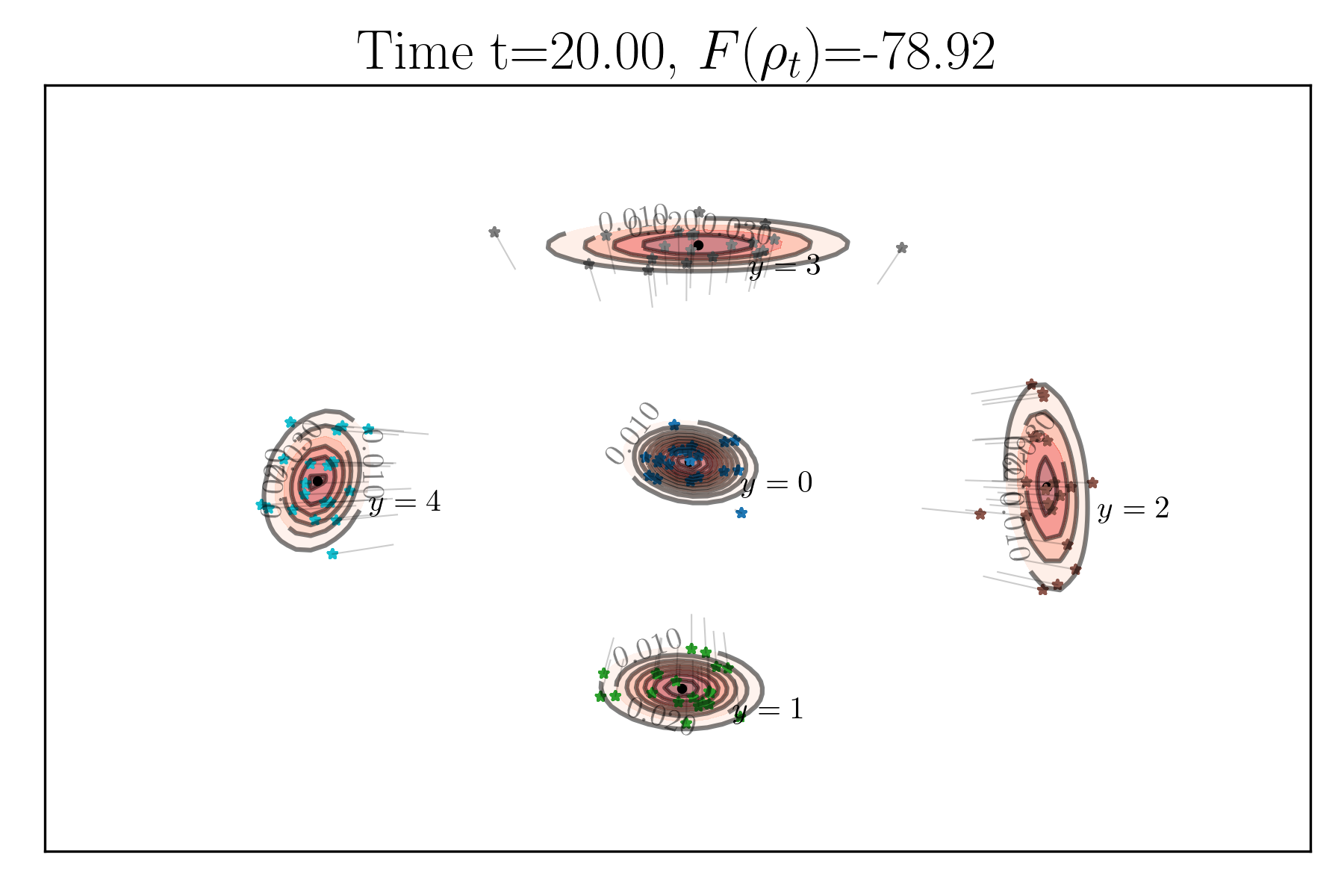}%
        \includegraphics[width=0.33\linewidth, trim={0.3cm 0.3cm 0.3cm 0.3cm},clip]{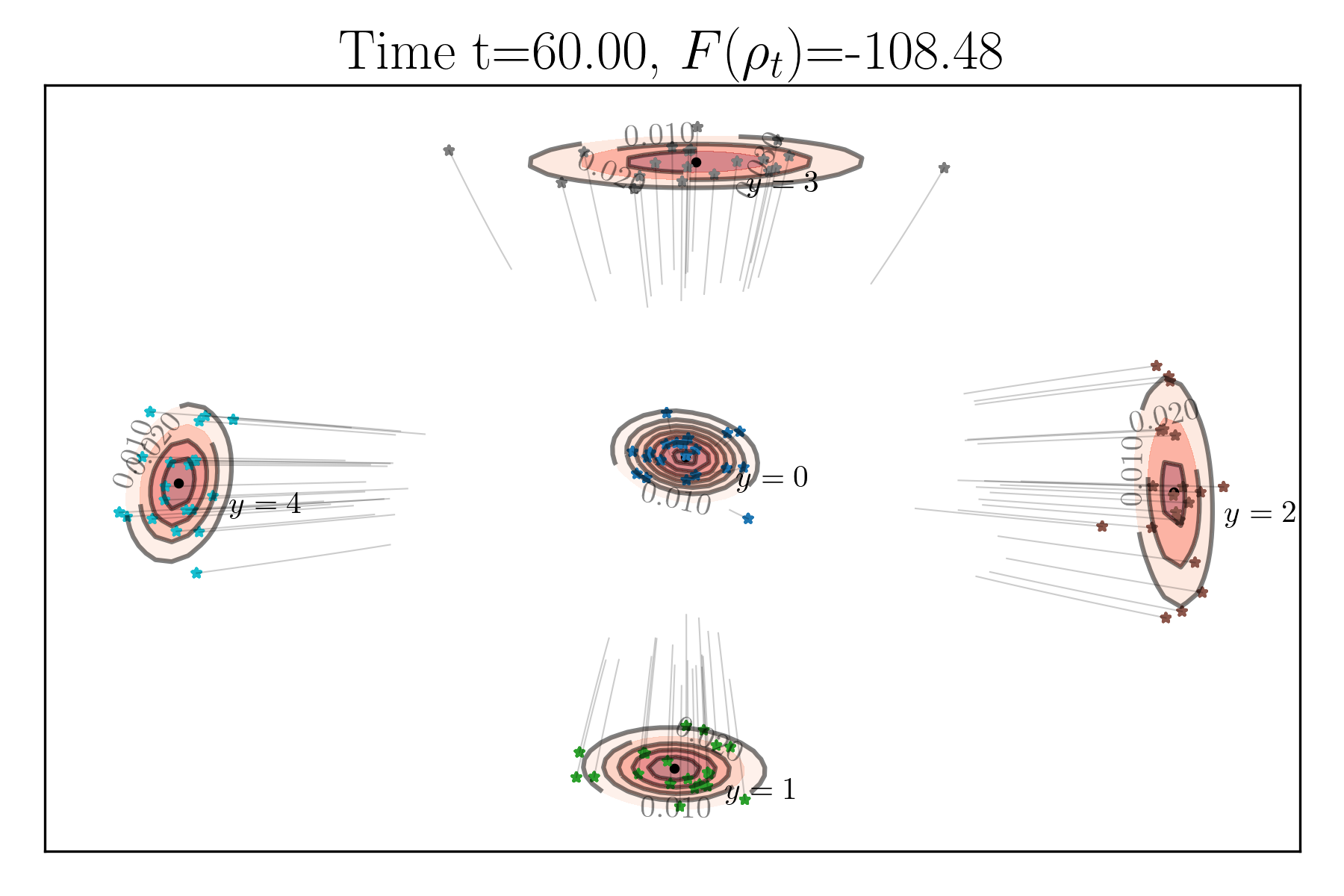}
    \caption{Functional: $F(\rho) =  \iint -||\x-\x'\|^2 \mathds{1}_{y\neq y'} \dif \rho(z)\dif\rho(z') $, \textsc{sgd} optimizer. }\label{fig:functionals_gaussians_3}
	\vspace{0.5cm}
	\end{subfigure}
    \begin{subfigure}{\linewidth}
        \includegraphics[width=0.33\linewidth, trim={0.3cm 0.3cm 0.3cm 0.3cm},clip]{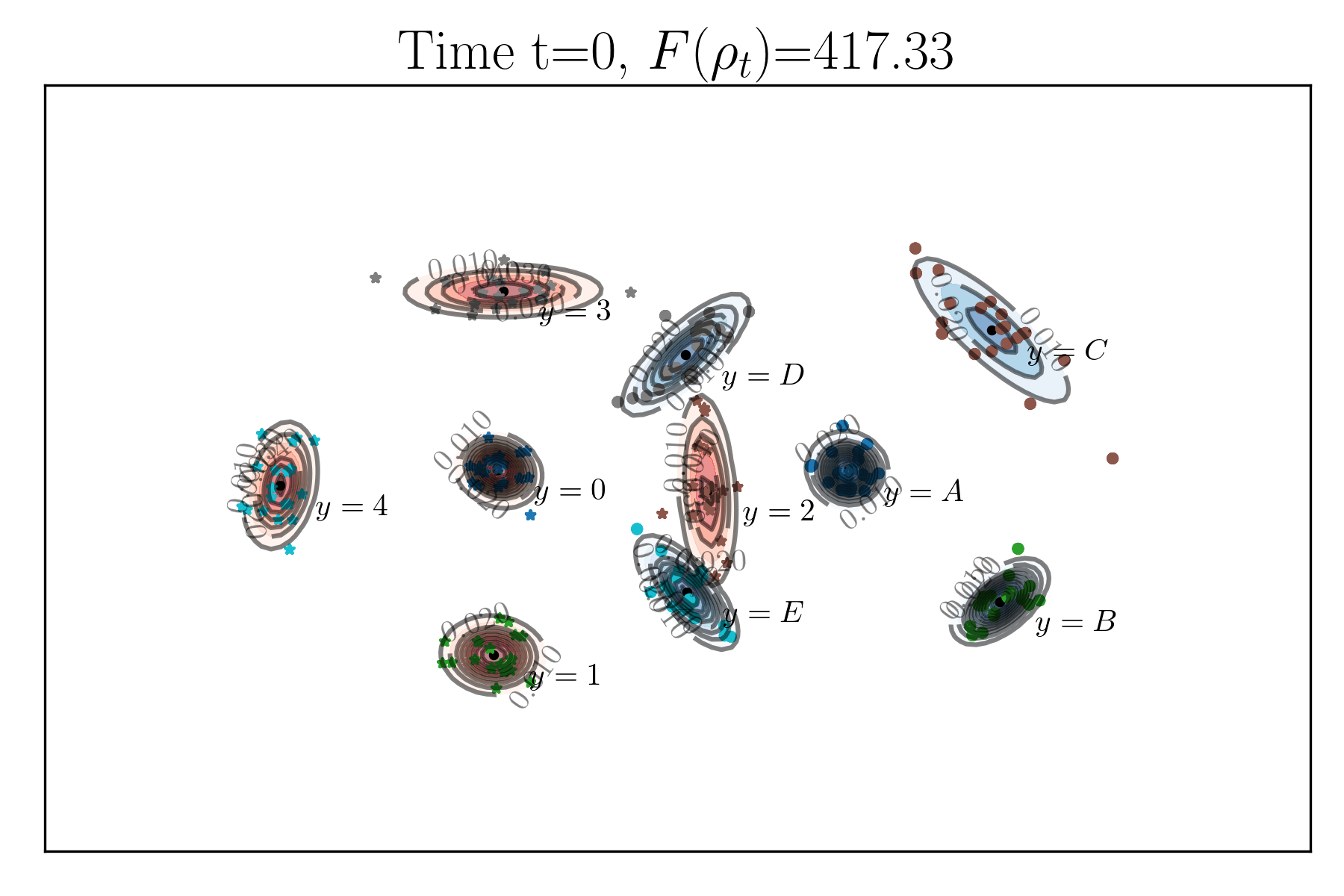}%
        \includegraphics[width=0.33\linewidth, trim={0.3cm 0.3cm 0.3cm 0.3cm},clip]{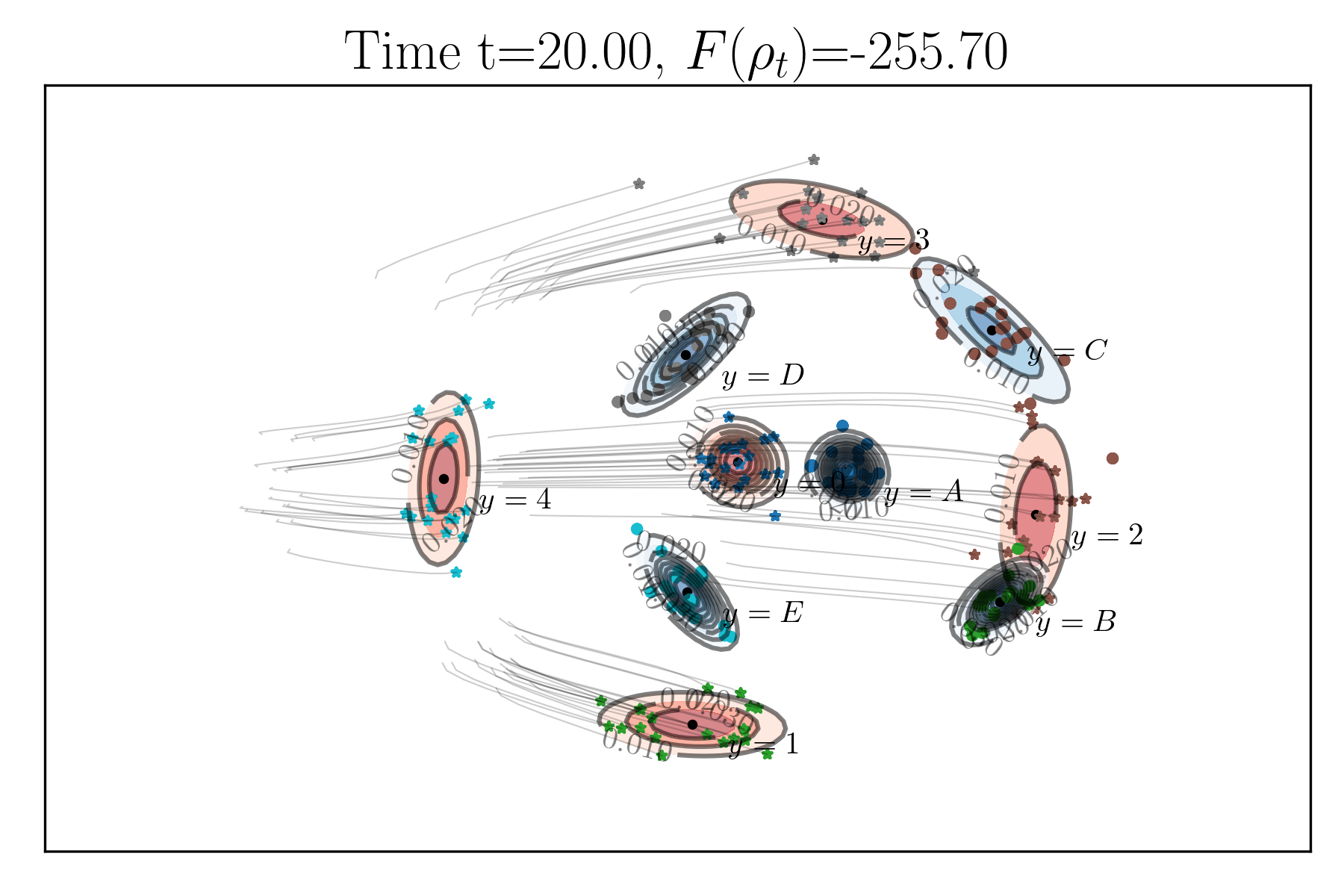}%
        \includegraphics[width=0.33\linewidth, trim={0.3cm 0.3cm 0.3cm 0.3cm},clip]{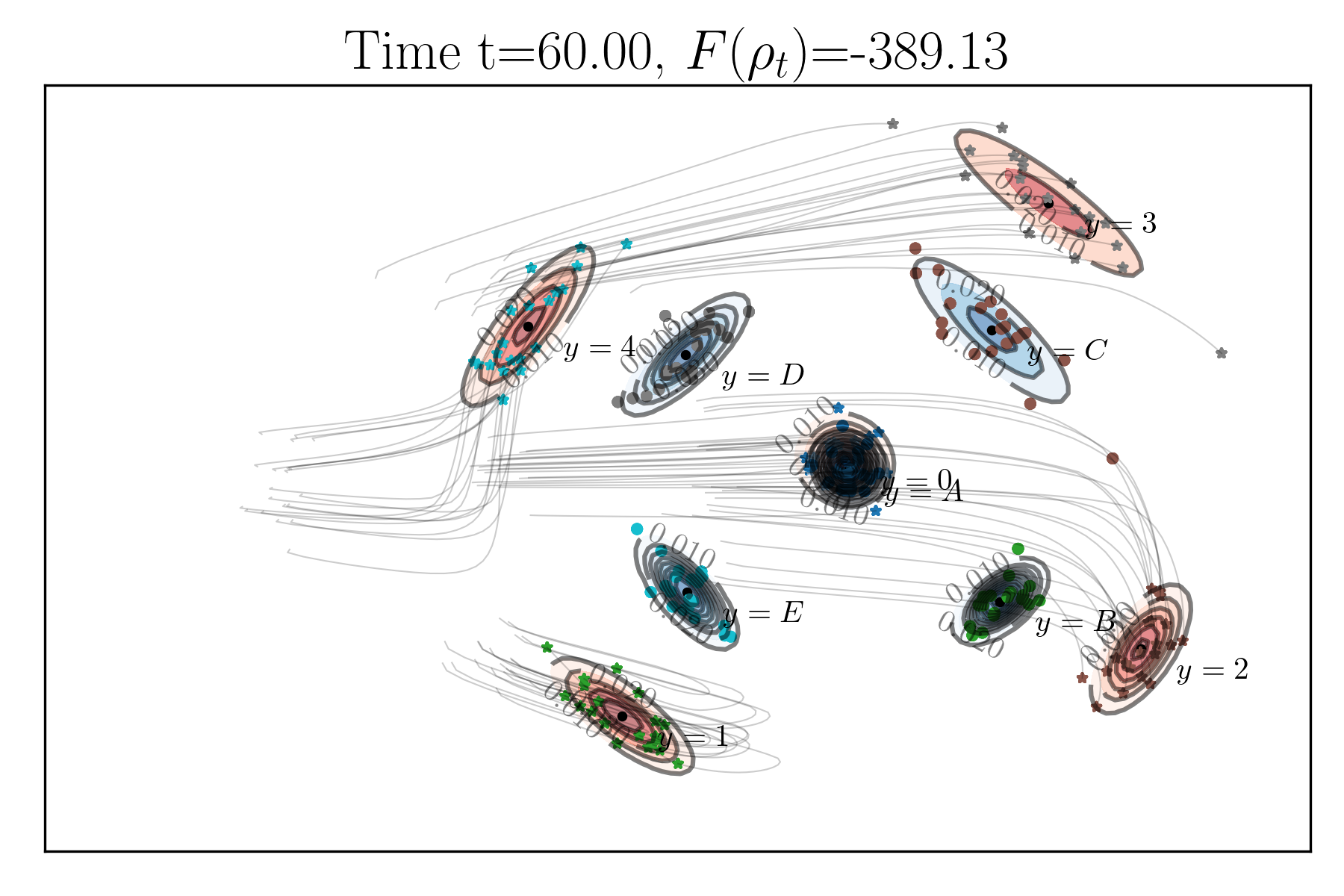}
    \caption{Functional: $F(\rho) = \text{OTDD}(\mathrm{D}_{\rho}, \mathrm{D}_{\beta}) + \lambda \iint -||\x-\x'\|^2 \mathds{1}_{y\neq y'} \dif \rho(z)\dif\rho(z') $, \textsc{sgd} optimizer. }\label{fig:functionals_gaussians_4}
	\end{subfigure}	
    \caption{Gradient flows starting from dataset $\rho_0$ (red) advecting towards $\beta$ (blue) driven by different functionals, using \textsc{sgd}+\texttt{jd-vl} dynamics in all cases.}\label{fig:functionals_gaussians}
\end{figure*}

For the simple synthetic dataset example of Section~\ref{sec:implementation}, we show a comparison of the three types of flow dynamics in Figure~\ref{fig:flow_methods_gaussians}, and experiments with various types of functionals in Figure~\ref{fig:functionals_gaussians}.

\clearpage
\pagebreak

\section{Additional Experimental Results on *\textsc{nist} Flows}\label{sec:additional_nist}

\begin{figure}[H]
    \centering
    \includegraphics[width=\textwidth]{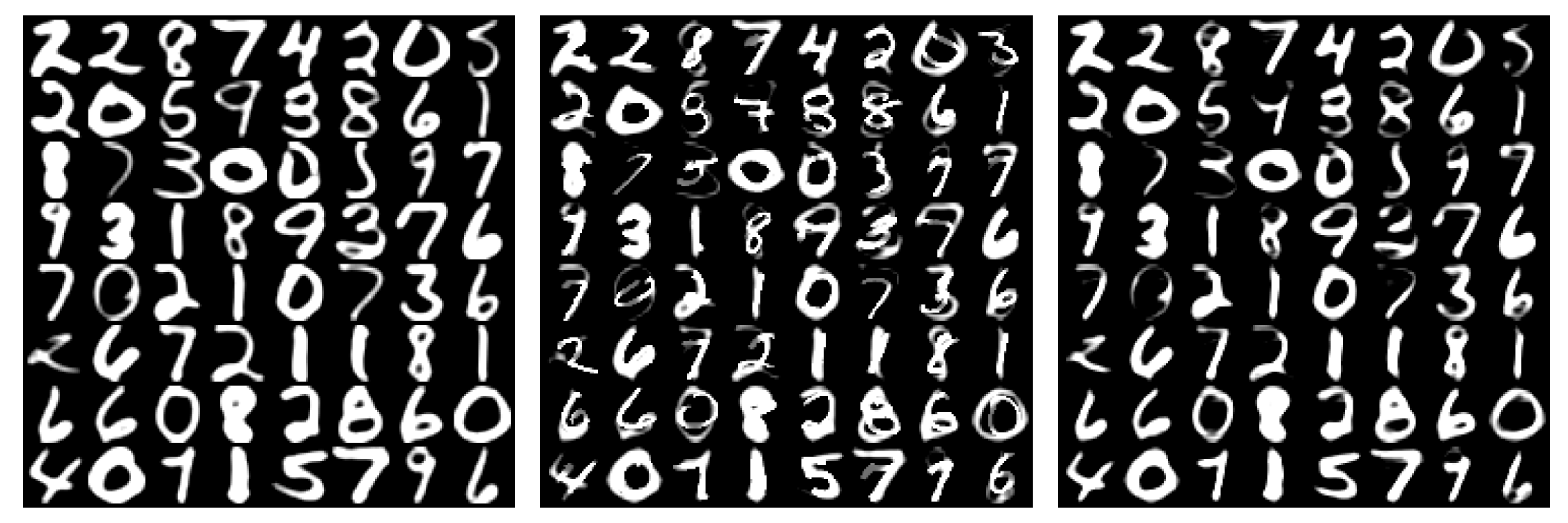}
    \caption{\textbf{Left}: initial particles $\x_0$ taken from \usps. \textbf{Center:} intermediate state of particles $\x_t$ after gradient flow driven by similarity-seeking functional $\cT_{\beta}(\rho) = \text{OTDD}(\mathrm{D}_{\rho}, \mathrm{D}_{\beta})$ for $\mathrm{D}_{\beta}:$\mnist. \textbf{Right}: particles mapped by using a parametric approximation of $h_{\text{flow}}$ learnt from data.}
    \label{fig:mnist_mapped}
    \vspace{-0.2cm}
\end{figure}

In Figure~\ref{fig:mnist_mapped} we show the effect of the transformation $h_{\theta}$ parametrized as a neural network and learnt from data to mimic the effect of the flow mapping $h_{\text{flow}}: \x_0 \mapsto \x_T$. 

As described in Section~\ref{sec:transfer_mnist}, we use gradient flows to approach to problem of transfer learning. Figure~\ref{fig:mnist_adapt_extra} shows results on the 5- and 10-shot tasks on the \textsc{nist} datasets. Notably, the results follow a similar trend as Figure~\ref{fig:mnist_adapt}, although, as expected by the smaller target datasets, the classification errors are higher.

\ifbool{twocol}{
\begin{figure}[H]
	\centering
    \includegraphics[width=0.5\linewidth, trim={0.7cm 0.75cm 0.74cm 1.7cm}, clip]{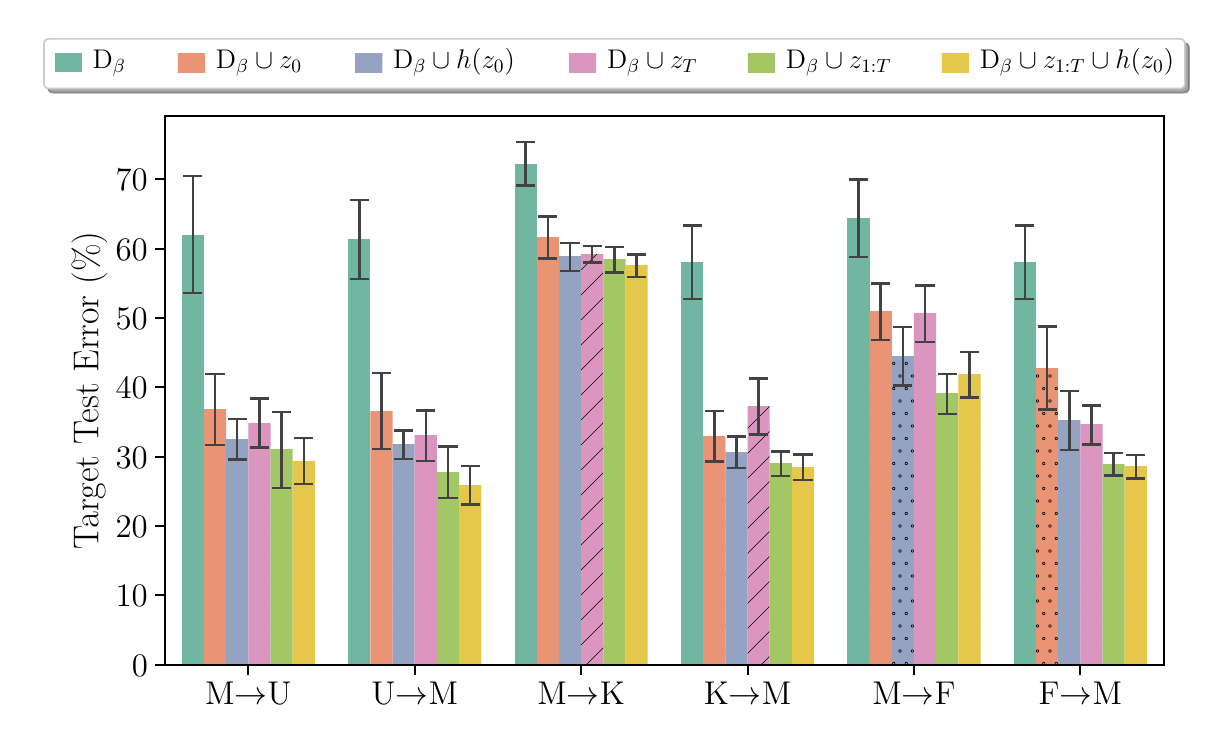}\\
    \includegraphics[width=0.5\linewidth, trim={0.7cm 0.75cm 0.74cm 0.6cm}, clip]{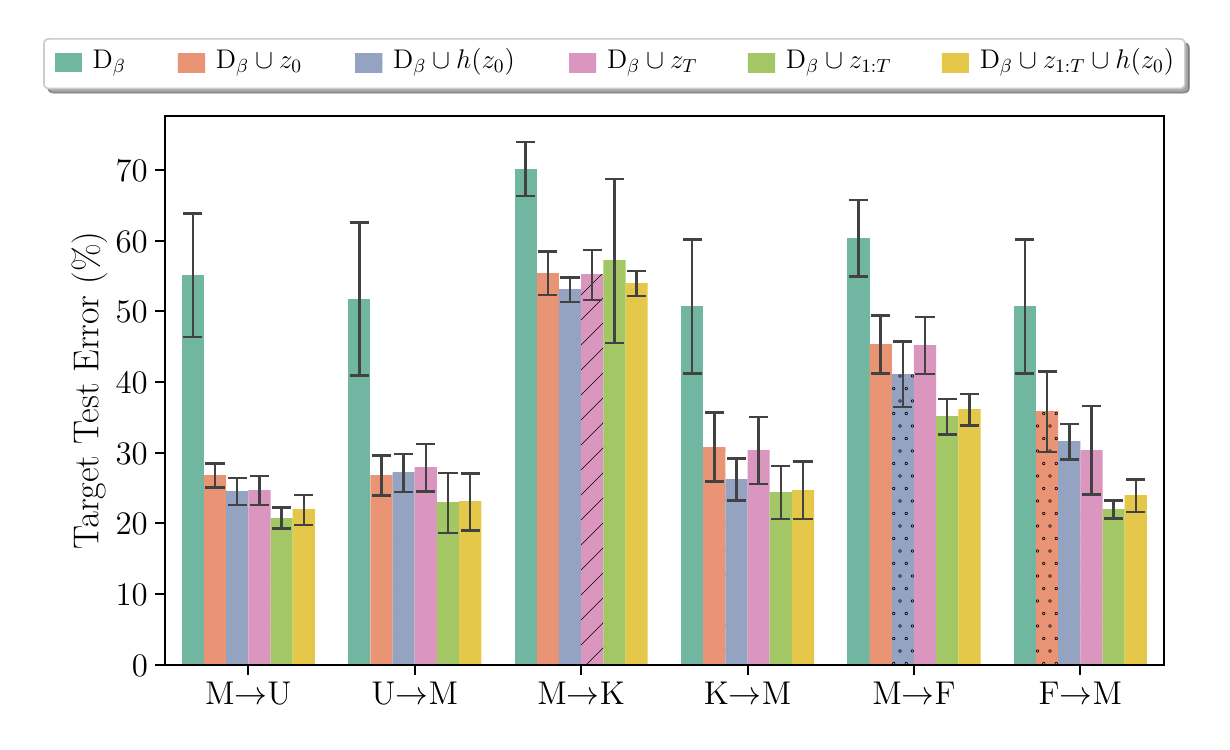}
	\caption{Transfer learning results on image datasets for $5$-shot (top) and $10$-shot (bottom) tasks. See Section~\ref{sec:transfer_mnist} for symbol key.}\label{fig:mnist_adapt_extra}
\end{figure}
}{
\begin{figure}[H]
	\centering
    \includegraphics[width=0.5\linewidth, trim={0.7cm 0.75cm 0.74cm 1.7cm}, clip]{figs/appendix/adaptation_plot_all_50.pdf}\\
    \includegraphics[width=0.5\linewidth, trim={0.7cm 0.75cm 0.74cm 0.6cm}, clip]{figs/appendix/adaptation_plot_all_100.pdf}
	\caption{Transfer learning results on image datasets for $5$-shot (top) and $10$-shot (bottom) tasks. See Section~\ref{sec:transfer_mnist} for naming key.}\label{fig:mnist_adapt_extra}
\end{figure}
}

\section{Additional Miscellaneous Experimental Results}\label{sec:additional_quant}

\begin{figure}[H]
    \centering
    \includegraphics[width=0.33\linewidth]{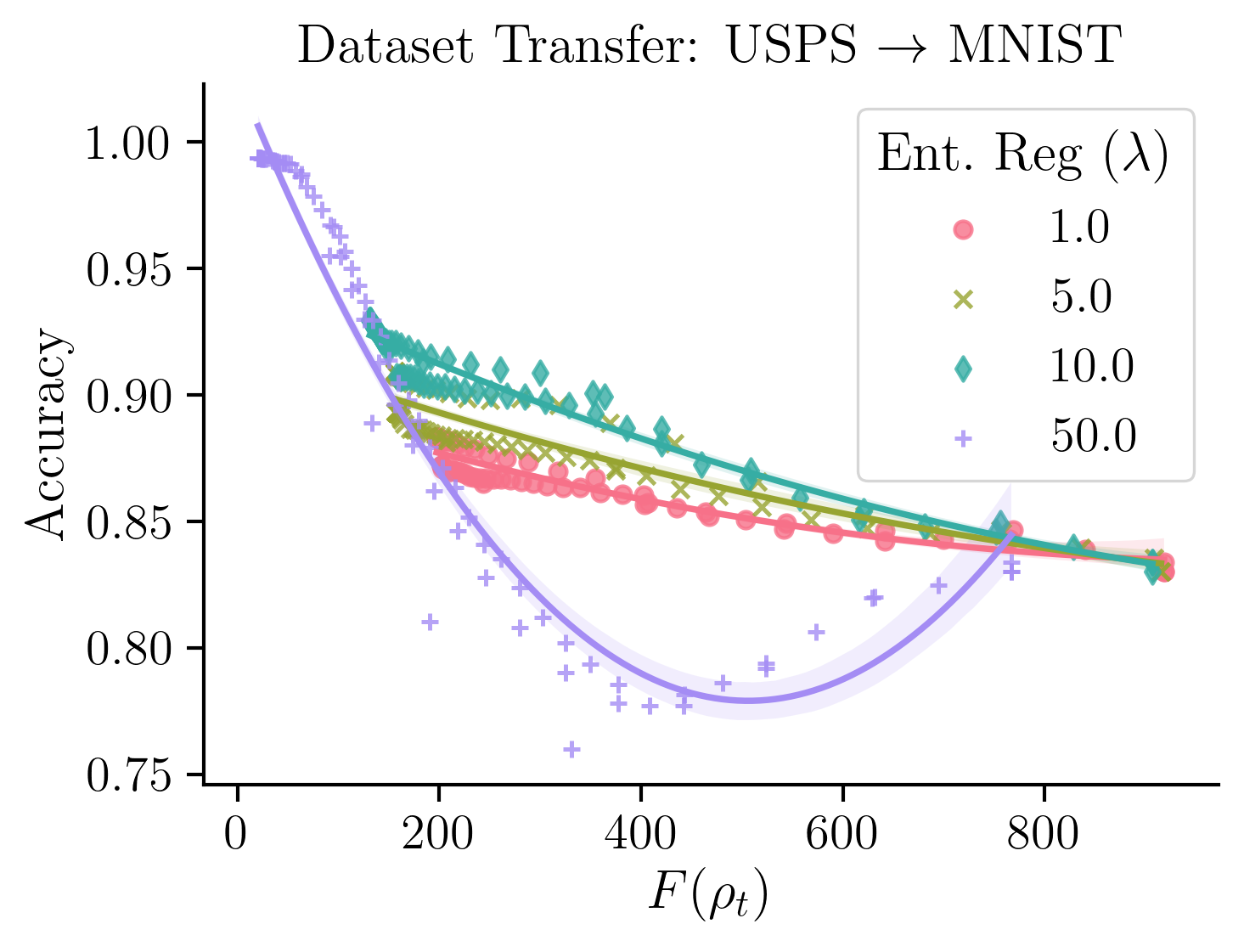}%
    \includegraphics[width=0.33\linewidth]{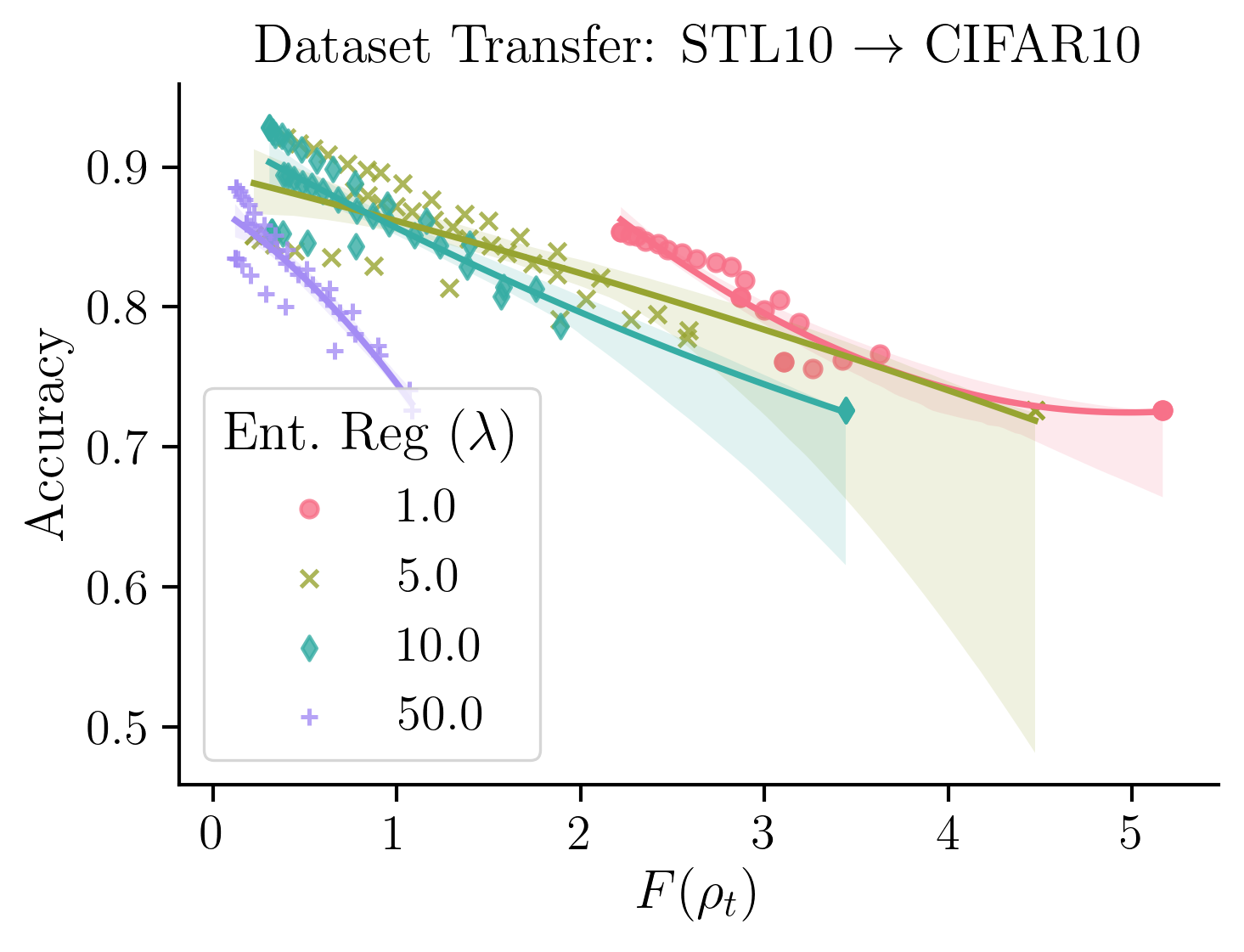}%
    \includegraphics[width=0.33\linewidth]{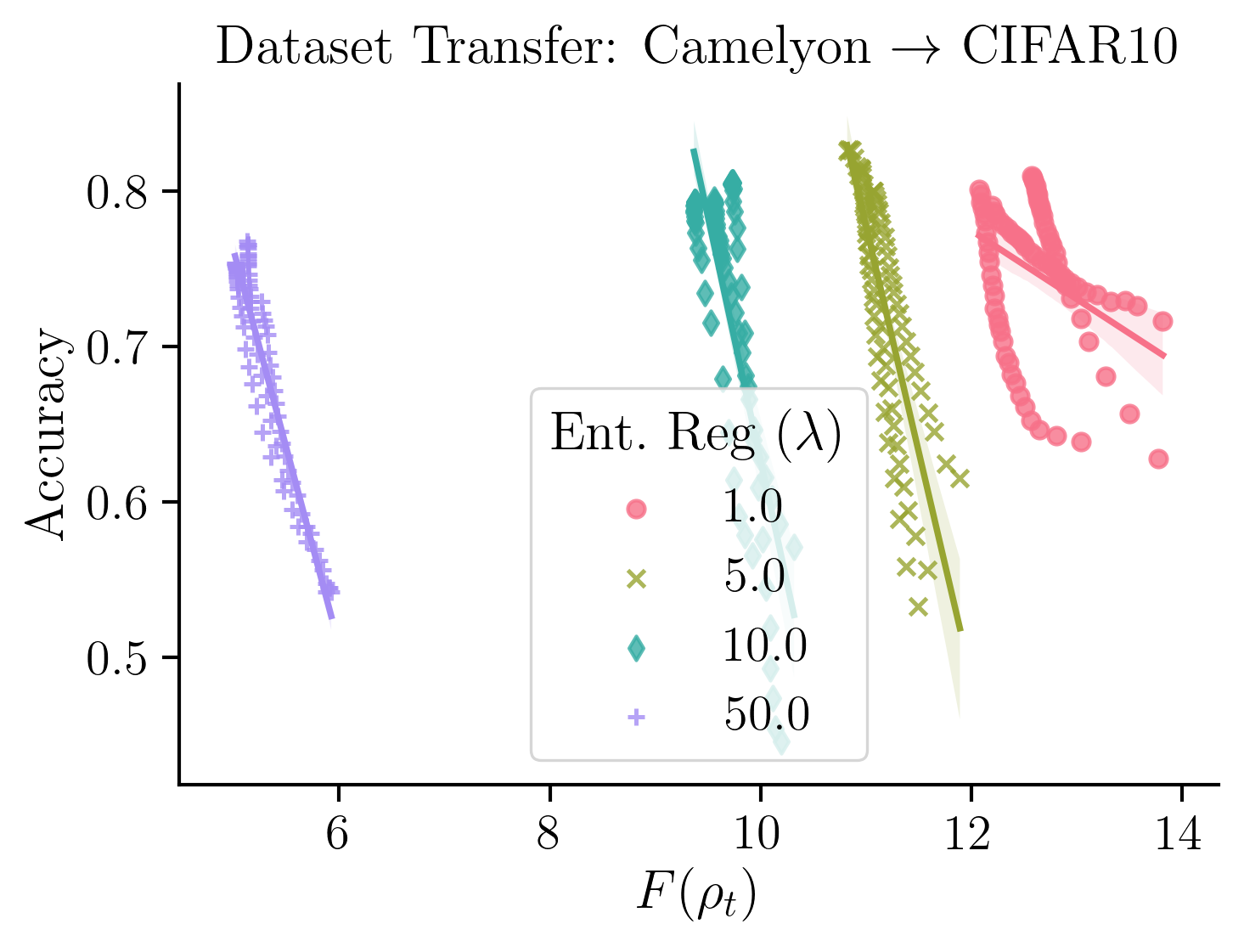}%
    \caption{\textbf{Qualitative evaluation of flows}: for the same setting as in Figure~\ref{fig:oracle_experiments}, we plot here the flow objective against the accuracy. The strong correlation between these quantities shows that the OTDD functional provides a good proxy for dataset transfer quality. The relation between these quantities is mostly linear and monotonic, except in configurations with an over-regularized OTDD objective, as shown for the $\lambda=50$ curve in the first plot.}
    \label{fig:oracle_experiments_correlations}
\end{figure}

\begin{figure}[H]
    \centering
    \includegraphics[width=\linewidth, trim={0 12cm 11cm 0}, clip]{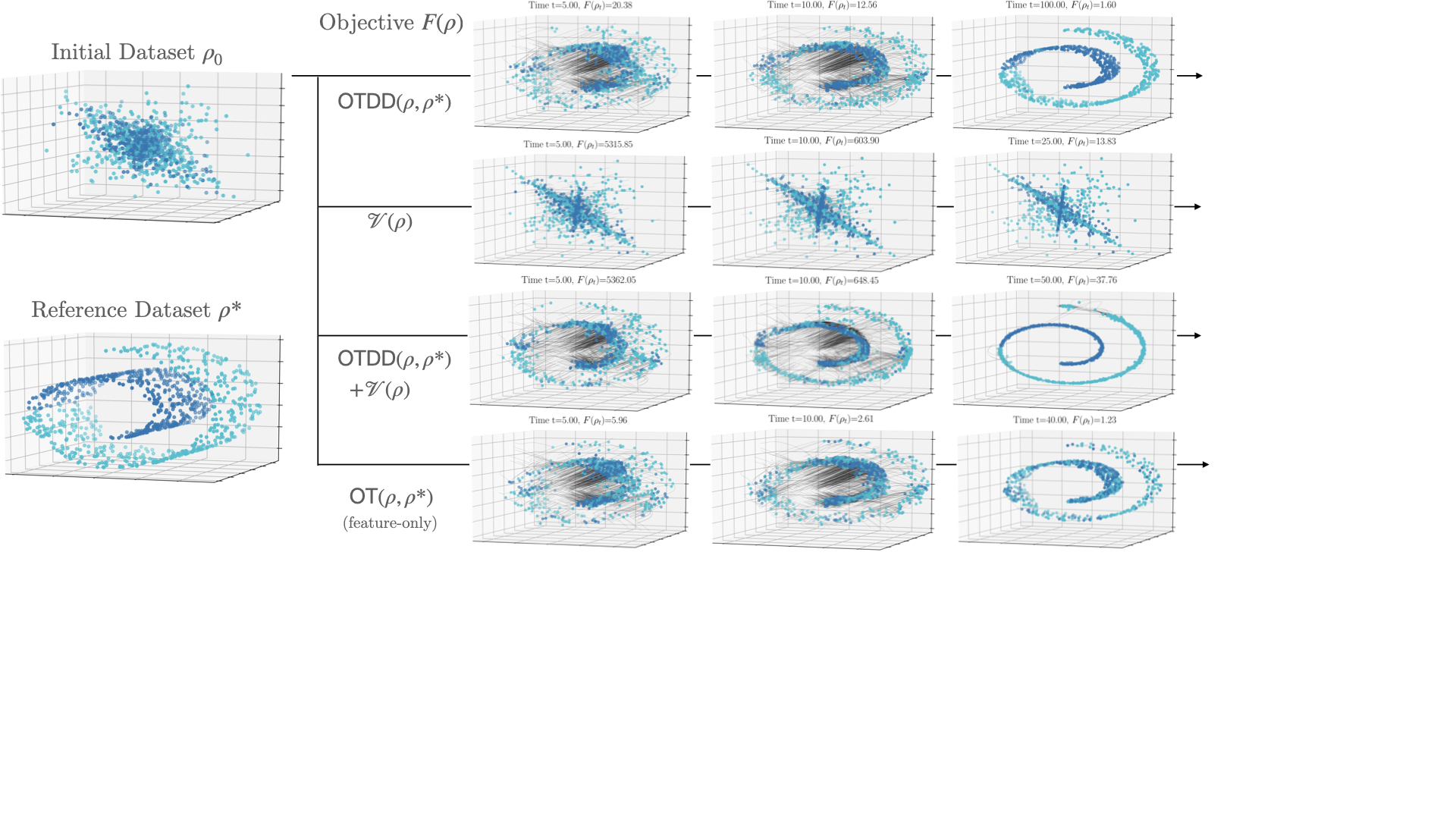}
    \caption{\textbf{Shaping datasets via flows}: our framework allows for simple and principled transformation of classification datasets by following the gradient flow of a functional objective, such as: similarity to a reference dataset (shown here as $\textrm{OTDD}(\cdot, \rho^*)$), a function enforcing collapse along a given dimension (shown here as $\mathcal{V}(\rho)$), or a combination thereof. In the bottom row we show the flow obtained with a vanilla (feature-only) optimal transport distance functional, which unsurpisingly misses the important class structure of the dataset.}
    \label{fig:additional_shaping}
\end{figure}

\clearpage

\end{document}